\documentclass[pdflatex,sn-mathphys-num]{sn-jnl}

\usepackage{amsmath,amssymb,amsfonts}
\usepackage{amsthm}
\usepackage[title]{appendix}
\usepackage{xcolor}
\usepackage{nicefrac}
\usepackage{booktabs}
\usepackage{tabularx}
\usepackage{algorithm}
\usepackage{algpseudocode}
\usepackage{microtype}

\theoremstyle{thmstyleone}
\newtheorem{theorem}{Theorem}
\newtheorem{lemma}[theorem]{Lemma}
\newtheorem{corollary}[theorem]{Corollary}
\newtheorem{assump}{Assumption}

\theoremstyle{thmstyletwo}
\newtheorem{remark}{Remark}

\theoremstyle{thmstylethree}
\newtheorem{definition}[theorem]{Definition}

\newcommand{\oo}[1]{{\cal O}(#1)}
\newcommand{\tildeo}[1]{\tilde{\cal O}(#1)}
\newcommand{\pp}[1]{\mathbb{P}(#1)}
\newcommand{\ee}[1]{\mathbb{E}[#1]}
\newcommand{\EE}[1]{\mathbb{E}\Big[#1\Big]}
\newcommand{\Ew}[1]{\mathbb{E}_W\Big[#1\Big]}
\newcommand{\sq}[1]{\Big[#1\Big]}
\newcommand{\normnu}[1]{\|#1\|_*}
\newcommand{\normsp}[1]{\|#1\|_2}
\newcommand{\normf}[1]{\|#1\|_{\scriptscriptstyle \textrm{\upshape F}}}
\newcommand{\norm}[1]{\|#1\|}
\newcommand{\normtv}[1]{\|#1\|_{\mathrm{TV}}}
\newcommand{\NORM}[1]{\Big\|#1\Big\|}
\newcommand{\inpf}[1]{\langle#1\rangle_{\scriptscriptstyle \textrm{\upshape F}}}
\newcommand{\rank}[1]{\operatorname{rank}(#1)}
\newcommand{\mix}{\mathrm{mix}}
\DeclareMathOperator*{\argmax}{arg\,max}
\DeclareMathOperator*{\argmin}{arg\,min}
\newcommand{\ind}[1]{\mathbf{1}\{#1\}}
\newcommand{\normd}[1]{\|#1\|_{\dagger}}

\hypersetup{
  hidelinks,
  pdftitle={Finite-Time Analysis of Importance-Corrected Momentum Actor-Critic over a Markovian Trajectory},
  pdfauthor={Yanjie Dong, Gang Wang, ShiSheng Cui},
  pdfkeywords={Actor-critic methods, importance-corrected momentum, least-squares temporal difference, Markovian sampling}
}

\begin{document}

\title[Finite-Time Analysis of Importance-Corrected Momentum Actor-Critic]{Finite-Time Analysis of Importance-Corrected Momentum Actor-Critic over a Markovian Trajectory}

\author*[1]{\fnm{Yanjie} \sur{Dong}}\email{ydong16@ece.ubc.ca}
\author[2]{\fnm{Gang} \sur{Wang}}\email{gangwang@bit.edu.cn}
\author[2]{\fnm{Shisheng} \sur{Cui}}\email{css@bit.edu.cn}

\affil*[1]{\orgdiv{School of Cyber-Science and Engineering}, 
	\mbox{\orgname{Xi'an Jiaotong University}, \orgaddress{\city{Xi'an}, \country{China}}}}
\affil[2]{\orgdiv{State Key Laboratory of Autonomous Intelligent Unmanned Systems},
  \orgname{Beijing Institute of Technology}, \orgaddress{\city{Beijing}, \country{China}}}

\abstract{We analyze importance-corrected momentum actor-critic methods on a single continuing Markovian trajectory. Each policy collects $T$ transitions starting where the previous frame ended. A ridge-regularized pathwise least-squares temporal-difference critic fitted by Newton-Schulz inversion on one frame supplies value estimates for the next.
Under uniform mixing and bounded, Lipschitz policy scores, the policy-gradient bias splits into a stationary-to-discounted visitation mismatch and a frame bias. In the momentum tracking error, the mismatch telescopes to a single term at the current policy and can leave a nonzero error floor. With the starting state and critic fixed, both the magnitude of the frame bias and its Lipschitz constant with respect to the policy parameters are $\mathcal{O}(T^{-1})$.
For fixed $T$ and the prescribed schedules, the bounds on the expected policy-gradient norm averaged over $K$ updates have transients of order $\mathcal{O}(K^{-1/3})$ for normalized steps and $\mathcal{O}(K^{-1/4})$ for unnormalized steps, with constants depending on $T$. The analysis covers a Muon-type actor, normalized and unnormalized stochastic gradient descent with momentum, and stabilized AdaNorm. For an $m\times n$ momentum matrix, $\mathcal{O}(\log(1+\min\{m,n\}))$ Newton-Schulz orthogonalization recursions retain a constant fraction of the exact polar step's inner product with the momentum.}

\keywords{Actor-critic methods, importance-corrected momentum, least-squares temporal difference, Markovian sampling}

\maketitle

\section{Introduction}\label{sec:intro}
Actor-critic (AC) methods improve the policy along a policy-gradient (PG) estimate where a learned critic replaces Monte Carlo returns to reduce gradient variance~\citep{Konda1999,Sutton1999,Sutton2018}.~Convergence analyses of AC methods bound the propagation of critic error into actor update asymptotically~\citep{Konda1999,Zhang2020a} and non-asymptotically for double-loop~\citep{Qiu2021,Kumar2023,Zhang2021}, two-timescale~\citep{Wu2020,Hong2023,Zeng2024,Zeng2024a}, and single-timescale~\citep{Olshevsky2023,ChenDec.2023,Shen2022} settings.~The vanilla critic performs \mbox{semi-gradient} temporal-difference (TD) policy evaluation~\citep{Bhandari2018}.
Moreover, the vanilla critic can be accelerated by adaptive stepsizes and momentum accelerate~\citep{Sun2022,Dong2024}, and freed from knowledge of the mixing time by multi-level Monte Carlo PG estimators~\citep{DorfmanJuly2022,Suttle2023,Patel2024mlmc,Ganesh2025}.

{\bfseries Research gaps.}~While current convergence analyses of AC methods update the actor by unnormalized gradients~\citep{Wu2020,Xu2020a,Xu2020b,Qiu2021,Kumar2023,Olshevsky2023,ChenDec.2023}, recent actor designs depart from such steps in two ways.~First, an importance-corrected momentum (ICM) estimator~\citep{Huang2020} can reduce momentum variance by reweighting the previous PG estimator on the current trajectory via importance sampling.~Second, the momentum could be normalized~\citep{CutkoskyMehta2020} and orthogonalized~\citep{Jordan2024, Chang2025muon, Lion2026}.~For example, Muon~\citep{Jordan2024} applies the Newton-Schulz orthogonalization (NSO)~\citep{Kelley1995} to the momentum matrix and moves the model parameter along an approximation of momentum polar factor.~Moreover, a normalized step can bound the importance weights without further assumptions~\citep{Barakat2023general}.~To our knowledge, such orthogonalization is absent from policy updates reshaped by Fisher preconditioning~\citep{kakade2001natural,peters2008natural,Agarwal2021,Khodadadian2023,Wu2017acktr}, mirror descent~\citep{Yang2022,Lan2023}, trust-region and proximal surrogates~\citep{Schulman2015,schulman2017proximal}, momentum~\citep{Huang2020,Chen2024}, and Adam-type steps~\citep{Xiong2021}.~Besides, \mbox{Muon-based} reinforcement learning has only been investigated empirically~\citep{Ruan2026muonrl,Fan2026pion}.~The current finite-time analyses of normalized and orthogonalized momentum methods assume unbiased independent and identically distributed (i.i.d.)\ gradients, e.g., normalized stochastic gradient descent (SGD) with momentum~\citep{CutkoskyMehta2020}, Muon with finite NSO~\citep{kim2026convergence} or variance reduction~\citep{Chang2025muon}, and linear-minimization-oracle methods~\citep{Pethick2025lmo,Kovalev2025,Shen2025muon,Riabinin2025gluon}.

The requirement of unbiased i.i.d.\ gradients cannot be met by reinforcement learning tasks along a single trajectory, whose PG estimators exhibit bias and correlated noise: the score-weighted TD-based PG estimator  relies on a critic that is still being learned, and the sample trajectory is drawn from a Markov chain induced by the evolving actor.
Prior works leave the underlying three aspects unaddressed when normalized and orthogonalized ICM actors learn along a single trajectory that consists of $T$ transitions per consecutive frame.
\begin{itemize}
\item[i.] Analyses of normalized and orthogonalized momentum are based on unbiased gradients~\citep{CutkoskyMehta2020,Barakat2023general,kim2026convergence,Chang2025muon,Pethick2025lmo,Kovalev2025,Shen2025muon,Riabinin2025gluon}, analyses of biased stochastic approximation control the bias only through its norm~\citep{Karimi2019biased,Ajalloeian2020biased,Demidovich2023zoo}, and variance-reduced methods with biased estimators assume a mean field that depends on the current iterate alone~\citep{Hu2020biased,Khanduri2021near}, whereas the frame estimates of an AC loop have a bias that depends on the iterate and on a start state generated by the previous frame.
\item[ii.] Existing finite-time AC analyses are confined to unnormalized steps, whose progress is controlled through second moments~\citep{Wu2020,Olshevsky2023,ChenDec.2023}.
Normalized steps, in contrast, necessitate first-moment control of the momentum's tracking error, and ICM actors are not accommodated.
\item[iii.] ICM has been analyzed for i.i.d.\ samples~\citep{Cutkosky2019} and for restarted episodes~\citep{Huang2020,Barakat2023general},whereas along a single trajectory the kernel evolves with the actor and the importance weight corrects the actions of a frame but not its start state, which couples consecutive frames.
\end{itemize}

{\bfseries Our contributions.}~Different from the AC framework of~\citet{Kumar2023} that combines a fixed SGD actor with several critic under i.i.d.\ sampling, our AC framework averages the PG estimator over a frame attenuates the critic error by the frame length such that the ICM-based actor update determines the convergence rate under a single Markovian trajectory.
Therefore, by fixing the critic, the PG estimator, and the ICM, we investigate how different actor updates govern the convergence rate and the dependence on the rank of the actor parameter.
The critic is a ridge-regularized pathwise least-squares TD (LSTD) critic, i.e., the pathwise LSTD critic of~\citet{Lazaric2012} with a ridge term.
By leveraging Newton-Schulz inversion (NSI) that requires only general matrix-matrix multiplications (GEMMs), we approximate the critic parameter while fully exploiting the parallel computing power of graphics processing units (GPUs).
Our contributions are summarized as follows.
\begin{itemize}
	\item \emph{An inverse-free LSTD critic.} Our proposed inverse-free critic can approximate the ridge-regularized pathwise LSTD solution with GEMMs alone for every ridge parameter $\lambda>0$ and every frame length $T$; therefore, does not require stepsize, knowledge of the mixing time, and rank condition on the actor parameter.
	Theorem~\ref{theorems:lstd-critic-error} quantifies both sources of critic error: the NSI error decays superlinearly in the number $P_1$ of NSI recursions, and the mean-square Markovian sampling error is of order $\nicefrac{\tau_{\mix}}{T}$. 
	\item \emph{Generic bounds over a single Markovian trajectory.} Conditioned on the start state of each frame and the lagged critic, the noise of the PG estimators forms a martingale-difference sequence across frames, and their bias has two parts: the stationary-to-discounted mismatch (that enters the convergence rate only once due to the cancellation of importance correction between consecutive iterates) and the frame bias (which is of order $\nicefrac{1}{T}$ and varies with the actor parameter at a rate of order $\nicefrac{1}{T}$).
	With such decomposition, every actor update inherits a finite-time convergence rate in the nuclear norm: an $\oo{K^{-1/3}}$ transient for normalized actor updates and an $\oo{K^{-1/4}}$ transient for unnormalized ones (Theorems~\ref{theorems:normalized-actor} and~\ref{theorems:unnormalized-actor}), plus a floor formed by the mismatch and the frame bias.
	
	\item \emph{Instances and comparisons.} We apply our theoretical results to normalized actor updates, such as Muon and normalized SGD with momentum (NSGDM), and to unnormalized ones, such as SGD with momentum (SGDM) and an adaptive-stepsize actor (AdaNorm).	
	We establish that the Muon actor update retains at least $78\%$ of the first-order progress of the exact polar step along the momentum without a conditioning assumption after $\oo{\log r}$ NSO recursions with the actor parameter rank $r\le\min\{m,n\}$.
	Comparing the instances, we identify two effects: normalization alone explains the faster rate, and orthogonalization with a dimension-free stepsize confines the factor $r^{1/2}$ to the noise and correction terms.
	We also characterize when orthogonalization guarantees a larger noiseless ascent than normalization: exactly when the nuclear rank of the gradient exceeds the ratio of the spectral and Frobenius curvature constants (Section~\ref{sec:comparison}).
\end{itemize}
	
To our knowledge, Theorems~\ref{theorems:normalized-actor} and~\ref{theorems:unnormalized-actor} and their instances are the first finite-time convergence rates for ICM actors along a single Markovian trajectory with a bootstrapped critic.~The floor of our bounds isolates the stationary-to-discounted mismatch of uniformly weighted samples, a bias that practical AC implementations share~\citep{Thomas2014bias,Nota2020policy,Zhang2022discounting}; every other floor term decays at least as $\nicefrac{1}{T}$.~Appendix~\ref{apdx:notation} collects our notation and defines the stationarity criterion.

\section{Problem Description}\label{sec:problem}
We consider a discounted Markov decision process $({\cal S}, {\cal A}, \mathbb{P}, r, \gamma)$ with 
a state space ${\cal S}$, 
an action space ${\cal A}$, 
an unknown transition kernel that maps each state-action pair $(s,a)$ to a distribution $\pp{\cdot|s,a}$ over ${\cal S}$, a bounded reward $r:{\cal S}\times{\cal A}\rightarrow[-R_r,R_r]$, and a discount factor $\gamma\in(0,1)$.
A policy $\pi_W$, parameterized by a matrix $W\in\mathbb{R}^{m\times n}$, maps each state $s$ to a distribution $\pi_W(\cdot|s)$ over ${\cal A}$.
Under parametric policy $\pi_W$, the state-value function and the \emph{Q}-function are
\begin{equation}\label{eq:action-value-bellman}
\begin{split}
	V_W(s) &= \Ew{\sum\nolimits_{t=0}^{\infty}\gamma^tr(s_t,a_t)\,\Big|\,s_0 = s} \\
	Q_W(s,a) &= \Ew{\sum\nolimits_{t=0}^{\infty}\gamma^tr(s_t,a_t)\,\Big|\,s_0=s,a_0=a} = r(s,a)+\gamma\ee{V_W(s')}
\end{split}
\end{equation}
where $a_t\sim\pi_W(\cdot|s_t)$,
$s_{t+1}\sim\pp{\cdot|s_t,a_t}$, and the last equality is the Bellman equation with $s'\sim\pp{\cdot|s,a}$.

{\bfseries Policy gradient theorem.}~Given an initial-state distribution $\rho$, our objective is to maximize the normalized expected discounted return $J(W) := (1-\gamma)\,\mathbb{E}_{s\sim\rho}[V_W(s)]$, whose gradient is given by the policy gradient theorem~\citep{Sutton2018} as
\begin{equation}\label{eq:policy-gradient}
\begin{split}
\nabla J(W) 
&= \EE{ [Q_W(s_t, a_t) - V_W(s_t)] \nabla\log\pi_W(a_t | s_t) } \\
&= \EE{ \sum\nolimits_{l=0}^{T-1}\gamma^l\delta_{W,t+l}  \nabla\log\pi_W(a_t | s_t) }
\end{split}
\end{equation}
where $s_t$ is drawn from the normalized discounted state-visitation distribution $\zeta_W(\cdot) := (1-\gamma) \sum_{t\ge0} \gamma^{t} \pp{ s_{t} \in \cdot | s_{0} \sim \rho, \pi_W }$, the subsequent actions and states are generated by $\pi_W \otimes \mathbb{P}$, and the one-step TD error is $\delta_{W,t} := r(s_t,a_t)+\gamma V_W(s_{t+1})-V_W(s_t)$.

The second equality in \eqref{eq:policy-gradient} follows from the three facts: 
(i) $\ee{\delta_{W,t} | s_t,a_t} = Q_W(s_t,a_t) - V_W(s_t)$ by leveraging definitions in \eqref{eq:action-value-bellman}; 
(ii) $\ee{\delta_{W,t+l} | s_{t+l}} = 0$ and  
$\ee{\delta_{W,t+l} | s_t,a_t} = 0$ for all $l\ge1$ by Markov property; and 
(iii) the score $\nabla\log\pi_W(a_t | s_t)$ is a function of state-action pair $(s_t,a_t)$.
Note that the state-visitation distribution $\zeta_W$ differs in general from the stationary distribution $\mu_W$ of the $\pi_W$-induced Markov chain.

{\bfseries Linear function approximation.}~Since the advantage in \eqref{eq:policy-gradient} is unavailable in the model-free setting, a linear critic is used to approximate the state-value function as
\begin{equation}\label{eq:linear-value-approximation}
	V_v(s) = \phi^\top(s) v \approx V_W(s)
\end{equation}
where the feature embedding $\phi: {\cal S} \rightarrow \mathbb{R}^{d}$ satisfies $\normsp{\phi(s)} \le 1$.

Taking the expectation of the last equality in \eqref{eq:action-value-bellman} over $a\sim\pi_W(\cdot|s)$ and replacing $V_W$ with $V_v$, we obtain the Bellman equation of state value as
\begin{equation}\label{eq:linear-bellman}
	\phi^\top(s) v = \ee{ r(s,a) + \gamma \phi^\top(s') v | s }
\end{equation}
where the expectation is taken over $\pi_W \otimes \mathbb{P}$.

Note that the Bellman equation of state value \eqref{eq:linear-bellman} imposes one equation per state on $d$ unknowns and has a solution if and only if $V_W$ lies in the function class of the critic.
Hence, we fix the target $r(s,a)+\gamma\phi^\top(s')v^*$ and require the critic parameter $v^*$ to minimize the ridge-regularized mean squared error as 
\begin{equation}\label{eq:ridge-bellman-projection}
	v^* = \argmin_{u \in \mathbb{R}^{d}}\; \ee{ ( \phi^\top(s) u - r(s,a) - \gamma \phi^\top(s') v^* )^2 } + \lambda \normsp{u}^2
\end{equation}
where $\lambda > 0$ is the ridge parameter and the observation $o = (s, a, s') \sim \mu_W\otimes \pi_W \otimes \mathbb{P}$.

Since the target $r(s,a)+\gamma\phi^\top(s')v^*$ is independent of $u$, the objective in \eqref{eq:ridge-bellman-projection} is a quadratic function of $u$ with positive definite Hessian $2(\Sigma_\phi+\lambda I_d)$ due to the positive semidefiniteness of $\Sigma_\phi := \ee{ \phi(s)\phi^\top(s) }$.~For a fixed target, the objective is strongly convex in $u$ and has a unique minimizer.~Thus, the optimal critic $v^*$ solves \eqref{eq:ridge-bellman-projection} if and only if it satisfies the first-order optimality condition at $u = v^*$, i.e., the ridge-regularized Bellman equation
\begin{equation}\label{eq:population-lstd-system}
	\ee{ \delta_{v^*}(o) \phi(s) } - \lambda v^* = b - (A + \lambda I_d) v^* = 0
\end{equation}
where the TD error $\delta_{v}(o) := r(s,a) + [\gamma \phi(s') - \phi(s)]^\top v$, 
$A := \ee{ \phi(s) ( \phi(s) - \gamma \phi(s') )^{\top} }$, and 
$b := \ee{ \phi(s) r(s,a) }$.

Note that \eqref{eq:population-lstd-system} is the population ridge-regularized pathwise LSTD system that will guide the design of the empirical LSTD system per each frame in Section~\ref{sec:critic}.
It is worth stressing that \eqref{eq:population-lstd-system} with $\lambda = 0$ recovers the projected Bellman equation~\citep{Tsitsiklis1997}, whose empirical counterpart is the well-known LSTD critic estimator~\citep{Bradtke1996, Lagoudakis2003}.
Since the successor state $s'$ shares the distribution $\mu_W$ with the current state $s$, we obtain $x^\top (A+\lambda I_d) x \ge (1-\gamma) x^\top \Sigma_\phi x + \lambda\normsp{x}^2 > 0$ for all $x\in\mathbb{R}^d$~\citep[Lemma~3]{Bhandari2018}.
Therefore, the ridge-regularized Bellman equation \eqref{eq:population-lstd-system} and the fixed-point problem \eqref{eq:ridge-bellman-projection} have the unique solution $v^*(W) := (A + \lambda I_d)^{-1} b$ with $\normsp{v^*(W)} \le \nicefrac{\normsp{b}}{\lambda} \le \nicefrac{R_r}{\lambda}$.

\section{Generic ICM-AC over a Single Markovian Trajectory}\label{sec:method}
The critic and the actor learn from a single trajectory of the policy-induced Markov chain that consists of consecutive frames of a fixed length $T\ge1$, one per update, for $K$ updates.
More specifically, frame $k$ is the $T$-step trajectory $O_k = [o_{k,t}]_{t=0}^{T-1}$ with the observations $o_{k,t} = (s_{k,t},a_{k,t},s_{k,t+1})$ that are generated by the current policy $\pi_k := \pi_{W_k}$ from the last state of the previous frame, i.e., $s_{k,0} = s_{k-1,T}$ for $k\ge1$.
The warm-up frame $O_0$ is generated by $\pi_1$ from $s_{0,0}\sim\rho$.
Based on the $T$-step trajectory $O_k$, Sections~\ref{sec:critic}--\ref{sec:generic-alg} develop the inverse-free ridge-regularized pathwise LSTD critic, the PG estimator, and the ICM of the actor, and the generic ICM-AC algorithm, respectively.

\subsection{Inverse-Free Ridge-Regularized Pathwise LSTD Critic}\label{sec:critic}
The empirical estimates of $A$ and $b$ are $\hat A_k = \Phi_kY_k^\top$ and $\hat b_k = \Phi_kr_k$ with
$Y_k = \Phi_k-\gamma\Phi_k'$,
$\Phi_k = T^{-1/2}[\phi_{k,t}]_{t=0}^{T-1}\in\mathbb{R}^{d\times T}$,
$\Phi_k' = T^{-1/2}[\phi_{k,t}\ind{t<T}]_{t=1}^{T}$, and
$r_k = T^{-1/2}[r_{k,t}]_{t=0}^{T-1}\in\mathbb{R}^{T}$, where $\phi_{k,t} = \phi(s_{k,t})$ and $r_{k,t} = r(s_{k,t},a_{k,t})$.
Note that the last column  of $\Phi_k'$ drops the bootstrap of state $s_{k,T} $ of the frame such that the LSTD system involves only the features $\phi_{k,0},\ldots,\phi_{k,T-1}$.
Hence, for each positive ridge parameter $\lambda >0$, the matrices $\hat A_k+\lambda I_d$ and $H_k := Y_k^\top\Phi_k+\lambda I_{T}$ are non-singular with proofs relegated to Appendix~\ref{apdx:bk}.

Replacing the population quantities $A$ and $b$ in \eqref{eq:population-lstd-system} with $\hat A_k$ and $\hat b_k$, we obtain the empirical ridge-regularized pathwise LSTD solution as
\begin{equation}\label{eq:reduced-lstd-solution}
\hat v_k = (\hat A_k+\lambda I_d)^{-1}\hat b_k = \Phi_kH_k^{-1}r_k
\end{equation}
where the last equality is the push-through identity~\citep[p.~218]{Boyd2018} with the non-singular $H_k$.

Different from a semi-gradient TD(0) with weight decay $\lambda$, which targets the same $v^*$ by taking only one stochastic-approximation step per transition, the ridge-regularized pathwise LSTD critic solves the empirical optimal $\hat v_k$ of a frame in closed form to avoid critic stepsize tuning and the slow transient of stochastic updates.
Hence, the reduced $T\times T$ system lowers the cost of forming and factoring the LSTD system from $\oo{d^2T+d^3}$ to $\oo{dT^2+T^3}$ operations which is a significantly saving when $T\ll d$.
Since factoring $H_k$ uses GPUs far less efficiently than GEMMs, we choose to approximate $H_k^{-1}$ by the GEMM-only NSI recursion, which runs close to the peak throughput of GPUs, as
\begin{equation}\label{eq:newton-schulz-critic}
	E_{k,q} = I_{T}-H_kX_{k,q},\enspace
	X_{k,q+1} = X_{k,q}\sum\nolimits_{i=0}^{\kappa_1-1}E_{k,q}^i,\enspace
	v_{k,q} = \Phi_kX_{k,q}r_k
\end{equation}
where $\kappa_1\ge2$ is the NSI order, $X_{k,0} = \nicefrac{H_k^\top}{c_k^2}$ with the safe scaling $c_k^2 := \norm{H_k}_1\norm{H_k}_\infty$ of~\citet{Pan1991newton}, and $\norm{\cdot}_1$ and $\norm{\cdot}_\infty$ are the maximum absolute column and row sums.
After $P_1$ NSI recursions, we obtain the inverse-free LSTD critic of frame $k$ as $v_k := v_{k,P_1}$.

\subsection{Policy Gradient and Importance-Corrected Momentum}\label{sec:spg}
For a critic parameter $v$, the discounted sum of the one-step TD errors from step $t$ to the end of frame $k$ telescopes into the bootstrapped advantage estimate as
\begin{equation}\label{eq:bootstrapped-advantage}
	\hat\delta_{k,t}(v)
	:= \sum\nolimits_{l=0}^{T-t-1}\gamma^l\delta_{v}(o_{k,t+l})
	= \sum\nolimits_{l=0}^{T-t-1}\gamma^l r_{k,t+l}
	+ \gamma^{T-t}\phi_{k,T}^\top v - \phi_{k,t}^\top v.
\end{equation}

For an actor parameter $W$, the PG estimator on trajectory $O_k$ is
\begin{equation}\label{eq:frame-policy-gradient}
	\hat g(W,v;O_k)
	:= \frac{1}{T}\sum\nolimits_{t=0}^{T-1}\hat\delta_{k,t}(v)\,
	\nabla\log\pi_{W}(a_{k,t}| s_{k,t}).
\end{equation}

At frame $k\in[K]$, the actor evaluates $O_k$ at $W_k$ with the lagged critic $v_{k-1}$, i.e., the PG estimator of frame $k$ is $G_k := \hat g(W_k,v_{k-1};O_k)$.
The critic error enters $G_k$ in two ways: through the baseline terms $-\phi_{k,t}^\top v_{k-1}$, which have zero conditional mean because the lagged critic $v_{k-1}$ is fixed during frame $k$ and the score has zero mean given $s_{k,t}$, and through the bootstrap $\gamma^{T-t}\phi_{k,T}^\top v_{k-1}$ at the end of the frame, whose weights average to $\nicefrac{\gamma(1-\gamma^T)}{[(1-\gamma)T]}$ over the frame.

A plain momentum $M_k = \eta M_{k-1}+(1-\eta)G_k$ averages $G_1,\ldots,G_k$ at the past iterates and thus lags behind $\nabla J(W_k)$ by a drift that accumulates the changes $\nabla J(W_j)-\nabla J(W_{j-1})$ over its memory $\nicefrac{1}{(1-\eta)}$.
Following STORM~\citep{Cutkosky2019}, we cancel this drift by adding to the momentum an estimate of the change of the estimator's mean from $W_{k-1}$ to $W_k$ over the same trajectory $O_k$.
Since $O_k$ is sampled from $\pi_k$ rather than $\pi_{k-1}$, we reweight $\hat g(W_{k-1},v_{k-1};O_k)$ by the importance weight $w_k$~\citep{Huang2020} as
\begin{equation}\label{eq:importance-corrected-gradient}
	\widetilde G_k := w_k\,\hat g(W_{k-1},v_{k-1};O_k),
	\enspace
	w_k := \prod\nolimits_{t=0}^{T-1}\frac{\pi_{k-1}(a_{k,t}|s_{k,t})}{\pi_{k}(a_{k,t}|s_{k,t})}
\end{equation}
where the start state and the transition probabilities are common to both policies such that $w_k$ requires no knowledge of transition kernel $\mathbb{P}$. 
Moreover, since the actor parameter enters the PG estimator \eqref{eq:frame-policy-gradient} only via the score, the reweighted PG estimator $\widetilde G_k$ can reuse the advantage estimates of $G_k$.

The ICM with momentum factor $\eta\in[0,1)$ shifts the previous momentum by the importance correction before exponential averaging as
\begin{equation}\label{eq:is-momentum}
  M_k =
  \begin{cases}
    G_1, & k = 1 \\
    (1-\eta)G_k + \eta[ M_{k-1} + G_k - \widetilde G_k ], & k \ge 2
  \end{cases}
\end{equation}
where $\eta = 0$ induces the vanilla SGD, and dropping the importance correction term $G_k - \widetilde{G}_k$ yields the plain momentum.

\subsection{Normalized  and  Unnormalized ICM-Based Actors}\label{sec:generic-alg}

Let ${\cal F}_k := \sigma(O_0,O_1,\ldots,O_k)$ denote the history up to frame $k$ with trivial ${\cal F}_{-1}$.
Based on the ICM $M_k$, we define the two classes of actor updates as follows.

\begin{definition}[Actor updates]\label{def:steps}
	Let $\norm{\cdot}$ be a norm on $\mathbb{R}^{m\times n}$ with $\norm{\cdot}\ge\normsp{\cdot}$, and let $\normd{\cdot}$ and $\kappa_\dagger$ be its dual norm and the constant of Appendix~\ref{apdx:notation}.
	An actor update $W_{k+1} = W_k+\mathrm{Step}(M_k)$ whose output is ${\cal F}_k$-measurable is
	\begin{itemize}
		\item[(i)] \emph{normalized} in $\norm{\cdot}$, with stepsize $\beta>0$ and alignment factor $\chi\ge1$, if $\mathrm{Step}(M_k) = \beta N_k$ with $\norm{N_k}\le1$ and $\inpf{M_k,N_k}\ge\chi^{-1}\normd{M_k}$ for all $k\in[K]$;
		\item[(ii)] \emph{unnormalized}, with stepsize bound $\bar\beta>0$, range $\vartheta\ge1$, and variation $V_u\ge0$, if $\mathrm{Step}(M_k) = \beta_kM_k$ with ${\cal F}_k$-measurable $\beta_k\in[\nicefrac{\bar\beta}{\sqrt{\vartheta}},\bar\beta]$ for all $k\in[K]$ and $\sum_{k=2}^{K}|\beta_k^{-1}-\beta_{k-1}^{-1}|\le\nicefrac{V_u}{\bar\beta}$ for every realization.
	\end{itemize}
\end{definition}

Since $\normsp{N_k}\le\norm{N_k}\le1$, a normalized actor update moves by at most $\beta$ in the spectral norm irrespective of the momentum, whereas an unnormalized actor update scales with the momentum.
Note that the alignment factor is $\chi = 1$ for an exact steepest-ascent direction $N_k\in\argmax_{\norm{N}\le1}\inpf{M_k,N}$, i.e., the polar factor of $M_k$ in the spectral norm, which keeps the singular vectors of $M_k$ and sets all nonzero singular values to one, and $\nicefrac{M_k}{\normf{M_k}}$ in the Frobenius norm.

\begin{table}[ht]
	\centering\small
	\caption{Actor updates of Algorithm~\ref{alg:01} as instances of Definition~\ref{def:steps}.}
	\label{tab:steps}
	\medskip
	\setlength{\tabcolsep}{4pt}
	\begin{tabularx}{\linewidth}{@{}lll>{\raggedright\arraybackslash}X@{}}
		\toprule
		Actor & $\mathrm{Step}(M_k)$ & Class & Constants \\
		\midrule
		Muon & $\beta N_{k,P_2}$ in \eqref{eq:nso-step} & normalized in $\normsp{\cdot}$ & $\normd{\cdot} = \normnu{\cdot}$, $\kappa_\dagger = r^{1/2}$, $\chi = \nicefrac{1}{(1-r\,s_{P_2})}$ \\
		NSGDM & $\nicefrac{\beta M_k}{\normf{M_k}}$ & normalized in $\normf{\cdot}$ & $\normd{\cdot} = \normf{\cdot}$, $\kappa_\dagger = 1$, $\chi = 1$ \\
		SGDM & $\beta M_k$ & unnormalized & $\bar\beta = \beta$, $\vartheta = 1$, $V_u = 0$ \\
		AdaNorm & $\nicefrac{\beta M_k}{\sqrt{n_k}}$ in \eqref{eq:madagrad-update} & unnormalized & $\bar\beta = \beta\sqrt{\nicefrac{2}{n_0}}$, $\vartheta = 2+\nicefrac{R_g^2}{n_0}$, $V_u = \nicefrac{(\vartheta-1)}{2}$ \\
		\bottomrule
	\end{tabularx}

	\smallskip
	\begin{minipage}{0.92\linewidth}\footnotesize
	Here $r = \min\{m,n\}$, $s_{P_2}$ is given in Lemma~\ref{lemmas:nso-polar-convergence}, $R_g$ in Lemma~\ref{lemmas:critic-gradient-bounds}, and the AdaNorm constants require $\alpha\le\nicefrac{1}{(2K)}$ (Section~\ref{sec:sgdm-adanorm}).
	\end{minipage}
\end{table}

Table~\ref{tab:steps} summarizes the four actor updates that are investigated in Section~\ref{sec:instances}.
More specifically, the Muon actor update~\citep{Jordan2024} applies $P_2$ NSO recursions to the momentum as
\begin{equation}\label{eq:nso-step}
	N_{k,0} = \frac{M_k}{\normf{M_k}},
	\quad
	N_{k,p} = N_{k,p-1}\,\psi(N_{k,p-1}),\ p\in[P_2],
	\quad
	\mathrm{Step}(M_k) = \beta N_{k,P_2}
\end{equation}
where $\psi(N) := \sum_{i=0}^{\kappa_2-1}4^{-i}\binom{2i}{i}(I_n - N^\top N)^i$ is the $\kappa_2$th-order NSO polynomial with $\kappa_2\ge2$, and $N_{k,p} = 0$ for all $p$ if $M_k = 0$.
The NSGDM step $\nicefrac{\beta M_k}{\normf{M_k}}$ normalizes the momentum in the Frobenius norm~\citep{CutkoskyMehta2020} and is zero if $M_k = 0$, and the SGDM step is $\beta M_k$.
Moreover, the AdaNorm step scales the momentum by the inverse square root of a running average of the squared gradient norms as
\begin{equation}\label{eq:madagrad-update}
	n_k = (1-\alpha)n_{k-1} + \alpha\normf{G_k}^{2},
	\quad
	\mathrm{Step}(M_k) = \frac{\beta}{\sqrt{n_k}} M_k
\end{equation}
where $n_0>0$ and $\alpha\in[0,1]$.

\begin{algorithm}[ht]
	\footnotesize
	\caption{Generic ICM-AC over a Single Markovian Trajectory}
	\label{alg:01}
	\begin{algorithmic}[1]
		\State \textbf{Initialize:} actor $W_1$, frame length $T\ge1$, ridge parameter $\lambda>0$, number $P_1$ and order $\kappa_1\ge2$ of NSI recursions, momentum factor $\eta\in[0,1)$, and actor update $\mathrm{Step}(\cdot)$ (Definition~\ref{def:steps})
		\State Roll out the warm-up frame $O_0$ of $T$ transitions via $\pi_1$ from $s_{0,0}\sim\rho$
		\State Run $P_1$ NSI recursions \eqref{eq:newton-schulz-critic} on $O_0$ and set the critic $v_0 := v_{0,P_1}$
		\For{$k = 1, 2, \ldots, K$}
		\State Roll out the frame $O_k$ of $T$ transitions via $\pi_k$ from $s_{k,0} = s_{k-1,T}$
		\State Compute $G_k = \hat g(W_k,v_{k-1};O_k)$ by \eqref{eq:frame-policy-gradient}
		\If{$k = 1$}
		\State Set $M_1 = G_1$
		\Else
		\State Compute $w_k$ and $\widetilde G_k$ by \eqref{eq:importance-corrected-gradient} and $M_k$ by \eqref{eq:is-momentum}
		\EndIf
		\State Update the actor $W_{k+1} = W_k+\mathrm{Step}(M_k)$
		\State Run $P_1$ NSI recursions \eqref{eq:newton-schulz-critic} on $O_k$ and set the critic $v_k := v_{k,P_1}$
		\EndFor
		\State \Return critic $v_K$ and actor $W_U$ with $U$ drawn uniformly from $[K]$ (in practice, $W_{K+1}$)
	\end{algorithmic}
\end{algorithm}

Based on the frames of a single trajectory, Algorithm~\ref{alg:01} runs the inverse-free LSTD critic \eqref{eq:newton-schulz-critic}, the PG estimator \eqref{eq:frame-policy-gradient}, and the ICM \eqref{eq:is-momentum}, and leaves the actor update $\mathrm{Step}(M_k)$ open.
The critic and the actor share the same sample trajectory such that $K$ updates consume $(K+1)T$ transitions.
Reservoir sampling implements the uniform selection of the returned iterate $W_U$ without storing the full sequence of iterates.
Since $\ee{\normd{\nabla J(W_U)}} = \frac{1}{K}\sum_{k=1}^{K}\ee{\normd{\nabla J(W_k)}}$, the bounds on the averaged gradient norm in Section~\ref{sec:analysis} also hold for the expected gradient norm at $W_U$.

\section{Generic Finite-Time Analysis}\label{sec:analysis}
In this section, we establish the finite-time analysis of Algorithm~\ref{alg:01} when all frames are consecutive segments of a single continuing trajectory such that the observations are correlated within and across frames.
More specifically, Sections~\ref{sec:critic-analysis}--\ref{sec:generic-bounds} bound the critic error, split the bias of the PG estimator, bound the tracking error of the ICM, and turn the tracking error into the convergence rates of the actor, respectively.
Note that the actor parameters $W_k$ and $W_{k-1}$, the lagged critic $v_{k-1}$, and the start state $s_{k,0}$ are ${\cal F}_{k-1}$-measurable; hence, conditionally on ${\cal F}_{k-1}$, $O_k$ is a $T$-step trajectory of $\pi_k$ from $s_{k,0}$.

\begin{assump}\label{assum:01}
	For each state-action pair $(s,a)  \in {\cal S}\times{\cal A}$, the policy $\pi_W$ has a density $\pi_W(a|s) > 0$ that is differentiable with respect to the actor parameter $W$.
	Moreover, the score function $\nabla\log\pi_{W}$ satisfies
	\begin{subequations}\label{eq:policy-regularity}
		\begin{align}
			&\normnu{ \nabla\log\pi_W(a|s) } \le R_\pi \label{eq:score-bound}\\
			&\normnu{ \nabla\log\pi_W(a|s) \!-\! \nabla\log\pi_{W'}(a|s) } \!\le\! L_\pi\normsp{W - W'} \!\!\label{eq:score-lipschitz}
		\end{align}
	\end{subequations}
	where $W$ and $W'$ are arbitrary actor parameters.
\end{assump}

With the Frobenius norm in place of the nuclear and spectral norms in \eqref{eq:policy-regularity}, Assumption~\ref{assum:01} is standard in PG analyses~\citep[see, e.g.,][]{Zou2019,Hong2023,Wang2020,ChenDec.2023,Wu2020,Shen2022,Olshevsky2023,Kumar2023,Bedi2022,Ganesh2025} and holds for, e.g., Cauchy, Boltzmann, and truncated Gaussian policies~\citep{Konda1999a, Bedi2022, Dong2026}.
Assumption~\ref{assum:01} guarantees that $\nabla J$ is Lipschitz continuous with proof relegated to Appendix~\ref{apdx:smoothness}.

\begin{lemma}\label{lemmas:return-smoothness}
	Under Assumption~\ref{assum:01}, the policy gradient $\nabla J$ is Lipschitz continuous  as
	\begin{equation}\label{eq:return-smoothness}
		\normnu{\nabla J(W)-\nabla J(W')}\le L\normsp{W-W'}
	\end{equation}
	where $W$ and $W'$ are arbitrary actor parameters with $L := \nicefrac{R_rL_\pi}{(1-\gamma)}+\nicefrac{(1+\gamma)R_rR_\pi^2}{(1-\gamma)^2}$.
\end{lemma}

Based on Lemma \ref{lemmas:return-smoothness}, applying the descent lemma~\citep{Nesterov2018} to $-J$, we obtain
\begin{equation}\label{eq:return-ascent-bound}
	\inpf{ \nabla J(W), W' - W }
	\le J(W')  - J(W) + \frac{1}{2}L\normsp{W' - W}^2.
\end{equation}

Note that \eqref{eq:return-ascent-bound} is the smoothness with respect to the spectral norm in the analyses of Muon, e.g.,~\citet{kim2026convergence}.
Since $\normsp{\cdot}\le\normf{\cdot}$, \eqref{eq:return-ascent-bound} also holds with $\nicefrac{L}{2}\normf{W'-W}^2$ in place of $\nicefrac{L}{2}\normsp{W'-W}^2$.

\begin{assump}\label{assum:is}
	The policy-induced chains mix uniformly (i.e., the mixing time is finite) as
	\begin{equation*}
		\tau_{\mix} := \sup\nolimits_W\min\big\{t\ge1:\sup\nolimits_{s}\normtv{\pp{s_t\in\cdot\,|\,s_0 = s,\pi_W}-\mu_W}\le\nicefrac{1}{4}\big\}
	\end{equation*}
	where $\mu_W$ is a stationary distribution of the chain induced by $\pi_W$.
\end{assump}

The uniform ergodicity in Assumption~\ref{assum:is} is standard in finite-time analyses of TD learning and AC methods under Markovian sampling~\citep{Bhandari2018,Zou2019,Wu2020}.
Since the mixing time is uniform in the start state, the bounds for a frame hold conditionally on the past although each frame starts from the last state of the previous frame.
Note that the mixing time is unknown to Algorithm~\ref{alg:01} and enters only the analysis.
Assumptions~\ref{assum:01} and~\ref{assum:is} also guarantee that the solution of \eqref{eq:population-lstd-system} is Lipschitz continuous in the actor parameter with proof relegated to Appendix~\ref{apdx:lemma:opt_lip}.

\begin{lemma}\label{lemmas:critic-target-lipschitz}
	Under Assumptions~\ref{assum:01} and~\ref{assum:is}, the solution $v^*(W) := (A_W+\lambda I_d)^{-1}b_W$ of \eqref{eq:population-lstd-system} under $\pi_W$ is Lipschitz continuous in the actor parameter as
	\begin{equation}\label{eq:critic-target-lipschitz}
		\normsp{v^*(W)-v^*(W')}\le L_*\normsp{W-W'}
	\end{equation}
	where $W$ and $W'$ are arbitrary actor parameters with $L_* := \nicefrac{(4\tau_{\mix}+1)R_\pi R_r(\lambda+1+\gamma)}{\lambda^2}$.
\end{lemma}

\subsection{Critic Error and Target Drift}\label{sec:critic-analysis}
For frame $\smash[t]{k\ge1}$, let $\smash[t]{A_k}$ and $\smash[t]{b_k}$ denote the population quantities in \eqref{eq:population-lstd-system} under $\smash[t]{ \mu_k \otimes \pi_{k} \otimes \mathbb{P} }$ with $\mu_k := \mu_{W_k}$, let $\smash[t]{v_k^* := v^*(W_k) = (A_k+\lambda I_d)^{-1}b_k}$, and let $v_0^* := v_1^*$ for the warm-up frame.
By leveraging a second-moment bound for averages along the policy-induced chain (Lemma~\ref{lemmas:markov-moment} in Appendix~\ref{apdx:prelim}; cf.~\citealp[Lemma~A.6]{DorfmanJuly2022}), we establish the error bound of the inverse-free LSTD critic in the following theorem with proof relegated to Appendix~\ref{apdx:thm:01}.

\begin{theorem}[Inverse-Free LSTD critic]\label{theorems:lstd-critic-error}
	Under Assumptions~\ref{assum:01} and~\ref{assum:is}, choose any ridge parameter $\lambda>0$, which lower-bounds $\sigma_{\min}(\hat A_k+\lambda I_d)$ (Appendix~\ref{apdx:bk}), and initialize NSI with $X_{k,0} = \nicefrac{H_k^\top}{c_k^2}$ for any $c_k\in[\normsp{H_k},\lambda+1+\gamma]$.
	Then, for every $k\ge0$, after $P_1$ NSI recursions \eqref{eq:newton-schulz-critic}, the critic $v_k := v_{k,P_1}$ satisfies $\normsp{v_k}\le R_v$ and
	\begin{equation}\label{eq:lstd-critic-error}
		\sqrt{\ee{ \normsp{v_{k} - v_k^*}^2\,|\,{\cal F}_{k-1} }}
		\le \epsilon_{\rm lstd} := R_v\exp(-C_E\kappa_1^{P_1}) + 2C_v\sqrt{\frac{1+4\tau_{\mix}}{T}}
	\end{equation}
	where $C_E := \log\nicefrac{1}{\rho_E}>0$, and
	\begin{gather*}
		R_v = \frac{R_r}{\lambda},
		\enspace
		C_v := \frac{\sqrt2 R_r(\lambda+1+\nicefrac{3\gamma}{2})}{\lambda^2},
		\enspace
		\rho_E := 1-\Big(\frac{\lambda}{\lambda+1+\gamma}\Big)^4\in(0,1).
	\end{gather*}
	In particular, $\ee{\normsp{v_k-v_k^*}^2\,|\,{\cal F}_{k-1}}\le2R_v^2\exp(-2C_E\kappa_1^{P_1})+\nicefrac{8(1+4\tau_{\mix})C_v^2}{T}$.
\end{theorem}

The two terms in \eqref{eq:lstd-critic-error} are the NSI error and the Markovian sampling error, respectively.
Since the residual satisfies $\normsp{E_{k,q+1}} = \normsp{E_{k,q}}^{\kappa_1}$, the NSI error is at most $\delta<R_v$ after $P_1 = \max\{0,\lceil\log_{\kappa_1}(\nicefrac{\log(\nicefrac{R_v}{\delta})}{C_E})\rceil\}$ NSI recursions, which is doubly logarithmic in $\nicefrac{R_v}{\delta}$ and logarithmic in the conditioning of $H_k$.
Moreover, the realized residual bounds the NSI error by $R_v\normf{E_{k,P_1}}$ and can be monitored during training.
It is worth stressing that the bounds hold for every ridge parameter $\lambda>0$ with the constants $R_v$, $C_v$, and $\rho_E$ independent of the frame length $T$ due to the two facts: (i) the ridge-regularized pathwise LSTD system is accretive for every frame length (Appendix~\ref{apdx:bk}); and (ii) dropping the bootstrap of the last transition leaves the target $v_k^*$ unchanged and makes $\hat A_k$ differ from the full average of $\phi(s)(\phi(s)-\gamma\phi(s'))^\top$ over the frame by at most $\nicefrac{\gamma}{T}$ in the Frobenius norm, which is absorbed by $C_v$.
Since the lagged critic $v_{k-1}$ used at frame $k$ is fitted to $\pi_{k-1}$, Lemma~\ref{lemmas:critic-target-lipschitz} bounds the resulting policy drift of the lagged critic by $L_*\normsp{W_k-W_{k-1}}$.

\subsection{Bias, Noise, and Importance Correction of the PG Estimator}\label{sec:spg-analysis}
Throughout the actor analysis, the ridge parameter and the NSI initialization satisfy the conditions of Theorem~\ref{theorems:lstd-critic-error}.
For a $T$-step trajectory $O = [o_t]_{t=0}^{T-1}$ with $o_t = (s_t,a_t,s_{t+1})$ that is induced by the policy $\pi_W$ from $s_0$, we write $\hat\delta_t(v)$ and $\hat g(W,v;O)$ for \eqref{eq:bootstrapped-advantage} and \eqref{eq:frame-policy-gradient} evaluated on $O$ in place of $O_k$.
The following lemma bounds the critic, the advantage estimates, and the PG estimators for every realization with proof relegated to Appendix~\ref{apdx:lemma:support}.

\begin{lemma}\label{lemmas:critic-gradient-bounds}
	Under Assumption~\ref{assum:01} and the conditions of Theorem~\ref{theorems:lstd-critic-error}, call a critic parameter $v$ admissible if $\sup_s|\phi^\top(s)v|\le R_v$, and let $R_A := \nicefrac{[R_r+(1+\gamma)R_v]}{(1-\gamma)}$ and $R_g := R_AR_\pi$.
	Then, the following hold for every realization.
	\begin{itemize}
		\item[(i)] Every NSI iterate satisfies $\normsp{v_{k,q}}\le R_v$; hence, since $\normsp{\phi(s)}\le1$, the critic $v_k = v_{k,P_1}$ is admissible.
		\item[(ii)] For every admissible $v$, every observation $o$, and every step $t$ of a $T$-step frame, the TD error and the advantage estimate \eqref{eq:bootstrapped-advantage} satisfy $|\delta_v(o)|\le R_r+(1+\gamma)R_v$ and $|\hat\delta_t(v)|\le R_A$.
		\item[(iii)] For every actor parameter $W$, every admissible $v$, and every $T$-step frame $O$, the PG estimator \eqref{eq:frame-policy-gradient} and the PG satisfy $\normnu{\hat g(W,v;O)}\le R_g$ and $\normnu{\nabla J(W)}\le\nicefrac{R_rR_\pi}{(1-\gamma)}\le R_g$; in particular, $\normnu{G_k}\le R_g$ for every $k$, and, since $\normf{\cdot}\le\normnu{\cdot}$, the same bounds hold in the Frobenius norm.
	\end{itemize}
\end{lemma}

For an admissible critic parameter $v$ and a start state $s_0$, we call $\omega = (v,s_0)$ an admissible frame context.
Since the uniform weights in \eqref{eq:frame-policy-gradient} average over the observations whose state distribution approaches $\mu_W$ rather than $\zeta_W$, we compare the PG estimator with the stationary counterpart $g(W)$ of \eqref{eq:policy-gradient} and split its bias into a context-free part $\bar B$ and a context-dependent part $D$ as
\begin{equation}\label{eq:actor-mean-field}
\begin{split}
g(W) &:= \mathbb{E}_{(s,a)\sim\mu_W\otimes\pi_W}\big[(Q_W(s,a) - V_W(s))\nabla\log\pi_W(a|s)\big],  \\
h(W;\omega) &:= \ee{\hat g(W,v;O) | s_0},
\qquad
D(W;\omega) := h(W;\omega)-g(W), \\
B(W;\omega) &:= h(W;\omega) - \nabla J(W),
\qquad
\bar B(W) := g(W)-\nabla J(W)
\end{split}
\end{equation}
where $\mu_W$ is unique by Assumption~\ref{assum:is}, the context-free part $\bar B$ is the mismatch between $\mu_W$ and $\zeta_W$, and the context-dependent part $D$ collects the transient from $s_0$ and the critic error.
By the ${\cal F}_{k-1}$-measurability in Section~\ref{sec:analysis}, the actor at frame $k$ operates under the admissible and ${\cal F}_{k-1}$-measurable context $\omega_k := (v_{k-1},s_{k,0})$ such that $\ee{G_k\,|\,{\cal F}_{k-1}} = h(W_k;\omega_k)$.
The critic enters the PG estimator through its value error at the current policy as
\begin{equation}\label{eq:critic-mismatch-definitions}
		\epsilon_k(s) := \phi^\top(s)v_{k-1} - V_{W_k}(s),
		\enspace
		\epsilon_{\rm crit} := \sup_{k\in[K]}\sqrt{\ee{\norm{\epsilon_k}_\infty^2}},
		\enspace
		\epsilon_{\rm dist} := \sup_W\normnu{\bar B(W)}
\end{equation}
and we let $G_k^\circ$ denote the oracle estimator, i.e., \eqref{eq:frame-policy-gradient} evaluated at $W_k$ and $O_k$ with $V_{W_k}$ in place of $\phi^\top v$ in \eqref{eq:bootstrapped-advantage}.
Moreover, we define the constants
\begin{equation}\label{eq:actor-error-constants}
	\begin{split}
		&C_F := R_A\big[L_\pi + (\exp(1)-1)\,TR_\pi^2\big],
		\quad
		D_F := \frac{\tau_{\mix}^2R_A(46R_\pi^2+5L_\pi)}{(1-\gamma)^2T}, \\
		&\epsilon_{\rm bias} := \epsilon_{\rm dist} + \frac{R_\pi(4R_r\tau_{\mix}+\gamma\,\epsilon_{\rm crit})}{(1-\gamma)T}, \\
		&V^\circ := \frac{R_rR_\pi}{1-\gamma}\Big(4\sqrt{1+4\tau_{\mix}}+\frac{\sqrt{2}\,\gamma}{1-\gamma}\Big),
		\quad
		\bar\sigma := \frac{V^\circ+R_\pi\epsilon_{\rm crit}}{\sqrt{T}}+\frac{\gamma R_\pi\epsilon_{\rm crit}}{(1-\gamma)T}.
	\end{split}
\end{equation}

Note that the bounds $\epsilon_{\rm crit}\le R_v+\nicefrac{R_r}{(1-\gamma)}$ and $\epsilon_{\rm dist}\le\nicefrac{2R_rR_\pi}{(1-\gamma)}\sup_W\normtv{\mu_W-\zeta_W}$ hold such that every constant in \eqref{eq:actor-error-constants} is finite and free of $r$.
Besides, the nuclear quantities $\epsilon_{\rm dist}$ and $\epsilon_{\rm bias}$ also bound the bias in the Frobenius norm, whereas $\bar\sigma$ bounds the noise in the Frobenius norm in which the ICM averages the noise.
Since \eqref{eq:score-bound} and \eqref{eq:score-lipschitz} pair the score with the spectral norm, the constants $C_F$ and $D_F$ bound the importance correction and the change of the frame bias per unit spectral norm of the step in the nuclear norm, respectively.
The following two lemmas are established with proofs relegated to Appendices~\ref{apdx:lemma:is_corr} and~\ref{apdx:lemma:is_bias}, respectively.

\begin{lemma}\label{lemmas:importance-correction}
	Under Assumption~\ref{assum:01}, the following conditions hold.
	\begin{itemize}
		\item[(i)] For actor parameters $W$ and $W'$ with $\ell_2 := \normsp{W-W'}$ and a trajectory $O$ of $T$ steps with $TR_\pi\ell_2\le1$, the importance weight $w := \prod_{t=0}^{T-1}\nicefrac{\pi_{W'}(a_t|s_t)}{\pi_W(a_t|s_t)}$ satisfies $|\log w|\le TR_\pi\ell_2\le1$ and $|w-1|\le(\exp(1)-1)TR_\pi\ell_2$.
		\item[(ii)] For every $k\ge2$, the reweighted estimate is conditionally unbiased, i.e., $\ee{\widetilde G_k\,|\,{\cal F}_{k-1}} = h(W_{k-1};\omega_k)$.
		\item[(iii)] If $TR_\pi\ell_2\le1$, then, for every admissible $v$ and every $T$-step frame $O$, with the importance weight $w$ of part (i), the importance correction satisfies
		\begin{equation*}
			\normnu{\hat g(W,v;O)-w\,\hat g(W',v;O)}\le C_F\,\ell_2;
		\end{equation*}
		since $\normf{\cdot}\le\normnu{\cdot}$, the same bound holds in the Frobenius norm.
	\end{itemize}
\end{lemma}

When $\normsp{W_k-W_{k-1}}\le\bar\ell_2$ with $TR_\pi\bar\ell_2\le1$, parts (i) and (iii) in Lemma \ref{lemmas:importance-correction} with $W = W_k$, $W' = W_{k-1}$, $O = O_k$, and $v = v_{k-1}$ induce $|\log w_k|\le1$ and $\normnu{G_k-\widetilde G_k}\le C_F\,\bar\ell_2$ for every $k\ge2$.
Since a normalized actor update satisfies $\normsp{W_k-W_{k-1}}\le\beta$, Lemma~\ref{lemmas:importance-correction} bounds its importance weights deterministically when $TR_\pi\beta\le1$ without the coupling between $\beta$ and $\eta$, the weight clipping, or the bounded variance of the importance weights that~\citet[Assumption~3]{Huang2020} impose for Euclidean steps; for softmax policies, normalized steps remove this assumption in the same way~\citep{Barakat2023general}.
Since an unnormalized actor update scales with $M_k$, its importance weights are controlled only when the momentum stays bounded for every realization (Lemma~\ref{lemmas:baseline-momentum-tracking} in Appendix~\ref{apdx:lemma:is_track}).
Moreover, the importance weight spans the whole frame such that the step condition involves $T$ and $C_F$ grows linearly with $T$, as the corresponding constant of~\citet{Huang2020} grows with the horizon; an analysis of the martingale structure of $\log w_k$ might reduce such growth.
Note that the per-decision weights $\smash[t]{ \prod_{i=0}^{t+l}\nicefrac{\pi_{k-1}(a_{k,i}|s_{k,i})}{\pi_k(a_{k,i}|s_{k,i})} }$ on the summand $\delta_v(o_{k,t+l})\nabla\log\pi_{k-1}(a_{k,t}|s_{k,t})$ keep $\widetilde G_k$ conditionally unbiased, satisfy the same bounds, and can reduce the variance of $\widetilde G_k$~\citep{Precup2000}.

\begin{lemma}\label{lemmas:frame-bias-noise}
	Under Assumptions~\ref{assum:01} and~\ref{assum:is}, consider an actor update whose iterates $W_k$ are ${\cal F}_{k-1}$-measurable, the following condition hold for every $k\in[K]$.
	\begin{itemize}
		\item[(i)] For every admissible context $\omega = (v,s_0)$ and every $W$, the context-dependent part of the bias satisfies $\normnu{D(W;\omega)}\le\nicefrac{R_\pi(4R_r\tau_{\mix}+\gamma\norm{\epsilon_v}_\infty)}{[(1-\gamma)T]}$ with $\epsilon_v := \phi^\top v-V_W$; in particular, $\sqrt{\ee{\normnu{B(W_k;\omega_k)}^2}}\le\epsilon_{\rm bias}$.
		\item[(ii)] For any two actor parameters $W$ and $W'$ and every admissible context $\omega$, the context-dependent part of the bias satisfies $\normnu{D(W;\omega)-D(W';\omega)}\le D_F\normsp{W-W'}$.
		\item[(iii)] The noise of the PG estimator satisfies $\sqrt{\ee{\normf{G_k-\ee{G_k\,|\,{\cal F}_{k-1}}}^2}}\le\bar\sigma$, and that of the oracle estimator satisfies $\sqrt{\ee{\normf{G_k^\circ-\ee{G_k^\circ\,|\,{\cal F}_{k-1}}}^2}}\le\nicefrac{V^\circ}{\sqrt{T}}$.
		\item[(iv)] When $\normsp{W_k-W_{k-1}}\le\bar\ell_2$ for all $k\ge2$, the critic error satisfies
		\begin{equation*}
			\epsilon_{\rm crit}\le\epsilon_{\rm lstd}+L_*\bar\ell_2+\epsilon_{\rm app}
		\end{equation*}
		where  $\bar\ell_2\ge0$ is a positive constant, $\epsilon_{\rm lstd}$ is given in Theorem~\ref{theorems:lstd-critic-error}, $L_*$ in Lemma~\ref{lemmas:critic-target-lipschitz}, and the approximation error of the linear critic at the solution $v^*(W)$ is
		\begin{equation*}
			\epsilon_{\rm app} := \sup\nolimits_{W,s}|\phi^\top(s)v^*(W)-V_W(s)|.
		\end{equation*}
	\end{itemize}
\end{lemma}

Part (i) of Lemma~\ref{lemmas:frame-bias-noise} shows that, besides $\epsilon_{\rm dist}$, the bias contains a frame bias of order $\nicefrac{1}{T}$ with two sources: the start-state transient, since the average over the frame forgets its start state after $\oo{\tau_{\mix}}$ steps; and the critic error at the end of the frame, which is attenuated by $\nicefrac{\gamma}{[(1-\gamma)T]}$ since the lagged critic enters only through the bootstrap (Section~\ref{sec:spg}).
Since the exact bootstrap weight $\nicefrac{\gamma(1-\gamma^T)}{[(1-\gamma)T]}$ is at most $\min\{\gamma,\nicefrac{\gamma}{[(1-\gamma)T]}\}$, every frame length $T>\nicefrac{1}{(1-\gamma)}$ improves on the weight $\gamma$ of a one-step TD error.
Part (ii) shows that $D$ varies slowly with the actor parameter since $D$ involves only the first $\oo{\tau_{\mix}}$ and the last $\oo{\nicefrac{1}{(1-\gamma)}}$ steps of the frame, each with weight $\nicefrac{1}{T}$; different from the change of measure over the whole frame behind $C_F$, the proof perturbs the policy-induced chain in the spirit of~\citet{Mitrophanov2005}.
Part (iii) quantifies the noise: the oracle part follows from Lemma~\ref{lemmas:markov-moment} and a martingale decomposition of the multi-step TD errors (Lemma~\ref{lemmas:oracle-noise} in Appendix~\ref{apdx:lemma:is_bias}), and the critic part averages the baseline terms over the frame without the factor $r^{1/2}$ that a nuclear-norm bound would cost.
Both parts decay as $T^{-1/2}$, and $\bar\sigma$ enters the bounds of Section~\ref{sec:generic-bounds} only through their transients.
Part (iv) carries Theorem~\ref{theorems:lstd-critic-error} into the actor analysis, whereas parts (i)--(iii) use the actor update only through the ${\cal F}_{k-1}$-measurability of $W_k$.

\subsection{Momentum Tracking}\label{sec:tracking}
The following lemma bounds the tracking error of the ICM for every actor update of bounded spectral norm and in every dual norm of Appendix~\ref{apdx:notation} with proof relegated to Appendix~\ref{apdx:lemma:is_track}.

\begin{lemma}\label{lemmas:momentum-tracking}
	Under Assumptions~\ref{assum:01} and~\ref{assum:is}, consider Algorithm~\ref{alg:01} with $\eta\in[0,1)$ and an actor update whose iterate $W_k$ is ${\cal F}_{k-1}$-measurable and satisfies $\normsp{W_k-W_{k-1}}\le\bar\ell_2$ for all $k\ge2$, where $\bar\ell_2\ge0$ is deterministic and, if $\eta>0$, $TR_\pi\bar\ell_2\le1$, and let $\bar\sigma$ and $\epsilon_{\rm bias}$ be evaluated along its iterates.
	Then, for all $k\in[K]$ and every norm $\norm{\cdot}\ge\normsp{\cdot}$ with dual norm $\normd{\cdot}$, the ICM \eqref{eq:is-momentum} tracks the PG as
	\begin{equation}\label{eq:momentum-tracking-generic}
		\ee{\normd{M_k-\nabla J(W_k)}}
		\le \kappa_\dagger\Big[\big(\eta^{k-1}+\sqrt{2(1-\eta)}\big)\,\bar\sigma + \frac{\sqrt{2}\,\eta\,C_F\bar\ell_2}{\sqrt{1-\eta}}\Big]
		+ \epsilon_{\rm bias} + \frac{\eta D_F\bar\ell_2}{1-\eta}
	\end{equation}
	and
	\begin{equation}\label{eq:momentum-tracking-second-moment}
		\ee{\normf{M_k-\nabla J(W_k)}^2}
		\le 2\eta^{2(k-1)}\bar\sigma^2+4(1-\eta)\,\bar\sigma^2+\frac{4\eta^2C_F^2\bar\ell_2^2}{1-\eta}
		+2\sq{ \epsilon_{\rm bias}+\frac{\eta D_F\bar\ell_2}{1-\eta} }^2.
	\end{equation}
\end{lemma}

By the proof of Lemma~\ref{lemmas:momentum-tracking} in Appendix~\ref{apdx:lemma:is_track}, the tracking error splits as $M_k-\nabla J(W_k) = e^{\rm n}_k+e^{\rm b}_k$, where the noise part $e^{\rm n}_k$, i.e., a weighted sum of martingale differences, satisfies $\ee{\normf{e^{\rm n}_k}}\le(\eta^{k-1}+\sqrt{2(1-\eta)})\,\bar\sigma+\nicefrac{\sqrt{2}\,\eta C_F\bar\ell_2}{\sqrt{1-\eta}}$, and the bias part $e^{\rm b}_k$, i.e., a weighted sum of conditional means, satisfies $\ee{\normnu{e^{\rm b}_k}}\le\epsilon_{\rm bias}+\nicefrac{\eta D_F\bar\ell_2}{(1-\eta)}$.
Hence, the ICM averages the noise $\bar\sigma$, including the critic noise, over about $\nicefrac{1}{(1-\eta)}$ frames at the price of the correction noise $\nicefrac{\sqrt{2}\eta C_F\bar\ell_2}{\sqrt{1-\eta}}$, which is small since the importance correction and the bounded step make $G_k-\widetilde G_k$ of the order of $\bar\ell_2$.
Since such averaging rests on the orthogonality of martingale differences in the Frobenius inner product, the noise part is converted to the dual norm at the cost of $\kappa_\dagger$, whereas the bias part carries no such factor since Lemma~\ref{lemmas:frame-bias-noise}(i),(ii) hold in the nuclear norm.
Moreover, since the importance correction makes the ICM track the mean field $h$, the context-free part $\bar B$ telescopes in the bias recursion \eqref{eq:context-free-bias-telescope} such that the drift stems from $D$ alone.
Note that the second-moment bound \eqref{eq:momentum-tracking-second-moment} is needed only for unnormalized actor updates.

\begin{remark}\label{rem:is_plain}
	For a normalized actor update with stepsize $\beta$, the plain momentum $M_k = \eta M_{k-1}+(1-\eta)G_k$ with $M_1 = G_1$ satisfies $\ee{\normd{M_k-\nabla J(W_k)}}\le\kappa_\dagger(\eta^{k-1}+\sqrt{1-\eta})\,\bar\sigma+\epsilon_{\rm bias}+\nicefrac{\eta L\beta}{(1-\eta)}$ for any $\beta>0$ (Appendix~\ref{apdx:lemma:is_track}), i.e., \eqref{eq:momentum-tracking-generic} with $C_F = 0$, $\bar\ell_2 = \beta$, and $L$ in place of $D_F$, since its drift accumulates the changes of $\nabla J$ whose Lipschitz constant does not decay with $T$.
	Hence, the plain momentum retains the floor term $(1+\chi)cL$ under the schedule of Theorem~\ref{theorems:normalized-actor}.
	Removing such a floor term requires $\nicefrac{\beta}{(1-\eta)}\to0$, and balancing the terms under this constraint gives $1-\eta = K^{-1/2}$, $\beta = cK^{-3/4}$, and only the rate $\oo{K^{-1/4}}$.
	Therefore, the importance correction improves the rate to $\oo{K^{-1/3}}$ and replaces $(1+\chi)cL$ with the $\oo{\nicefrac{1}{T}}$ term $(1+\chi)cD_F$ at the price of the correction noise $(1+\chi)\sqrt{2}\,\kappa_\dagger cC_FK^{-1/3}$.
\end{remark}

\begin{remark}[Why the bias is split]\label{rem:bias}
	By Appendix~\ref{apdx:lemma:is_track}, the bias part of the tracking error is $e_k^{\rm b} = \eta^{k-1}B_1+\sum_{j=2}^{k}\eta^{k-j}[(1-\eta)B_j+\eta(B(W_j;\omega_j)-B(W_{j-1};\omega_j))]$ with $B_j := B(W_j;\omega_j)$, where $B(W_{j-1};\omega_j)$ carries the current context since $G_j$ and $\widetilde G_j$ share a frame whose start state is not changed by the importance weight.
	The split of the bias is justified by the three facts.
	(i) A bound $\normnu{B}\le b$ alone, as in the analyses of SGD with biased gradients~\citep{Ajalloeian2020biased,Demidovich2023zoo}, bounds the accumulated differences only by $\nicefrac{2\eta b}{(1-\eta)}$, which diverges under the schedule of Theorem~\ref{theorems:normalized-actor}; moreover, a Lipschitz constant $L_B$ of $B$ in $W$ leaves the floor term $(1+\chi)cL_B$ that does not decay with $T$ since the Lipschitz constant of $\bar B$, like that of $\nabla J$, does not decay with $T$.
	(ii) Tracking $h(W_k;\omega_k)$ instead of $\nabla J(W_k)$, as~\citet{Khanduri2021near} do for a mean field that depends on the iterates alone, introduces the change of the context between frames, which is of order $\nicefrac{1}{T}$ rather than $\beta$, such that the separate bounds grow as $\nicefrac{K^{2/3}}{T}$.
	(iii) The split $B = \bar B+D$ combines both views: $\bar B$ telescopes to $\bar B(W_k)$ such that the mismatch enters only once, and $D$ enters only through its size of order $\nicefrac{1}{T}$ and its change at a fixed context of order $D_F\bar\ell_2$.
\end{remark}

\subsection{Convergence Rates of Generic ICM-Based Actors}\label{sec:generic-bounds}
Based on Lemmas~\ref{lemmas:importance-correction}--\ref{lemmas:momentum-tracking}, we establish the convergence rates of the two classes of actor updates in Definition~\ref{def:steps} with proofs relegated to Appendices~\ref{apdx:thm:normalized} and~\ref{apdx:thm:unnormalized}, respectively.
Let $\Delta_1 := J^*-J(W_1)$ with $J^* := \sup_W J(W)$.

\begin{theorem}[Normalized actor updates]\label{theorems:normalized-actor}
	Under Assumptions~\ref{assum:01} and~\ref{assum:is}, let the actor update be normalized in a norm $\norm{\cdot}\ge\normsp{\cdot}$ with stepsize $\beta>0$ and alignment factor $\chi$ (Definition~\ref{def:steps}), let $\eta\in[0,1)$ with $TR_\pi\beta\le1$ if $\eta>0$, and let $\bar\sigma$ and $\epsilon_{\rm bias}$ be evaluated along the iterates.
	Then Algorithm~\ref{alg:01} satisfies
	\begin{equation}\label{eq:normalized-bound}
		\begin{split}
			&\frac{1}{K}\sum_{k=1}^{K}\ee{\normd{\nabla J(W_k)}}
			\le \chi\Big[\frac{\Delta_1}{\beta K} + \frac{L\beta}{2}\Big] \\
			&\quad+ (1+\chi)\,\kappa_\dagger\Big[\frac{\bar\sigma}{K(1-\eta)} + \sqrt{2(1-\eta)}\,\bar\sigma + \frac{\sqrt{2}\,\eta\,C_F\beta}{\sqrt{1-\eta}}\Big]
			+ (1+\chi)\Big(\epsilon_{\rm bias} + \frac{\eta D_F\beta}{1-\eta}\Big).
		\end{split}
	\end{equation}
	In particular, for any $c>0$ and a prescribed horizon $K\ge(cTR_\pi)^{3/2}$, the choice $1-\eta = K^{-2/3}$ and $\beta = cK^{-2/3}$ induces
	\begin{equation}\label{eq:normalized-rate}
		\begin{split}
			&\frac{1}{K}\sum_{k=1}^{K}\ee{\normd{\nabla J(W_k)}}
			\le \chi\Big[\frac{\Delta_1}{c}\,\frac{1}{K^{1/3}} + \frac{cL}{2 K^{2/3}}\Big] \\
			&\quad+ (1+\chi)\,\kappa_\dagger\big[(1+\sqrt{2})\,\bar\sigma + \sqrt{2}\,cC_F\big]\frac{1}{K^{1/3}}
			+ (1+\chi)\big(\epsilon_{\rm bias}+cD_F\big).
		\end{split}
	\end{equation}
\end{theorem}

The proof uses the actor update only through the ${\cal F}_k$-measurability, the step bound $\normsp{W_{k+1}-W_k}\le\beta$, and the alignment, which turn \eqref{eq:return-ascent-bound} into the inequality $\normd{\nabla J(W_k)}\le\chi\nicefrac{[J(W_{k+1})-J(W_k)]}{\beta}+(1+\chi)\normd{M_k-\nabla J(W_k)}+\nicefrac{\chi L\beta}{2}$ that is linear in the tracking error.
The exponent of the $\oo{K^{-1/3}}$ transient matches those of the ICM actor~\citep{Huang2020} and the variance-reduced Muon~\citep{Chang2025muon}, which rely on unbiased estimates from restarted episodes or i.i.d.\ samples, whereas \eqref{eq:normalized-rate} holds for biased and Markov-correlated frame estimates (Remark~\ref{rem:scope}).
Note that the noise $\bar\sigma$ does not enter the floor since the ICM averages the noise across frames.
For $\eta = 0$, the bound retains $\sqrt{2}(1+\chi)\kappa_\dagger\bar\sigma = \oo{T^{-1/2}}$ for every stepsize since normalizing a noisy estimate does not average its noise; hence, only longer frames reduce the noise (Remark~\ref{rem:complexity}).

\begin{theorem}[Unnormalized actor updates]\label{theorems:unnormalized-actor}
	Under Assumptions~\ref{assum:01} and~\ref{assum:is}, let the actor update be unnormalized with stepsize bound $\bar\beta$, range $\vartheta$, and variation $V_u$ (Definition~\ref{def:steps}), let $\eta\in[0,1)$ with $L\bar\beta\le1$ and, if $\eta>0$, $2TR_\pi R_g\bar\beta\le1$ and $2\eta C_F\bar\beta\le1-\eta$, let $\bar\sigma$ and $\epsilon_{\rm bias}$ be evaluated along the iterates, and let $\Delta_{\max} := \sup_W[J^*-J(W)]\le2R_r$ and $\Lambda := 2\sqrt{\vartheta}\,\Delta_1+2V_u\Delta_{\max}$.
	Then, Algorithm~\ref{alg:01} satisfies
	\begin{equation}\label{eq:unnormalized-bound}
		\begin{split}
			\frac{1}{K}\sum_{k=1}^K \ee{\normf{\nabla J(W_k)}}
			\le \sqrt{\frac{\Lambda}{\bar\beta K}} &+ \frac{\sqrt{2}\,\bar\sigma}{\sqrt{K(1-\eta^2)}} + 2\sqrt{1-\eta}\,\bar\sigma \\
			&+ \frac{4\eta C_F\bar\beta R_g}{\sqrt{1-\eta}} + \sqrt{2}\sq{ \epsilon_{\rm bias}+\frac{2\eta D_F\bar\beta R_g}{1-\eta} }.
		\end{split}
	\end{equation}
	In particular, for any $0<c\le\nicefrac{1}{(2C_F)}$ and $K\ge(cL)^2$, the choice $1-\eta = K^{-1/2}$ and $\bar\beta = cK^{-1/2}$ induces
	\begin{equation}\label{eq:unnormalized-rate}
		\frac{1}{K}\sum_{k=1}^K \ee{\normf{\nabla J(W_k)}}
		\le \sq{  \sqrt{\frac{\Lambda}{c}} + (2 + \sqrt{2})\bar\sigma + 4cC_FR_g }\frac{1}{K^{1/4}}
		+\sqrt{2}\big(\epsilon_{\rm bias} + 2cD_FR_g\big).
	\end{equation}
\end{theorem}

The proof in Appendix~\ref{apdx:thm:unnormalized} first bounds the momentum by $2R_g$ for every realization (Lemma~\ref{lemmas:baseline-momentum-tracking}) such that Lemma~\ref{lemmas:momentum-tracking} applies with $\bar\ell_2 = 2\bar\beta R_g$, and then sums the descent inequality, which is quadratic in the gradient norm, by parts to obtain $\Lambda$.
Note that the bound holds in the nuclear norm with the factor $r^{1/2}$, and the stepsize condition $c\le\nicefrac{1}{(2C_F)}$ tightens as $\nicefrac{1}{T}$.
For both classes of actor updates, the following corollary counts the numbers of updates and transitions after which the returned iterate comes within $\epsilon$ of the floor.

\begin{corollary}[Finite-time error guarantee]\label{cor:eps-guarantee}
	Let $\bar\epsilon_{\rm crit} := \epsilon_{\rm lstd}+\epsilon_{\rm app}$ denote the part of the bound of Lemma~\ref{lemmas:frame-bias-noise}(iv) that does not depend on $K$, and let $\epsilon>0$.
	\begin{itemize}
		\item[(i)] Under the conditions and the schedule of Theorem~\ref{theorems:normalized-actor}, let $\hat\sigma$ denote $\bar\sigma$ of \eqref{eq:actor-error-constants} with $\hat\epsilon := \bar\epsilon_{\rm crit}+cL_*$ in place of $\epsilon_{\rm crit}$, and define
		\begin{equation}\label{eq:generic-floor}
			\begin{split}
				&\mathcal{E} := (1+\chi)\sq{ \epsilon_{\rm dist}+\frac{R_\pi(4R_r\tau_{\mix}+\gamma\,\bar\epsilon_{\rm crit})}{(1-\gamma)T}+cD_F }, \\
				&\Gamma := \chi\Big[\frac{\Delta_1}{c}+\frac{cL}{2}\Big]+(1+\chi)\Big[\kappa_\dagger(1+\sqrt{2})\,\hat\sigma+\sqrt{2}\,\kappa_\dagger cC_F+\frac{\gamma R_\pi cL_*}{(1-\gamma)T}\Big].
			\end{split}
		\end{equation}
		Then the returned iterate satisfies $\ee{\normd{\nabla J(W_U)}}\le\mathcal{E}+\epsilon$ for every horizon $K\ge\max\{(\nicefrac{\Gamma}{\epsilon})^3,(cTR_\pi)^{3/2}\}$.
		\item[(ii)] Under the conditions and the schedule of Theorem~\ref{theorems:unnormalized-actor}, let $\hat\sigma_u$ denote $\bar\sigma$ with $\hat\epsilon_u := \bar\epsilon_{\rm crit}+2cR_gL_*$ in place of $\epsilon_{\rm crit}$, and define
		\begin{equation}\label{eq:generic-floor-unnormalized}
			\begin{split}
				&\mathcal{E}_u := \sqrt{2}\sq{ \epsilon_{\rm dist}+\frac{R_\pi(4R_r\tau_{\mix}+\gamma\,\bar\epsilon_{\rm crit})}{(1-\gamma)T}+2cD_FR_g }, \\
				&\Gamma_u := \sqrt{\frac{\Lambda}{c}}+(2+\sqrt{2})\,\hat\sigma_u+4cC_FR_g+\frac{2\sqrt{2}\,\gamma R_\pi cR_gL_*}{(1-\gamma)T}.
			\end{split}
		\end{equation}
		Then the returned iterate satisfies $\ee{\normf{\nabla J(W_U)}}\le\mathcal{E}_u+\epsilon$ for every horizon $K\ge\max\{(\nicefrac{\Gamma_u}{\epsilon})^4,(cL)^2\}$.
	\end{itemize}
	In both cases, Algorithm~\ref{alg:01} consumes $(K+1)T$ transitions, and the floors $\mathcal{E}$ and $\mathcal{E}_u$ do not depend on $K$; Remark~\ref{rem:scope} gives the limit of $\mathcal{E}$ as $T\to\infty$.
\end{corollary}

\begin{proof}
	(i) By Lemma~\ref{lemmas:frame-bias-noise}(iv) with $\bar\ell_2 = \beta = cK^{-2/3}$, we have $\epsilon_{\rm crit}\le\bar\epsilon_{\rm crit}+cL_*K^{-2/3}\le\hat\epsilon$ such that $\bar\sigma\le\hat\sigma$ since $\bar\sigma$ is nondecreasing in $\epsilon_{\rm crit}$, and the floor term of~\eqref{eq:normalized-rate} is at most $\mathcal{E}+\nicefrac{(1+\chi)\gamma R_\pi cL_*}{[(1-\gamma)TK^{2/3}]}$.
	Inserting both bounds into~\eqref{eq:normalized-rate} and using $K^{-2/3}\le K^{-1/3}$, we bound the transient by $\Gamma K^{-1/3}\le\epsilon$.

	(ii) By Lemma~\ref{lemmas:baseline-momentum-tracking}, every step satisfies $\normsp{W_{k+1}-W_k}\le\normf{W_{k+1}-W_k}\le2\bar\beta R_g$.
	Hence, Lemma~\ref{lemmas:frame-bias-noise}(iv) with $\bar\ell_2 = 2\bar\beta R_g = 2cR_gK^{-1/2}$ gives $\epsilon_{\rm crit}\le\bar\epsilon_{\rm crit}+2cR_gL_*K^{-1/2}\le\hat\epsilon_u$ such that $\bar\sigma\le\hat\sigma_u$, and the floor term of~\eqref{eq:unnormalized-rate} is at most $\mathcal{E}_u+\nicefrac{2\sqrt{2}\gamma R_\pi cR_gL_*}{[(1-\gamma)TK^{1/2}]}$.
	Inserting both bounds into~\eqref{eq:unnormalized-rate} and using $K^{-1/2}\le K^{-1/4}$, we bound the transient by $\Gamma_uK^{-1/4}\le\epsilon$.

	In both cases, the identity $\ee{\normd{\nabla J(W_U)}} = \frac{1}{K}\sum_{k=1}^{K}\ee{\normd{\nabla J(W_k)}}$, which also holds in the Frobenius norm, completes the proof since every frame consists of $T$ transitions.
\end{proof}

\begin{remark}[Residual error]\label{rem:scope}
The floor $\mathcal E$ in Corollary~\ref{cor:eps-guarantee}(i) separates the error sources with different remedies.
The mismatch term $(1+\chi)\epsilon_{\rm dist}$ is the bias of uniformly weighted samples, which estimate the stationary means rather than the discounted means as in practical AC implementations~\citep{Thomas2014bias,Nota2020policy,Zhang2022discounting}.
Note that the mismatch term depends on neither the critic nor $T$ or $K$, enters only once since $\bar B$ telescopes (Remark~\ref{rem:bias}), and vanishes when $\mu_W = \zeta_W$.
For fixed $\lambda>0$ and $c$, every other term decays at least as $\nicefrac{1}{T}$.
Therefore, Corollary~\ref{cor:eps-guarantee} is a finite-time guarantee up to $(1+\chi)\epsilon_{\rm dist}+\oo{\nicefrac{1}{T}}$ rather than the convergence to a stationary point of $J$; removing $\epsilon_{\rm dist}$ calls for a different sampling protocol rather than a different actor update.
Moreover, the critic error enters only through the bootstrap with weight $\nicefrac{\gamma R_\pi}{[(1-\gamma)T]}$ and through the noise constant $\bar\sigma$.
Hence, the transient exponents hold even under the uniform bound $\epsilon_{\rm crit}\le R_v+\nicefrac{R_r}{(1-\gamma)}$ such that the actor guarantees are robust to an inaccurate critic, whereas a more accurate critic via more NSI recursions, longer frames, or richer features lowers both the floor and the noise constant.
Appendix~\ref{apdx:actor-floor} expands the error terms and gives an example with nonzero mismatch despite one-step mixing.
\end{remark}

\section{Instances}\label{sec:instances}
In this section, we apply Theorems~\ref{theorems:normalized-actor} and~\ref{theorems:unnormalized-actor} to the four actor updates of Table~\ref{tab:steps}.
The detailed proofs can be found in Appendix~\ref{apdx:instances}. 
Since all instances share the inverse-free LSTD critic, the PG estimator, and the ICM, we compare them in the nuclear norm.

\subsection{Muon Actor}\label{sec:nsac}
A vanilla SGD actor moves along $G_k$, which can be dominated by a few large singular directions~\citep{Jordan2024}.
Different from the vanilla SGD actor, the Muon actor moves along an approximation of the polar factor of $M_k$ by the NSO recursions \eqref{eq:nso-step}, whose $\kappa_2$th-order NSO polynomial keeps the nonzero singular values of every iterate in $(0,1]$.
The following lemma establishes the convergence and alignment of finite NSO recursions with proof relegated to Appendix~\ref{apdx:instances}.

\begin{lemma}[Convergence and alignment of finite NSO]\label{lemmas:nso-polar-convergence}
	Let
	\begin{equation*}
		f(x) := x\sum_{i=0}^{\kappa_2-1}4^{-i}\binom{2i}{i}(1-x^2)^i,
		\qquad
		s_P := \sup_{x\in[0,1]}x\,\big(1-f^{P}(x)\big)
	\end{equation*}
	where $f^{P}$ denotes the $P$-fold composition of $f$.
	Then $s_P\le(2e\,\kappa_2^{P})^{-1/2}$, and, for every $k$ and every realization, the NSO iterates in \eqref{eq:nso-step} satisfy the following.
	\begin{itemize}
		\item[(i)] If $M_k\neq0$, every iterate $N_{k,p}$ keeps the singular vectors and the rank of $M_k$, has the singular values $f^{p}(\sigma_i(M_k)/\normf{M_k})$, and satisfies $\normsp{N_{k,p}}\le1$, and $N_{k,p}$ converges to the polar factor $\Theta_k$ of $M_k$ locally with order $\kappa_2$ in $p$.
		Moreover, with $a_k := \sigma_{\min}^+(M_k)/\normf{M_k}$, the polar residual $R_k := N_{k,P_2}-\Theta_k$ satisfies
		\begin{equation*}
			\normsp{R_k} = 1-\sigma_{\min}^+(N_{k,P_2})\le\big(1-a_k^2\big)^{\kappa_2^{P_2}}<1.
		\end{equation*}
		If $M_k = 0$, then $N_{k,p} = 0$, and we set $\Theta_k = R_k = 0$.
		\item[(ii)] With $r = \min\{m,n\}$, the NSO output is aligned with the momentum as
		\begin{equation}\label{eq:nso-alignment}
			\inpf{M_k,N_{k,P_2}}\ge(1-r\,s_{P_2})\normnu{M_k}
		\end{equation}
		where $\kappa_2^{P_2}\ge4r^2$ implies $\inpf{M_k,N_{k,P_2}}\ge0.78\,\normnu{M_k}$.
	\end{itemize}
\end{lemma}

By Lemma~\ref{lemmas:nso-polar-convergence}, the Muon actor update is normalized in the spectral norm with the alignment factor $\chi = \nicefrac{1}{(1-r\,s_{P_2})}$ whenever $r\,s_{P_2}<1$, i.e., with the constants of Table~\ref{tab:steps}.
Note that Lemma~\ref{lemmas:nso-polar-convergence} requires no conditioning assumption on the momentum, which is imposed by the finite-NSO analysis of~\citet{kim2026convergence}.

\begin{corollary}[Muon]\label{cor:nsac}
	Under Assumptions~\ref{assum:01} and~\ref{assum:is}, let $P_2$ satisfy $r\,s_{P_2}<1$ (e.g., $\kappa_2^{P_2}\ge4r^2$) with $s_{P_2}$ of Lemma~\ref{lemmas:nso-polar-convergence}, and let $\chi := \nicefrac{1}{(1-r\,s_{P_2})}$ (hence $\chi\le1.28$ if $\kappa_2^{P_2}\ge4r^2$).
	Then Algorithm~\ref{alg:01} with the Muon step \eqref{eq:nso-step} satisfies \eqref{eq:normalized-bound} with $\normd{\cdot} = \normnu{\cdot}$ and $\kappa_\dagger = r^{1/2}$ under the conditions of Theorem~\ref{theorems:normalized-actor}; in particular, for any $c>0$ and $K\ge(cT R_\pi)^{3/2}$, the choice $1-\eta = K^{-2/3}$ and $\beta = cK^{-2/3}$ induces
	\begin{equation}\label{eq:nsac-stationarity-rate}
		\begin{split}
			&\frac{1}{K}\sum_{k=1}^{K}\ee{\normnu{\nabla J(W_k)}}
			\le \chi\Big[\frac{\Delta_1}{c}\,\frac{1}{K^{1/3}} + \frac{cL}{2 K^{2/3}}\Big] \\
			&\quad+ (1+\chi)\,r^{1/2}\big[(1+\sqrt{2})\,\bar\sigma + \sqrt{2}\,cC_F\big]\frac{1}{K^{1/3}}
			+ (1+\chi)\big(\epsilon_{\rm bias}+cD_F\big)
		\end{split}
	\end{equation}
	and the guarantee of Corollary~\ref{cor:eps-guarantee}(i) holds in the nuclear norm with $\kappa_\dagger = r^{1/2}$.
\end{corollary}

The NSI approximation in the critic enters only through $\epsilon_{\rm crit}$, whereas the NSO approximation in the actor enters multiplicatively through $\chi\le1.28$ once $\kappa_2^{P_2}\ge4r^2$, i.e., after $\oo{\log r}$ GEMM-only NSO recursions.
Moreover, if $\nicefrac{\sigma_{\min}^+(M_k)}{\normf{M_k}}$ is bounded away from zero, the gap between the realized alignment and one decays doubly exponentially in $P_2$ since $\inpf{M_k,N_{k,P_2}}\ge(1-\normsp{R_k})\normnu{M_k}$.
In \eqref{eq:nsac-stationarity-rate}, only the noise and correction terms carry the factor $r^{1/2}$ due to the three facts: (i) the ascent of the Muon step along the momentum is at least $\nicefrac{\beta\normnu{M_k}}{\chi}$, and \eqref{eq:policy-regularity} pairs the score and its change with the spectral norm of the step such that the optimization term and the constants $C_F$, $D_F$, and $L$ are free of $r$; (ii) the bias part of the tracking error, which is controlled in the nuclear norm, is paired with $\normsp{N_{k,P_2}}\le1$; and (iii) only the noise part, which is averaged in the Frobenius norm, is converted by $\normnu{e^{\rm n}_k}\le r^{1/2}\normf{e^{\rm n}_k}$.
Note that the mismatch $\epsilon_{\rm dist}$ can exceed $\sup_W\normf{\bar B(W)}$ by up to $r^{1/2}$, whereas its a priori bound is free of $r$.
Hence, the noise term fixes the order $r^{1/2}$ of the transient for every stepsize constant, and $c$ only shifts $r^{1/2}$ between the optimization and correction terms.
More specifically, for a dimension-free constant $c_0$, the choice $c = c_0$ makes the optimization and second-order terms, the floor, and the condition on $K$ free of $r$, whereas $c = \nicefrac{c_0}{r^{1/2}}$ makes the correction term free of $r$; for both choices, the horizon of Corollary~\ref{cor:eps-guarantee}(i) is of order $r^{3/2}\epsilon^{-3}$.

\subsection{NSGDM Actor}\label{sec:nsgdm}
The NSGDM step $\nicefrac{\beta M_k}{\normf{M_k}}$ is the normalized momentum of~\citet{CutkoskyMehta2020} with our importance correction, and is closely related to the normalized variance-reduced PG method of~\citet{Barakat2023general} that uses Monte Carlo episodes restarted from the initial distribution without a critic.
Since $\inpf{M_k,\nicefrac{M_k}{\normf{M_k}}} = \normf{M_k}$, the NSGDM actor update is normalized in the self-dual Frobenius norm with $\chi = \kappa_\dagger = 1$.

\begin{corollary}[NSGDM]\label{cor:nsgdm}
	Under Assumptions~\ref{assum:01} and~\ref{assum:is}, Algorithm~\ref{alg:01} with the NSGDM step satisfies \eqref{eq:normalized-bound} with $\normd{\cdot} = \normf{\cdot}$ and $\chi = \kappa_\dagger = 1$ under the conditions of Theorem~\ref{theorems:normalized-actor}; in particular, for any $c>0$ and $K\ge(cT R_\pi)^{3/2}$, the choice $1-\eta = K^{-2/3}$ and $\beta = cK^{-2/3}$ induces
	\begin{equation}\label{eq:nsgdm-stationarity-rate}
		\begin{split}
			\frac{1}{K}\sum_{k=1}^{K}\ee{\normnu{\nabla J(W_k)}}
			\le r^{1/2}\bigg[&\Big(\frac{\Delta_1}{c}+2(1+\sqrt{2})\,\bar\sigma+2\sqrt{2}\,cC_F\Big)\frac{1}{K^{1/3}} \\
			&+\frac{cL}{2K^{2/3}}+2\big(\epsilon_{\rm bias}+cD_F\big)\bigg]
		\end{split}
	\end{equation}
	where the bracket bounds $\frac{1}{K}\sum_{k=1}^{K}\ee{\normf{\nabla J(W_k)}}$.
\end{corollary}

\subsection{SGDM and AdaNorm Actors}\label{sec:sgdm-adanorm}
The SGDM step $\beta M_k$ is unnormalized with $\bar\beta = \beta$, $\vartheta = 1$, and $V_u = 0$.
Since the accumulator of the AdaNorm step \eqref{eq:madagrad-update} with $\alpha\le\nicefrac{1}{(2K)}$ stays in $[\nicefrac{n_0}{2},n_0+\nicefrac{R_g^2}{2}]$ and varies little, the AdaNorm step is unnormalized with the constants of Table~\ref{tab:steps} (Appendix~\ref{apdx:instances}).

\begin{corollary}[SGDM and AdaNorm]\label{cor:sgdm-adanorm}
	Under Assumptions~\ref{assum:01} and~\ref{assum:is}, let $\bar\beta := \beta$ for the SGDM step and $\bar\beta := \beta\sqrt{\nicefrac{2}{n_0}}$ for the AdaNorm step with a $K$-independent $n_0>0$ and $\alpha\in[0,\nicefrac{1}{(2K)}]$, and let $\Lambda := 2\Delta_1$ for SGDM and $\Lambda := 2\sqrt{\vartheta}\,\Delta_1+(\vartheta-1)\Delta_{\max}$ with $\vartheta := 2+\nicefrac{R_g^2}{n_0}$ for AdaNorm.
	Then Algorithm~\ref{alg:01} with either step satisfies \eqref{eq:unnormalized-bound}, multiplied by $r^{1/2}$, in the nuclear norm under the conditions of Theorem~\ref{theorems:unnormalized-actor}; in particular, for any $0<c\le\nicefrac{1}{(2C_F)}$ and $K\ge(cL)^2$, the choice $1-\eta = K^{-1/2}$ and $\bar\beta = cK^{-1/2}$, i.e., $\beta = cK^{-1/2}$ for SGDM and $\beta = c\sqrt{\nicefrac{n_0}{2}}\,K^{-1/2}$ for AdaNorm, induces
	\begin{equation}\label{eq:sgdm-stationarity-rate}
		\begin{split}
			&\frac{1}{K}\sum_{k=1}^K \ee{\normnu{\nabla J(W_k)}} \\
			&\le r^{1/2}\sq{ \Big(\sqrt{\frac{\Lambda}{c}} + (2 + \sqrt{2})\bar\sigma + 4cC_FR_g\Big)\frac{1}{K^{1/4}} 
				+\sqrt{2}\big(\epsilon_{\rm bias} + 2cD_FR_g\big) }.
		\end{split}
	\end{equation}
\end{corollary}

The AdaNorm bound is the SGDM bound with $\Lambda\ge2\sqrt{2}\,\Delta_1+\Delta_{\max}$ in place of $2\Delta_1$, which accounts for the variation of the effective stepsize $\nicefrac{\beta}{\sqrt{n_k}}\in[\nicefrac{\bar\beta}{\sqrt{\vartheta}},\bar\beta]$ through the summation by parts.
Note that our analysis covers AdaNorm in its stabilized regime $\alpha\le\nicefrac{1}{(2K)}$, in which the effective stepsize varies within the factor $\sqrt{\vartheta}$.
Otherwise, a $K$-independent $\alpha$ allows $n_k$ to decay as $(1-\alpha)^kn_0$ such that the effective stepsizes, the momentum, and the importance weights admit no deterministic bound independent of $K$.

\subsection{Comparison}\label{sec:comparison}

\begin{table}[ht]
	\centering\small
	\caption{Bounds on $\frac{1}{K}\sum_{k=1}^{K}\ee{\normnu{\nabla J(W_k)}}$ term by term (Corollaries~\ref{cor:nsac}--\ref{cor:sgdm-adanorm}).}
	\label{tab:rate}
	\medskip
	\setlength{\tabcolsep}{2pt}
	\begin{tabularx}{\linewidth}{@{}lccccc>{\centering\arraybackslash}X@{}}
		\toprule
		Actor & Rate & Optimization & \shortstack{Noise\\($\times r^{1/2}\bar\sigma$)} & \shortstack{Importance\\correction\\($\times r^{1/2}cC_F$)} & \shortstack{Second\\order} & Error floor\textsuperscript{c} \\
		\midrule
		Muon\textsuperscript{a} & $K^{-1/3}$ & $\chi\Delta_1/c$ & $(1+\sqrt{2})(1+\chi)$ & $\sqrt{2}(1+\chi)$ & $\chi cL/2$ & \shortstack{$(1+\chi)$\\$\times(\epsilon_{\rm bias}+cD_F)$} \\
		NSGDM\textsuperscript{a} & $K^{-1/3}$ & $r^{1/2}\Delta_1/c$ & $2(1+\sqrt{2})$ & $2\sqrt{2}$ & $r^{1/2}cL/2$ & \shortstack{$2r^{1/2}$\\$\times(\epsilon_{\rm bias}+cD_F)$} \\
		SGDM\textsuperscript{b} & $K^{-1/4}$ & $r^{1/2}(2\Delta_1/c)^{1/2}$ & $2+\sqrt{2}$ & $4R_g$ & -- & \shortstack{$\sqrt{2r}\,(\epsilon_{\rm bias}$\\$+2cD_FR_g)$} \\
		AdaNorm\textsuperscript{b} & $K^{-1/4}$ & $r^{1/2}(\Lambda/c)^{1/2}$ & $2+\sqrt{2}$ & $4R_g$ & -- & \shortstack{$\sqrt{2r}\,(\epsilon_{\rm bias}$\\$+2cD_FR_g)$} \\
		\bottomrule
	\end{tabularx}

	\smallskip
	\begin{minipage}{0.92\linewidth}\footnotesize
	For fixed $T$ and $\lambda$ and a dimension-free stepsize constant $c$, the optimization, noise, and importance-correction entries multiply the rate, and the second-order entries multiply $K^{-2/3}$.\par
	\textsuperscript{a}Any $c>0$ with $K\ge(cT R_\pi)^{3/2}$; for Muon, $\chi\le1.28$ once $\kappa_2^{P_2}\ge4r^2$, i.e., after $\oo{\log r}$ NSO recursions per update.\par
	\textsuperscript{b}With $c\le\nicefrac{1}{(2C_F)}$ and $K\ge(cL)^2$, which absorbs the second-order term.\par
	\textsuperscript{c}All floors are stated with the nuclear-norm mismatch $\epsilon_{\rm dist}$; the Frobenius analyses behind the last three rows could use the possibly smaller $\sup_W\normf{\bar B(W)}$ instead.
	\end{minipage}
\end{table}

Since the four actors share the critic, the frames, the PG estimator, and the ICM, the differences among the bounds in Table~\ref{tab:rate} stem from the actor updates up to the evaluation of $\bar\sigma$ and $\epsilon_{\rm bias}$ along the respective iterates and the constant factors, e.g., $\sqrt{2}$ and $1+\chi$, that reflect the proof technique.

{\bfseries Rate.}~The bounds decay as $\oo{K^{-1/3}}$ for Muon and NSGDM and as $\oo{K^{-1/4}}$ for SGDM and AdaNorm.
Since all bounds contain the drift $\nicefrac{\eta D_F\beta}{(1-\eta)}$ of the frame bias (cf.~\eqref{eq:normalized-bound}), they require $\beta = \oo{1-\eta}$.
Moreover, the SGDM and AdaNorm steps, which scale with $M_k$, additionally require $2\eta C_F\bar\beta\le1-\eta$ and $2T R_\pi R_g\bar\beta\le1$ to keep $\normf{M_k}\le2R_g$ and the importance weights bounded (Lemma~\ref{lemmas:baseline-momentum-tracking}), whereas the Muon and NSGDM actors admit any $c>0$.
Under the proportional schedule, the descent inequalities of normalized actor updates, which are linear in the tracking error, balance $\nicefrac{\Delta_1}{(\beta K)}$ against $\sqrt{1-\eta}\,\bar\sigma$ at $K^{-1/3}$, whereas those of unnormalized actor updates, which are quadratic in the gradient norm, balance $\nicefrac{\Delta_1}{(\beta K)}$ against $(1-\eta)\bar\sigma^2$ at $K^{-1/2}$ for the squared norm.
Hence, the better exponent in these bounds is due to normalization, which the Muon actor update shares with NSGDM.
Note that the coupling $1-\eta\propto\beta^2$ behind the complexity $\oo{\epsilon^{-3}}$ of the unnormalized ICM actor with unbiased estimates~\citep{Huang2020} is unavailable here since it lets the drift of the frame bias grow as $\nicefrac{D_F}{\beta}$.
Besides, without the importance correction, both normalized bounds decay only as $\oo{K^{-1/4}}$ (Remark~\ref{rem:is_plain}); therefore, the faster rate in these upper bounds requires both normalization and the importance correction.

{\bfseries Dimension dependence.}~All four transients carry the factor $r^{1/2}$ through their noise terms, whereas the bounds separate term by term.
More specifically, the SGDM, AdaNorm, and NSGDM analyses bound the averaged Frobenius norm of the PG, in which every quantity other than $\epsilon_{\rm dist}$ is free of $r$, and convert the result to the nuclear norm as a whole, whereas only the noise and correction terms of the Muon bound carry $r^{1/2}$ with a dimension-free $c$ and $\kappa_2^{P_2}\ge4r^2$.
It is worth stressing that the descent arguments that control the bias only through its norm cannot remove the factor $r^{1/2}$ of the baseline floors: a tracking error of nuclear norm $b$ can cancel the first-order progress of the SGDM, AdaNorm, and NSGDM steps although $\normnu{\nabla J(W_k)}$ is of the order $r^{1/2}b$, whereas it cannot cancel the progress of the polar factor once $\normnu{\nabla J(W_k)}>2b$.
At the same $c$, the NSGDM row is the Muon row with $r^{1/2}$ in place of $\chi$ on the optimization and second-order terms, $2$ in place of $1+\chi$ on the noise and correction terms, and $2r^{1/2}$ in place of $1+\chi$ on the floor.
Hence, when NSGDM uses a dimension-free constant $c_0$, the Muon bound is no larger than the NSGDM bound term by term up to the factor $\chi$ for every $c\in[\nicefrac{c_0}{r^{1/2}},c_0]$: the noise terms coincide up to $\nicefrac{(1+\chi)}{2}$; at $c = c_0$, the optimization and second-order terms are smaller by $r^{1/2}$; at $c = \nicefrac{c_0}{r^{1/2}}$, the correction and second-order terms are smaller by $r^{1/2}$ and $r$, respectively; and the floor with the common mismatch $\epsilon_{\rm dist}$ is smaller by $r^{1/2}$.
With each constant tuned to minimize the sum of the optimization and correction terms, these terms scale as $r^{1/4}$ for Muon and as $r^{1/2}$ for the other actors.

{\bfseries When orthogonalization helps.}~The advantage of Muon is stated in the nuclear norm, i.e., the dual norm of the spectral norm in which the Muon actor update is normalized, and its size depends on the curvature constants.
In the Frobenius norm, the roles reverse: the bracket of \eqref{eq:nsgdm-stationarity-rate} carries no factor $r^{1/2}$, whereas the Muon bound keeps $r^{1/2}$ on its noise and correction terms such that the NSGDM bound is the smaller one.
Moreover, NSGDM needs only the Frobenius counterparts of \eqref{eq:policy-regularity} and \eqref{eq:return-ascent-bound}, whose constants can be smaller; even for scores of rank one, the curvature of $J$, which averages over states, can differ between the two norms.
Let $L_2$ and $L_F$ denote the smallest constants for which \eqref{eq:return-ascent-bound} holds with $\nicefrac{L_2}{2}\normsp{W'-W}^2$ and with $\nicefrac{L_F}{2}\normf{W'-W}^2$, respectively, such that $L_F\le L_2\le\min\{L,rL_F\}$.
Without noise and bias, an exact polar step with the best stepsize gains at least $\nicefrac{\normnu{\nabla J(W)}^2}{(2L_2)}$, and an exact NSGDM step gains at least $\nicefrac{\normf{\nabla J(W)}^2}{(2L_F)}$.
Therefore, orthogonalization guarantees the larger gain if and only if the nuclear rank $\nicefrac{\normnu{\nabla J(W)}^2}{\normf{\nabla J(W)}^2}$ of the PG exceeds the curvature ratio $\nicefrac{L_2}{L_F}$, where both ratios lie in $[1,r]$.
For example, $W\mapsto-\frac{1}{2}\mathrm{tr}(W\Sigma W^\top)$ with a positive semidefinite $\Sigma\in\mathbb{R}^{n\times n}$ and $m\ge n$ has $L_2 = \mathrm{tr}(\Sigma)$ and $L_F = \lambda_{\max}(\Sigma)$, whose ratio, i.e., the effective rank of $\Sigma$, can be as large as $r$.
Hence, orthogonalization pays off for the PGs whose nuclear rank exceeds the curvature ratio, and the above term-by-term advantage with a common stepsize constant holds for the upper bounds in the nuclear criterion.

{\bfseries Computation and frame length.}~The actor evaluates $\log\pi$ and $\nabla\log\pi$ at $W_k$ and $W_{k-1}$ on the frame.
Orthogonalization costs the factor $\chi\le1.28$ and $\oo{\log r}$ NSO recursions per update, i.e., $\oo{P_2\kappa_2mnr}$ operations in total when the Gram matrix is formed on the smaller side, against $\oo{mn}$ operations for the other actor updates.
Note that such an overhead is dominated by the cost of forming the frame estimates, which is at least of order $Tmn$ for a dense layer, once $T\ge P_2\kappa_2r$.
All actors consume the same frames and share the critic of Section~\ref{sec:critic}, which costs $\oo{dT^2+\kappa_1P_1T^3}$ operations per update, i.e., $\oo{dT^2}$ operations to form the $T\times T$ matrix $Y_k^\top\Phi_k$ and $\oo{\kappa_1T^3}$ operations per NSI recursion.

A longer frame reduces the context-dependent bias as $\nicefrac{1}{T}$ and the noise bound as $T^{-1/2}$, whereas it increases $C_F$ and tightens the step condition $TR_\pi\beta\le1$.
Moreover, each update consumes $T$ transitions.
Since the ridge-regularized pathwise LSTD critic is well posed for every $\lambda>0$ and $T\ge1$, the choice of $T$ is governed by such a bias--cost tradeoff.

For $\Delta_1>0$, the sufficient horizon of Corollary~\ref{cor:eps-guarantee}(i) grows at least as $T^{3/2}\epsilon^{-3}$ for every $c$ (Appendix~\ref{apdx:frame-complexity}), whereas the cap $c\le\nicefrac{1}{(2C_F)}$ makes that of Corollary~\ref{cor:eps-guarantee}(ii) grow at least as $T^2\epsilon^{-4}$.
Therefore, normalized actor updates also scale better with the frame length.

\begin{remark}[Transition complexity]\label{rem:complexity}
At a fixed frame length, the ICM removes the noise term $\oo{T^{-1/2}}$ that a momentum-free normalized actor update retains in its floor.
Instead, when the frame length grows with a target $\epsilon\in(0,1]$ for the residual in \eqref{eq:normalized-bound} after excluding the mismatch term $(1+\chi)\epsilon_{\rm dist}$, for fixed $\lambda$ and problem constants, $T$ of order $\epsilon^{-1}$ with a tuned momentum factor and stepsize gives $\oo{\epsilon^{-9/2}}$ transitions, or $\oo{\epsilon^{-11/2}}$ transitions under the schedule $1-\eta = K^{-2/3}$ of Theorem~\ref{theorems:normalized-actor}, whereas a momentum-free actor update with $T$ of order $\epsilon^{-2}$ gives $\oo{\epsilon^{-4}}$ transitions.
Note that these counts are certified by the sufficient bounds whose correction constant $C_F$ grows linearly with $T$ (Section~\ref{sec:spg-analysis}).
The schedules and derivations are relegated to Appendix~\ref{apdx:frame-complexity}.
\end{remark}

\section{Conclusions}\label{sec:conclusion}
In this work, we have established the first finite-time convergence rates for ICM actors that learn along a single Markovian trajectory with an inverse-free ridge-regularized pathwise LSTD critic.
Up to the stationary-to-discounted mismatch and an $\oo{\nicefrac{1}{T}}$ frame bias, the averaged gradient norm decays as $\oo{K^{-1/3}}$ for normalized actor updates and as $\oo{K^{-1/4}}$ for unnormalized ones.
Among the instances, Muon with a dimension-free stepsize confines the factor $r^{1/2}$ to the noise and correction terms, and orthogonalization guarantees a larger noiseless ascent than normalization exactly when the nuclear rank of the PG exceeds the spectral-to-Frobenius curvature ratio.

\backmatter

\begin{appendices}

\section{Notation and Stationarity Criterion}\label{apdx:notation}
Lowercase letters denote vectors and uppercase letters matrices, unless they denote scalars such as $d$, $m$, $n$, $T$, $K$, and $L$; $I_d$ is the $d\times d$ identity, $[K] := \{1,\ldots,K\}$, $\ind{\cdot}$ is the indicator of an event, and $(\cdot)^{\top}$ is the transpose.
For a vector, $\normsp{\cdot}$ is the Euclidean norm; for a matrix, $\normsp{\cdot}$, $\normf{\cdot}$, and $\normnu{\cdot}$ are the spectral, Frobenius, and nuclear norms, $\inpf{\cdot,\cdot}$ is the Frobenius inner product, $\sigma_i(\cdot)$, $\sigma_{\min}(\cdot)$, and $\sigma_{\max}(\cdot)$ are the $i$th largest, smallest, and largest singular values, and $\sigma_{\min}^+(\cdot)$ is the smallest nonzero singular value of a nonzero matrix.
For a bounded function $f$ on ${\cal S}$, $\norm{f}_\infty := \sup_s|f(s)|$, and, for a finite signed measure $\nu$, $\normtv{\nu} := \sup_{\cal B}|\nu({\cal B})|$ over measurable sets ${\cal B}$, which is half the total variation of $\nu$ if $\nu$ has total mass zero.
The symbols $\mathbb{E}$ and $\mathbb{P}$ denote expectation and probability ($\mathbb{P}$ also denotes the transition kernel), and $\oo{\cdot}$ hides constant factors, which may depend on problem constants such as $R_r$, $R_\pi$, $\gamma$, and $\lambda$ but not on the quantities displayed in its argument; $\tildeo{\cdot}$ additionally hides polylogarithmic factors.
The symbol $r$ denotes both the reward function and $r := \min\{m,n\}$, the largest rank of an $m\times n$ matrix, such that $\normnu{X}\le r^{1/2}\normf{X}$; likewise, $n$ denotes the number of columns of $W$ and, as $n_k$, the accumulator of the AdaNorm actor (Section~\ref{sec:generic-alg}); the meaning is clear from context.

For a norm $\norm{\cdot}$ on $\mathbb{R}^{m\times n}$, the dual norm is $\normd{Y} := \sup_{\norm{X}\le1}\inpf{Y,X}$, such that the spectral and nuclear norms are dual to each other and the Frobenius norm is self-dual, and $\kappa_\dagger := \sup_{X\neq0}\nicefrac{\normd{X}}{\normf{X}}$, which is $r^{1/2}$ for the nuclear norm and $1$ for the Frobenius norm.
If $\norm{\cdot}\ge\normsp{\cdot}$, its unit ball lies in that of the spectral norm, such that $\normd{\cdot}\le\normnu{\cdot}$.
Since $J$ is nonconcave, we measure the progress of an actor by $\frac{1}{K}\sum_{k=1}^{K}\ee{\normd{\nabla J(W_k)}}$ in the dual norm of its step (Definition~\ref{def:steps}), and we compare actors in the nuclear norm, which bounds every such dual norm because $\normd{\cdot}\le\normnu{\cdot}\le r^{1/2}\normf{\cdot}$.

\section{Non-Singularity of the Ridge-Regularized Pathwise LSTD}\label{apdx:bk}
We show that every $\lambda>0$ makes both ridge-regularized pathwise LSTD systems invertible, for every frame length and without any rank condition on $\Phi_k$ or $Y_k$.

\begin{proof}
Per frame $k$ with ridge parameter $\lambda\neq0$, we have $\det(\lambda I_d+\Phi_kY_k^\top) = \lambda^{d-T}\det(\lambda I_T+Y_k^\top\Phi_k)$ with $\hat A_k = \Phi_kY_k^\top$ and $H_k = \lambda I_T + Y_k^\top\Phi_k$.
Therefore, we can show that $H_k$ is non-singular if and only if $\hat A_k+\lambda I_d$ is.

For $v\in\mathbb{R}^d$, we define $z_t := v^\top\phi_{k,t}$ for $t<T$ and $z_T := 0$ such that the zero last column of $\Phi_k'$ induces $T v^\top\hat A_kv = \sum_{t=0}^{T-1}(z_t^2-\gamma z_tz_{t+1})$.
Based on the facts $z_t^2-\gamma z_tz_{t+1} = (1-\gamma)z_t^2+\frac{\gamma}{2}(z_t^2-z_{t+1}^2)+\frac{\gamma}{2}(z_t-z_{t+1})^2$ and $z_T = 0$, we obtain
\begin{equation}\label{eq:empirical-lstd-accretivity}
	v^\top(\hat A_k+\lambda I_d)v
	= \frac{1}{T}\sum_{t=0}^{T-1}\sq{ (1-\gamma)z_t^2+\frac{\gamma}{2}(z_t-z_{t+1})^2 } + \frac{\gamma z_0^2}{2T}+\lambda\normsp{v}^2
	\ge \lambda\normsp{v}^2.
\end{equation}

We observe that  \eqref{eq:empirical-lstd-accretivity} secures the smallest singular value of $\hat A_k+\lambda I_d$ to be positive ridge parameter $\lambda$.
Therefore, we conclude that the two matrices 
$\lambda I_d + \Phi_kY_k^\top$ and 
$\lambda I_T + Y_k^\top\Phi_k$ are non-singular.
\end{proof}

\section{Proof of Lemma \ref{lemmas:return-smoothness}}\label{apdx:smoothness}
\begin{proof}
	We make the argument of \citet[Lemma~3.2]{Zhang2020} explicit for the normalized return.
	Fix a start state $s_0$, let ${\cal H}_t := (s_0,a_0,s_1,\ldots,s_t,a_t)$, and let $p^W_t$ denote its law under $\pi_W$ from $s_0$.
	Only the policy factors of $p^W_t$ depend on $W$, and $\nabla_W\log p^W_t({\cal H}_t) = \sum_{i=0}^{t}\nabla\log\pi_W(a_i|s_i)$; hence, differentiating $V_W(s_0) = \sum_{t\ge0}\gamma^t\,\mathbb{E}_{p^W_t}[r(s_t,a_t)]$ term by term and under each expectation gives
	\begin{equation*}
		\nabla V_W(s_0) = \sum_{t\ge0}\gamma^t\,\mathbb{E}_{p^W_t}\big[\Upsilon^W_t\big],
		\qquad
		\Upsilon^W_t({\cal H}_t) := r(s_t,a_t)\sum_{i=0}^{t}\nabla\log\pi_W(a_i|s_i).
	\end{equation*}
	
	Both interchanges are justified by dominated convergence: for $\normsp{W''-W}\le\delta$ with $\gamma e^{R_\pi\delta}<1$, H\"older's inequality and \eqref{eq:score-bound}, integrated along the segment from $W$ to $W''$, give $|\log\pi_{W''}(a|s)-\log\pi_W(a|s)|\le R_\pi\delta$, such that the density of $p^{W''}_t$ is at most $e^{(t+1)R_\pi\delta}$ times that of $p^W_t$, the gradient of the $t$th integrand at $W''$ is at most $(t+1)R_rR_\pi e^{(t+1)R_\pi\delta}$ times the density of $p^W_t$ in the nuclear norm, and $\sum_{t\ge0}\gamma^t(t+1)e^{(t+1)R_\pi\delta}<\infty$.
	Fix $W,W'$, let $\ell_2 := \normsp{W-W'}$, and split $\nabla V_W(s_0)-\nabla V_{W'}(s_0) = \sum_{t\ge0}\gamma^t\,\mathbb{E}_{p^W_t}[\Upsilon^W_t-\Upsilon^{W'}_t]+\sum_{t\ge0}\gamma^t(\mathbb{E}_{p^W_t}-\mathbb{E}_{p^{W'}_t})[\Upsilon^{W'}_t]$.
	
	\emph{Change of the score.}~By \eqref{eq:score-lipschitz} and the triangle inequality, $\normnu{\Upsilon^W_t-\Upsilon^{W'}_t}\le(t+1)R_rL_\pi\ell_2$, such that $\sum_{t\ge0}(t+1)\gamma^t = (1-\gamma)^{-2}$ bounds the first sum by $\nicefrac{R_rL_\pi\ell_2}{(1-\gamma)^2}$ in the nuclear norm.
	
	\emph{Change of the trajectory distribution.}~Along $W_\tau := W'+\tau(W-W')$, $\tau\in[0,1]$, we have $\frac{d}{d\tau}\pi_{W_\tau}(a|s) = \pi_{W_\tau}(a|s)\inpf{\nabla\log\pi_{W_\tau}(a|s),W-W'}$, where H\"older's inequality and \eqref{eq:score-bound} bound the inner product by $R_\pi\ell_2$; integrating over $\tau$ and $a$ induces $\sup_s\normtv{\pi_W(\cdot|s)-\pi_{W'}(\cdot|s)}\le\nicefrac{R_\pi\ell_2}{2}$.
	Since $p^W_t$ and $p^{W'}_t$ differ only in the $t+1$ policy factors, replacing them one at a time gives $\normtv{p^W_t-p^{W'}_t}\le\nicefrac{(t+1)R_\pi\ell_2}{2}$, and $\normnu{\Upsilon^{W'}_t}\le(t+1)R_rR_\pi$ and the bound on $\norm{\int f\,d\nu}$ of Appendix~\ref{apdx:prelim} bound the $t$th summand of the second sum by $2\gamma^t(t+1)R_rR_\pi\normtv{p^W_t-p^{W'}_t}\le\gamma^t(t+1)^2R_rR_\pi^2\ell_2$; hence, $\sum_{t\ge0}(t+1)^2\gamma^t = \nicefrac{(1+\gamma)}{(1-\gamma)^3}$ bounds the second sum by $\nicefrac{(1+\gamma)R_rR_\pi^2\ell_2}{(1-\gamma)^3}$.
	
	\emph{Normalization.}~Averaging over $s_0\sim\rho$, which commutes with the gradient by dominated convergence since $\normnu{\nabla V_{W''}(s_0)}\le\nicefrac{R_rR_\pi}{(1-\gamma)^2}$ for all $W''$ and $s_0$, and multiplying by $1-\gamma$, since $J(W) = (1-\gamma)\mathbb{E}_{s_0\sim\rho}[V_W(s_0)]$, yield $\normnu{\nabla J(W)-\nabla J(W')}\le L\ell_2$, which is \eqref{eq:return-smoothness}.
	Finally, by H\"older's inequality, $J(W')-J(W)-\inpf{\nabla J(W),W'-W} = \int_0^1\inpf{\nabla J(W+\tau(W'-W))-\nabla J(W),W'-W}\,d\tau\ge-\int_0^1\tau L\normsp{W'-W}^2\,d\tau = -\frac{L}{2}\normsp{W'-W}^2$, which is \eqref{eq:return-ascent-bound}.
\end{proof}

\section{Proof of Lemma \ref{lemmas:critic-target-lipschitz}}\label{apdx:lemma:opt_lip}
\begin{proof}Fix $W$ and $W'$, let $\ell_2 := \normsp{W-W'}$ and $\epsilon_\pi := \tfrac{1}{2}R_\pi\ell_2$, recall that the ridge parameter satisfies $\lambda>0$, and let $\mathbb{P}_W := \mu_W\otimes\pi_W\otimes\mathbb{P}$ denote the law of $o = (s,a,s')$ under $\pi_W$, such that $A_W = \mathbb{E}_{\mathbb{P}_W}[\phi(s)(\phi(s)-\gamma\phi(s'))^\top]$ and $b_W = \mathbb{E}_{\mathbb{P}_W}[\phi(s)r(s,a)]$.
	
	\emph{Change of the observation law.}~Step~1 of the proof of Lemma~\ref{lemmas:frame-bias-noise}(ii) in Appendix~\ref{apdx:lemma:is_bias}, which holds for any two parameters, gives \eqref{eq:policy-total-variation} and \eqref{eq:kernel-stationary-perturbation}, i.e., $\sup_s\normtv{\pi_W(\cdot|s)-\pi_{W'}(\cdot|s)}\le\epsilon_\pi$ and $\normtv{\mu_W-\mu_{W'}}\le4\tau_{\mix}\epsilon_\pi$.
	For a measurable set ${\cal B}$ of observations, let $p_{\cal B}(s,a) := \mathbb{P}(\{s':(s,a,s')\in {\cal B}\}\,|\,s,a)$ and $f(s) := \int\pi_W(da|s)\,p_{\cal B}(s,a)$, both $[0,1]$-valued.
	Then $\mathbb{P}_W({\cal B})-\mathbb{P}_{W'}({\cal B}) = \int f\,d(\mu_W-\mu_{W'})+\int\mu_{W'}(ds)\int p_{\cal B}(s,a)\,(\pi_W-\pi_{W'})(da|s)$, and, since $|\int f\,d\nu|\le\normtv{\nu}$ for every $[0,1]$-valued $f$ and every signed measure $\nu$ with total mass zero,
	\begin{equation}\label{eq:observation-law-perturbation}
		\normtv{\mathbb{P}_W-\mathbb{P}_{W'}}
		\le \normtv{\mu_W-\mu_{W'}}+\sup\nolimits_s\normtv{\pi_W(\cdot|s)-\pi_{W'}(\cdot|s)}
		\le (4\tau_{\mix}+1)\,\epsilon_\pi.
	\end{equation}
	
	\emph{Change of $v^*$.}~Since $\normsp{\phi(s)}\le1$ gives $\normsp{\phi(s)r(s,a)}\le R_r$ and $\normsp{\phi(s)(\phi(s)-\gamma\phi(s'))^\top}\le1+\gamma$, the bound on $\norm{\int f\,d\nu}$ of Appendix~\ref{apdx:prelim}, applied to measures on observations, gives $\normsp{b_W-b_{W'}}\le2R_r\normtv{\mathbb{P}_W-\mathbb{P}_{W'}}$ and $\normsp{A_W-A_{W'}}\le2(1+\gamma)\normtv{\mathbb{P}_W-\mathbb{P}_{W'}}$.
	Moreover, $\normsp{b_{W'}}\le R_r$, and $x^\top A_Wx\ge0$ for all $x$ (see below \eqref{eq:population-lstd-system}) gives $\lambda\normsp{x}^2\le\normsp{x}\normsp{(A_W+\lambda I_d)x}$, i.e., $\normsp{(A_W+\lambda I_d)^{-1}}\le\nicefrac{1}{\lambda}$, and likewise for $W'$.
	Hence, the resolvent identity $(A_W+\lambda I_d)^{-1}-(A_{W'}+\lambda I_d)^{-1} = (A_W+\lambda I_d)^{-1}(A_{W'}-A_W)(A_{W'}+\lambda I_d)^{-1}$ gives
	\begin{equation*}
		\normsp{v^*(W)-v^*(W')}
		\le \frac{\normsp{b_W-b_{W'}}}{\lambda}+\frac{R_r\normsp{A_W-A_{W'}}}{\lambda^2}
		\le \frac{2R_r(\lambda+1+\gamma)}{\lambda^2}\normtv{\mathbb{P}_W-\mathbb{P}_{W'}}
	\end{equation*}
	and \eqref{eq:observation-law-perturbation} with $\epsilon_\pi = \tfrac{1}{2}R_\pi\ell_2$ induces $\normsp{v^*(W)-v^*(W')}\le L_*\ell_2$.
\end{proof}

\section{Preliminaries and a Markovian Moment Bound}\label{apdx:prelim}
Throughout the appendices, $P_W^t$ denotes the $t$-step kernel of the chain induced by $\pi_W$, $d_{\mix}(t) := \sup_W\sup_s\normtv{P^t_W(s,\cdot)-\mu_W}$, and H\"older's inequality refers to $|\inpf{X,Y}|\le\normnu{X}\normsp{Y}$.
For every finite signed measure $\nu$ on a measurable space ${\cal X}$, with $\normtv{\nu} := \sup_{{\cal B}\subseteq{\cal X}}|\nu({\cal B})|$, and every bounded measurable $f$ on ${\cal X}$ with values in a finite-dimensional normed space, $\norm{\int f\,d\nu}\le\sup\norm{f}\,|\nu|({\cal X})\le2\sup\norm{f}\,\normtv{\nu}$; we use this bound for measures on states, actions, observations, and frames.
By \citep[Sections~4.4--4.5]{Levin2017}, whose arguments do not use the finiteness of the state space, $d_{\mix}$ is nonincreasing and, by Assumption~\ref{assum:is}, satisfies $d_{\mix}(j\,\tau_{\mix})\le2^{-j}$ for all integers $j\ge0$; hence,
\begin{equation}\label{eq:mixing-distance-sum}
	\sum\nolimits_{t\ge0}d_{\mix}(t)\le\tau_{\mix}\sum\nolimits_{j\ge0}2^{-j} = 2\tau_{\mix}.
\end{equation}

The following second-moment bound for averages along the chain replaces the concentration bounds of~\citet[Lemma~A.6]{DorfmanJuly2022} for finite-state chains and holds for general state spaces.

\begin{lemma}\label{lemmas:markov-moment}
	Under Assumption~\ref{assum:is}, let $f$ map observations $o=(s,a,s')$ into a real inner-product space of finite dimension with $\norm{f}\le B$, fix $W$ and a start state $s_0$, let $o_t = (s_t,a_t,s_{t+1})$, $t\ge0$, be generated by $\pi_W$ and $\mathbb{P}$ from $s_0$, and let $m := \mathbb{E}_{\mu_W\otimes\pi_W\otimes\mathbb{P}}[f]$.
	Then, for every $T\ge1$,
	\begin{equation*}
		\mathbb{E}\NORM{\frac{1}{T}\sum_{t=0}^{T-1}f(o_t)-m}^2\le\frac{4B^2(1+4\tau_{\mix})}{T}.
	\end{equation*}
\end{lemma}

\begin{proof}Let $X_t := f(o_t)-m$, such that $\norm{X_t}\le2B$, and let $\bar f(s)$ denote the mean of $f(s,a,s')$ over $a\sim\pi_W(\cdot|s)$ and $s'\sim\mathbb{P}(\cdot|s,a)$, such that $m = \int\bar f\,d\mu_W$.
For $j\ge1$, the Markov property gives $\ee{X_{t+j}\,|\,s_0,a_0,\ldots,a_t,s_{t+1}} = \int\bar f\,d(P_W^{j-1}(s_{t+1},\cdot)-\mu_W)$, whose norm is at most $2B\,d_{\mix}(j-1)$; since $X_t$ is a function of $(s_t,a_t,s_{t+1})$, the tower property gives $|\mathbb{E}\langle X_t,X_{t+j}\rangle|\le4B^2d_{\mix}(j-1)$.
Expanding the square and using $T-j\le T$ and \eqref{eq:mixing-distance-sum} yield
\begin{equation*}
	\mathbb{E}\NORM{\frac{1}{T}\sum_{t=0}^{T-1}X_t}^2
	\le \frac{4B^2}{T}+\frac{8B^2}{T^2}\sum_{j=1}^{T-1}(T-j)\,d_{\mix}(j-1)
	\le \frac{4B^2(1+4\tau_{\mix})}{T}.
\end{equation*}

We complete the proof.
\end{proof}

\section{Proof of Theorem \ref{theorems:lstd-critic-error}}\label{apdx:thm:01}
\begin{proof}Fix $k\ge0$.
We bound the NSI and the statistical errors and then combine them.

\emph{NSI error.}~The safe scaling $c_k^2 = \norm{H_k}_1\norm{H_k}_\infty$ of Section~\ref{sec:critic} satisfies $\normsp{H_k}\le c_k\le\lambda+1+\gamma$, as the theorem requires: the first inequality holds because $\normsp{H}^2\le\norm{H}_1\norm{H}_\infty$ for every square matrix $H$, and the second because every entry of $Y_k^\top\Phi_k$, the inner product of a column of $Y_k$ with norm at most $\nicefrac{(1+\gamma)}{\sqrt{T}}$ and a column of $\Phi_k$ with norm at most $\nicefrac{1}{\sqrt{T}}$, is at most $\nicefrac{(1+\gamma)}{T}$ in absolute value, such that every absolute row and column sum of $H_k$ is at most $\lambda+1+\gamma$.
By the NSI recursion \eqref{eq:newton-schulz-critic}, $E_{k,q+1} = I_T-H_kX_{k,q}\sum_{i=0}^{\kappa_1-1}E_{k,q}^i = I_T-(I_T-E_{k,q})\sum_{i=0}^{\kappa_1-1}E_{k,q}^i = E_{k,q}^{\kappa_1}$, such that $X_{k,0} = \nicefrac{H_k^\top}{c_k^2}$ gives $E_{k,q} = (I_T-\nicefrac{H_kH_k^\top}{c_k^2})^{\kappa_1^q}$, a symmetric matrix with the eigenvalues $(1-\nicefrac{\sigma_i^2(H_k)}{c_k^2})^{\kappa_1^q}\in[0,1)$, since $c_k\ge\normsp{H_k}$ and $H_k$ is non-singular (Appendix~\ref{apdx:bk}).
Hence, $\normsp{E_{k,q}} = \epsilon_E^{\kappa_1^q}$ with $\epsilon_E := 1-\nicefrac{\sigma_{\min}^2(H_k)}{c_k^2}\in[0,1)$.
Since $X_{k,q}-H_k^{-1} = -H_k^{-1}E_{k,q}$, \eqref{eq:newton-schulz-critic} and \eqref{eq:reduced-lstd-solution} give $v_{k,q}-\hat v_k = -\Phi_kH_k^{-1}E_{k,q}r_k$, and the push-through identity $\Phi_kH_k^{-1} = (\hat A_k+\lambda I_d)^{-1}\Phi_k$ in \eqref{eq:reduced-lstd-solution}, \eqref{eq:empirical-lstd-accretivity}, and $\normsp{\Phi_k}\le\normf{\Phi_k}\le1$ give $\normsp{\Phi_kH_k^{-1}}\le\nicefrac{1}{\lambda}$.
Together with $\normsp{r_k}\le R_r$, we obtain, for every realization,
\begin{equation}\label{eq:nsi-critic-error}
	\normsp{v_{k,q}-\hat v_k}
	\le R_v\,\epsilon_E^{\kappa_1^q}\le R_v\exp(-C_E\kappa_1^q)
\end{equation}
where the last inequality holds because $\epsilon_E\le\rho_E = \exp(-C_E)$: indeed, every column of $Y_k$ has norm at most $\nicefrac{(1+\gamma)}{\sqrt{T}}$, such that $\normsp{Y_k^\top\Phi_k}\le\normf{Y_k}\normsp{\Phi_k}\le1+\gamma$ gives $\sigma_{\max}(H_k)\le\lambda+1+\gamma$, such that the range of $c_k$ in Theorem~\ref{theorems:lstd-critic-error} is nonempty, and the Woodbury identity $H_k^{-1} = \lambda^{-1}[I_T-Y_k^\top(\hat A_k+\lambda I_d)^{-1}\Phi_k]$ and $\normsp{(\hat A_k+\lambda I_d)^{-1}}\le\nicefrac{1}{\lambda}$ give $\sigma_{\min}(H_k)\ge\nicefrac{\lambda^2}{(\lambda+1+\gamma)}$, such that $c_k\le\lambda+1+\gamma$ induces $\epsilon_E\le1-\nicefrac{\sigma_{\min}^2(H_k)}{(\lambda+1+\gamma)^2}\le\rho_E$.

\emph{Statistical error.}~Since $\hat v_k = (\hat A_k+\lambda I_d)^{-1}\hat b_k$ and $(A_k+\lambda I_d)v_k^* = b_k$, we have $\hat v_k-v_k^* = (\hat A_k+\lambda I_d)^{-1}[(\hat b_k-b_k)-(\hat A_k-A_k)v_k^*]$, where $\normsp{(\hat A_k+\lambda I_d)^{-1}}\le\nicefrac{1}{\lambda}$ and $\normsp{v_k^*}\le\nicefrac{R_r}{\lambda}$.
For $k\ge1$, conditionally on ${\cal F}_{k-1}$, the frame $O_k$ is generated by $\pi_{W_k}$ from $s_{k,0}$, its law is measurable in $(W_k,s_{k,0})$ by Assumption~\ref{assum:01}, and $v_k^* = v^*(W_k)$ is ${\cal F}_{k-1}$-measurable because $v^*$ is continuous (Lemma~\ref{lemmas:critic-target-lipschitz}).
Lemma~\ref{lemmas:markov-moment}, applied conditionally on ${\cal F}_{k-1}$ to $f_b(o) := \phi(s)r(s,a)$ with $B = R_r$ and to $f_A(o) := \phi(s)(\phi(s)-\gamma\phi(s'))^\top$ in the Frobenius norm with $B = 1+\gamma$, bounds the errors of $\hat b_k$ and of the average $\bar A_k := \frac{1}{T}\sum_{t<T}f_A(o_{k,t})$ over the frame, which differs from the pathwise LSTD matrix only in the dropped bootstrap, $\hat A_k = \bar A_k+\frac{\gamma}{T}\phi_{k,T-1}\phi_{k,T}^\top$.
Hence, $\normsp{\cdot}\le\normf{\cdot}$, Minkowski's inequality, and $\nicefrac{\gamma}{T}\le\gamma\sqrt{\nicefrac{(1+4\tau_{\mix})}{T}}$ give
\begin{equation}\label{eq:markov-sample-moments}
	\ee{\normsp{\hat{b}_k-b_k}^2\,|\,{\cal F}_{k-1}}
	\le \frac{4R_r^2(1+4\tau_{\mix})}{T},
	\quad
	\ee{\normsp{\hat{A}_k-A_k}^2\,|\,{\cal F}_{k-1}}
	\le \frac{(2+3\gamma)^2(1+4\tau_{\mix})}{T}
\end{equation}
for every frame, since $\tau_{\mix}$ is uniform over the policies and the start states; for the warm-up frame, the bounds hold for every start state and hence for $s_{0,0}\sim\rho$.
Hence, $(x+y)^2\le2x^2+2y^2$ and $1+\nicefrac{(1+\nicefrac{3\gamma}{2})^2}{\lambda^2}\le\nicefrac{(\lambda+1+\nicefrac{3\gamma}{2})^2}{\lambda^2}$ give
\begin{equation}\label{eq:lstd-sampling-error}
	\ee{\normsp{\hat v_k-v_k^*}^2\,|\,{\cal F}_{k-1}}
	\le \frac{8R_r^2(1+4\tau_{\mix})}{\lambda^2T}\Big(1+\frac{(1+\nicefrac{3\gamma}{2})^2}{\lambda^2}\Big)
	\le \frac{4(1+4\tau_{\mix})\,C_v^2}{T}.
\end{equation}

\emph{Combination.}~Since $v_k = v_{k,P_1}$, \eqref{eq:nsi-critic-error} with $q = P_1$, \eqref{eq:lstd-sampling-error}, and the conditional Minkowski inequality give $\sqrt{\ee{\normsp{v_k-v_k^*}^2\,|\,{\cal F}_{k-1}}}\le R_v\exp(-C_E\kappa_1^{P_1})+2C_v\sqrt{\nicefrac{(1+4\tau_{\mix})}{T}} = \epsilon_{\rm lstd}$, which is \eqref{eq:lstd-critic-error}.
The critic is an NSI iterate, whose norm is at most $R_v$ by Lemma~\ref{lemmas:critic-gradient-bounds}(i); the mean-square form follows from $(x+y)^2\le2x^2+2y^2$, and, for $k = 0$, ${\cal F}_{-1}$ is trivial and the bounds hold for every start state and hence for $s_{0,0}\sim\rho$.
\end{proof}

\section{Proof of Lemma \ref{lemmas:critic-gradient-bounds}}\label{apdx:lemma:support}
\begin{proof}
\emph{Part (i).}~By the proof of Theorem~\ref{theorems:lstd-critic-error}, $E_{k,q}$ is symmetric with eigenvalues in $[0,1)$, such that $\normsp{I_{T}-E_{k,q}}\le1$, and $\normsp{\Phi_kH_k^{-1}}\le\nicefrac{1}{\lambda}$.
Since $v_{k,q} = \Phi_kX_{k,q}r_k$ with $X_{k,q} = H_k^{-1}(I_{T}-E_{k,q})$ and $\normsp{r_k}\le R_r$, we obtain $\normsp{v_{k,q}}\le\nicefrac{R_r}{\lambda} = R_v$.
The critic $v_k$ is the iterate $v_{k,P_1}$, such that $\normsp{v_k}\le R_v$, and the Cauchy-Schwarz inequality with $\normsp{\phi(s)}\le1$ gives $\sup_s|\phi^\top(s)v_k|\le R_v$, i.e., $v_k$ is admissible.

\emph{Part (ii).}~For an admissible $v$ and an observation $o = (s,a,s')$, the bounds $|r(s,a)|\le R_r$, $|\phi^\top(s')v|\le R_v$, and $|\phi^\top(s)v|\le R_v$ give $|\delta_v(o)|\le R_r+(1+\gamma)R_v$, such that \eqref{eq:bootstrapped-advantage} gives $|\hat\delta_t(v)|\le\sum_{l\ge0}\gamma^l(R_r+(1+\gamma)R_v) = R_A$.

\emph{Part (iii).}~Since the weights $\nicefrac{1}{T}$ in \eqref{eq:frame-policy-gradient} sum to one, the triangle inequality, part (ii), and \eqref{eq:score-bound} give $\normnu{\hat g(W,v;O)}\le R_AR_\pi = R_g$; since the critic $v_{k-1}$ used in $G_k$ is admissible by part (i), $\normnu{G_k}\le R_g$.
For the PG, the score has zero mean by dominated convergence and \eqref{eq:score-bound}, such that the per-state PG satisfies
\begin{equation*}
	g_W(s) := \mathbb{E}_{a\sim\pi_W}[(Q_W(s,a)-V_W(s))\nabla\log\pi_W(a|s)] = \mathbb{E}_{a\sim\pi_W}[Q_W(s,a)\nabla\log\pi_W(a|s)]
\end{equation*}
and $|Q_W|\le\nicefrac{R_r}{(1-\gamma)}$ and \eqref{eq:score-bound} give $\normnu{g_W(s)}\le\nicefrac{R_rR_\pi}{(1-\gamma)}$.
Since \eqref{eq:policy-gradient} gives $\nabla J(W) = \mathbb{E}_{s\sim\zeta_W}[g_W(s)]$ and $\zeta_W$ is a probability measure, we obtain $\normnu{\nabla J(W)}\le\nicefrac{R_rR_\pi}{(1-\gamma)}\le R_AR_\pi = R_g$; the Frobenius bounds follow from $\normf{\cdot}\le\normnu{\cdot}$.
\end{proof}

\section{Proof of Lemma \ref{lemmas:importance-correction}}\label{apdx:lemma:is_corr}
\begin{proof}\emph{Part (i).}~Write $O = \{(s_t,a_t)\}_{t=0}^{T-1}\cup\{s_T\}$, abbreviate $\nabla_{W'',t} := \nabla\log\pi_{W''}(a_t|s_t)$, and let $W_\tau := W+\tau(W'-W)$, $\tau\in[0,1]$.
By Assumption~\ref{assum:01}, $\tau\mapsto\log\pi_{W_\tau}(a_t|s_t)$ is differentiable with derivative $\inpf{\nabla_{W_\tau,t},W'-W}$, such that H\"older's inequality and \eqref{eq:score-bound} give $|\log\nicefrac{\pi_{W'}(a_t|s_t)}{\pi_W(a_t|s_t)}|\le R_\pi\ell_2$; summing over $t<T$ induces $|\log w|\le TR_\pi\ell_2\le1$.
Since $\exp(x)-1\le(\exp(1)-1)x$ for $x\in[0,1]$ by convexity and $1-\exp(-x)\le x$ for $x\ge0$, $|w-1|\le(\exp(1)-1)TR_\pi\ell_2$.

\emph{Part (ii).}~Fix $k\ge2$ and abbreviate $v := v_{k-1}$.
Conditionally on ${\cal F}_{k-1}$, the law of $O_k$ is the product of the point mass at the ${\cal F}_{k-1}$-measurable start state $s_{k,0}$, the policy kernels $\pi_k(\cdot|s_t)$, and the transition kernels $\mathbb{P}(\cdot|s_t,a_t)$, $t<T$, and each reward $r_t = r(s_t,a_t)$ is a function of the state-action pair.
By the positivity clause of Assumption~\ref{assum:01}, the policy kernels of $\pi_{k-1}$ and $\pi_k$ are mutually absolutely continuous with density ratio $\nicefrac{\pi_{k-1}(a|s)}{\pi_k(a|s)}$; since the start state and the transition kernels are common to both laws, the Radon--Nikodym derivative of the law of a $T$-step trajectory under $\pi_{k-1}$ with respect to that under $\pi_k$ is $w_k$.
Since $W_{k-1}$, $W_k$, $v$, and $s_{k,0}$ are ${\cal F}_{k-1}$-measurable, this change of measure gives $\ee{w_kf(O_k;\theta)\,|\,{\cal F}_{k-1}} = \mathbb{E}_{O\sim\pi_{k-1}}[f(O;\theta)]$ for every bounded $f$ that is jointly measurable in the frame and an ${\cal F}_{k-1}$-measurable parameter $\theta$, where $O\sim\pi_{k-1}$ denotes a $T$-step trajectory of the chain induced by $\pi_{k-1}$ from $s_0 = s_{k,0}$.
We take $f(O;\theta) = \hat g(W_{k-1},v;O)$ with $\theta = (W_{k-1},v)$, which is bounded by $R_g$ (Lemma~\ref{lemmas:critic-gradient-bounds}(iii)), and obtain $\ee{\widetilde G_k\,|\,{\cal F}_{k-1}} = \mathbb{E}_{O\sim\pi_{k-1}}[\hat g(W_{k-1},v;O)] = h(W_{k-1};\omega_k)$ by \eqref{eq:actor-mean-field}.

\emph{Part (iii).}~Since $\hat\delta_t(v)$ does not depend on the actor parameter, \eqref{eq:frame-policy-gradient} gives
\begin{equation*}
	\hat g(W,v;O)-w\,\hat g(W',v;O)
	= \frac{1}{T}\sum_{t=0}^{T-1}\hat\delta_t(v)\big[\nabla_{W,t}-\nabla_{W',t}\big]
	+ (1-w)\,\hat g(W',v;O).
\end{equation*}

By $|\hat\delta_t(v)|\le R_A$ (Lemma~\ref{lemmas:critic-gradient-bounds}(ii)) and \eqref{eq:score-lipschitz}, the first term is at most $R_AL_\pi\ell_2$ in the nuclear norm; by $\normnu{\hat g(W',v;O)}\le R_AR_\pi$ (Lemma~\ref{lemmas:critic-gradient-bounds}(iii)) and part (i), the second term is at most $(\exp(1)-1)TR_\pi^2R_A\ell_2$, and the two bounds sum to $C_F\ell_2$.
The Frobenius bound follows from $\normf{\cdot}\le\normnu{\cdot}$.
Finally, with $W = W_k$, $W' = W_{k-1}$, $O = O_k$, and $v = v_{k-1}$, the weight $w$ is $w_k$ and, by \eqref{eq:frame-policy-gradient} and \eqref{eq:importance-corrected-gradient}, the difference in part (iii) is $G_k-\widetilde G_k$, such that parts (i) and (iii) with $\ell_2\le\bar\ell_2$ give the last claim.
\end{proof}

\section{Proof of Lemma \ref{lemmas:frame-bias-noise}}\label{apdx:lemma:is_bias}
We use the notation of Appendix~\ref{apdx:prelim}.
For every $W$ and every signed measure $\nu$ on ${\cal S}$ with $\nu({\cal S}) = 0$, we write $\nu = \normtv{\nu}(p-q)$ with probability measures $p$ and $q$, and $\normtv{pP^t_W-qP^t_W}\le\sup_{s,s'}\normtv{P^t_W(s,\cdot)-P^t_W(s',\cdot)}\le\tilde d_{\mix}(t) := \min\{1,2d_{\mix}(t)\}$ gives the contraction $\normtv{\nu P^t_W}\le\tilde d_{\mix}(t)\normtv{\nu}$; moreover, \eqref{eq:mixing-distance-sum} gives $\sum_{t\ge0}\tilde d_{\mix}(t)\le4\tau_{\mix}$.
We prove parts (i) and (ii) in the nuclear norm, which implies the Frobenius norm since $\normf{\cdot}\le\normnu{\cdot}$.
For frame $k$, conditionally on ${\cal F}_{k-1}$, the frame $O_k$ is a $T$-step trajectory of the chain induced by $\pi_k$ from $s_{k,0}$.

\emph{Part (i).}~Fix $W$ and an admissible $\omega = (v,s_0)$, let $O = \{(s_t,a_t)\}_{t=0}^{T-1}\cup\{s_T\}$ be a $T$-step trajectory of the chain induced by $\pi_W$ from $s_0$, such that all expectations below are conditional on $s_0$, abbreviate $\nabla_t := \nabla\log\pi_W(a_t|s_t)$, and let $\hat\delta^\circ_t$ denote $\hat\delta_t$ in \eqref{eq:bootstrapped-advantage} with $V_W$ in place of the critic.
In the telescoped form \eqref{eq:bootstrapped-advantage}, the rewards cancel, such that $\hat\delta_t(v)-\hat\delta^\circ_t = \gamma^{T-t}\epsilon_v(s_T)-\epsilon_v(s_t)$.
Facts (i) and (ii) below \eqref{eq:policy-gradient} hold for any number of summands and, by the Markov property, conditionally on ${\cal H}_t := (s_0,a_0,\ldots,s_t,a_t)$, such that $\ee{\hat\delta^\circ_t\,|\,{\cal H}_t} = Q_W(s_t,a_t)-V_W(s_t)$; since $\nabla_t$ is a function of $(s_t,a_t)$ and $a_t\sim\pi_W(\cdot|s_t)$ given $s_0,a_0,\ldots,s_t$, the tower property gives $\ee{\hat\delta^\circ_t\nabla_t} = \ee{g_W(s_t)}$ with $g_W$ of Appendix~\ref{apdx:lemma:support}, and, since $v$ is fixed and the score has zero mean, $\ee{\epsilon_v(s_t)\nabla_t} = 0$.
Therefore, since $g(W) = \mathbb{E}_{s\sim\mu_W}[g_W(s)]$, \eqref{eq:frame-policy-gradient} and \eqref{eq:actor-mean-field} split $D$ into a transient part and a critic part,
\begin{equation}\label{eq:frame-bias-decomposition}
	D(W;\omega)
	= \underbrace{\frac{1}{T}\sum_{t=0}^{T-1}\Big(\ee{g_W(s_t)}-\mathbb{E}_{s\sim\mu_W}[g_W(s)]\Big)}_{=:\,D^{\rm tr}(W;s_0)}
	+ \underbrace{\frac{1}{T}\sum_{t=0}^{T-1}\gamma^{T-t}\,\ee{\epsilon_v(s_T)\nabla_t}}_{=:\,D^{\rm cr}(W;\omega)}.
\end{equation}
Since $\normnu{g_W(s)}\le\nicefrac{R_rR_\pi}{(1-\gamma)}$ (Appendix~\ref{apdx:lemma:support}) and $\ee{g_W(s_t)} = \int g_W\,dP^t_W(s_0,\cdot)$, \eqref{eq:mixing-distance-sum} gives $\normnu{D^{\rm tr}(W;s_0)}\le\frac{2R_rR_\pi}{(1-\gamma)T}\sum_{t<T}d_{\mix}(t)\le\nicefrac{4R_rR_\pi\tau_{\mix}}{[(1-\gamma)T]}$, and \eqref{eq:score-bound} and $\sum_{t<T}\gamma^{T-t}\le\nicefrac{\gamma}{(1-\gamma)}$ give $\normnu{D^{\rm cr}(W;\omega)}\le\nicefrac{\gamma R_\pi\norm{\epsilon_v}_\infty}{[(1-\gamma)T]}$, which proves the first claim.
For frame $k$, $\ee{G_k\,|\,{\cal F}_{k-1}} = h(W_k;\omega_k)$ (Section~\ref{sec:spg-analysis}), where $\omega_k$ is admissible by Lemma~\ref{lemmas:critic-gradient-bounds}(i) and $\epsilon_{v_{k-1}} = \epsilon_k$ at $W = W_k$; hence, $\normnu{B(W_k;\omega_k)}\le\normnu{\bar B(W_k)}+\normnu{D(W_k;\omega_k)}$, the definition of $\epsilon_{\rm dist}$, and Minkowski's inequality with $\sqrt{\ee{\norm{\epsilon_k}_\infty^2}}\le\epsilon_{\rm crit}$ yield $\sqrt{\ee{\normnu{B(W_k;\omega_k)}^2}}\le\epsilon_{\rm bias}$.

\emph{Part (ii).}~Fix $W$, $W'$, and an admissible $\omega = (v,s_0)$, let $\ell_2 := \normsp{W-W'}$, and let $C_D^R := \nicefrac{\tau_{\mix}^2R_AR_\pi^2}{(1-\gamma)^2}$ and $C_D^S := \nicefrac{\tau_{\mix}^2R_AL_\pi}{(1-\gamma)^2}$, such that $D_F = \nicefrac{(46C_D^R+5C_D^S)}{T}$; recall that $\tau_{\mix}\ge1$, $\nicefrac{R_r}{(1-\gamma)}\le R_A$, and $\norm{\phi^\top v-V_{W''}}_\infty\le\bar\epsilon := R_v+\nicefrac{R_r}{(1-\gamma)}\le R_A$ for every $W''$.
By \eqref{eq:frame-bias-decomposition}, it suffices to bound the changes of $D^{\rm tr}$ and $D^{\rm cr}$, for which we use, for bounded matrix-valued $f,f'$ and finite signed measures $\nu,\nu'$ with $(\nu-\nu')({\cal S}) = 0$,
\begin{equation}\label{eq:integral-perturbation}
	\normnu{\int f\,d\nu-\int f'\,d\nu'}\le\sup\nolimits_s\normnu{f(s)-f'(s)}\,|\nu|({\cal S})+2\sup\nolimits_s\normnu{f'(s)}\,\normtv{\nu-\nu'}.
\end{equation}

\emph{Step 1: policy and chain.}~Along $W_\tau := W'+\tau(W-W')$, $\tau\in[0,1]$, we have $\frac{d}{d\tau}\pi_{W_\tau}(a|s) = \pi_{W_\tau}(a|s)\inpf{\nabla\log\pi_{W_\tau}(a|s),W-W'}$, where H\"older's inequality and \eqref{eq:score-bound} bound the inner product by $R_\pi\ell_2$; integrating over $\tau$ and $a$ induces
\begin{equation}\label{eq:policy-total-variation}
	\sup\nolimits_s\normtv{\pi_W(\cdot|s)-\pi_{W'}(\cdot|s)}\le\epsilon_\pi := \tfrac{1}{2}R_\pi\ell_2
\end{equation}
and the same bound for the state kernels $P_W(s,\cdot) := \int\pi_W(da|s)\,\mathbb{P}(\cdot|s,a)$, such that $\normtv{p(P_W-P_{W'})}\le\epsilon_\pi$ for every probability measure $p$ and $\normtv{\nu(P_W-P_{W'})}\le2\epsilon_\pi\normtv{\nu}$ if $\nu({\cal S}) = 0$.
The identity $P^t_W-P^t_{W'} = \sum_{i=0}^{t-1}P^i_W(P_W-P_{W'})P^{t-1-i}_{W'}$ and the contraction give, for all $s$ and $t$,
\begin{equation}\label{eq:kernel-stationary-perturbation}
	\normtv{P^t_W(s,\cdot)-P^t_{W'}(s,\cdot)}\le\epsilon_\pi\sum_{j=0}^{t-1}\tilde d_{\mix}(j)\le4\tau_{\mix}\epsilon_\pi,
	\qquad
	\normtv{\mu_W-\mu_{W'}}\le4\tau_{\mix}\epsilon_\pi
\end{equation}
where the second bound follows from the first, since, for every $t$, the distance $\normtv{\mu_W-\mu_{W'}}$ is at most $\normtv{P^t_W(s,\cdot)-P^t_{W'}(s,\cdot)}+2d_{\mix}(t)$, and $d_{\mix}(t)\to0$ as $t\to\infty$; Step~1 holds for any two parameters $W$ and $W'$.

\emph{Step 2: value function and per-state gradient.}~By the Bellman equation, $V_W-V_{W'} = \sum_{t\ge0}\gamma^tP^t_Wu$ with $u(s) := (\mathbb{E}_{a\sim\pi_W(\cdot|s)}-\mathbb{E}_{a\sim\pi_{W'}(\cdot|s)})[Q_{W'}(s,a)]$, where $|u(s)|\le\nicefrac{2\epsilon_\pi R_r}{(1-\gamma)}$ by \eqref{eq:policy-total-variation}; hence, $\norm{V_W-V_{W'}}_\infty\le L_V\ell_2$ and $|Q_W-Q_{W'}|\le\gamma L_V\ell_2$ with $L_V := \nicefrac{R_\pi R_r}{(1-\gamma)^2}\le\nicefrac{R_\pi R_A}{(1-\gamma)}$.
Splitting $g_W(s)-g_{W'}(s)$, with $g_W(s) = \mathbb{E}_{a\sim\pi_W(\cdot|s)}[Q_W(s,a)\nabla\log\pi_W(a|s)]$, into the changes of the policy, of $Q$, and of the score, \eqref{eq:policy-total-variation}, \eqref{eq:score-bound}, and \eqref{eq:score-lipschitz} give
\begin{equation}\label{eq:state-gradient-perturbation}
	\sup\nolimits_s\normnu{g_W(s)-g_{W'}(s)}
	\le \frac{R_r\big(R_\pi^2+(1-\gamma)L_\pi\big)}{(1-\gamma)^2}\ell_2
	\le \frac{R_A(R_\pi^2+L_\pi)}{1-\gamma}\ell_2.
\end{equation}

\emph{Step 3: transient part.}~Let $\nu_t := P^t_W(s_0,\cdot)-\mu_W$ and $\nu'_t := P^t_{W'}(s_0,\cdot)-\mu_{W'}$, such that $D^{\rm tr}(W;s_0) = \frac{1}{T}\sum_{t<T}\int g_W\,d\nu_t$ and $\normtv{\nu_t}\le d_{\mix}(t)$.
Since $\mu_W$ and $\mu_{W'}$ are invariant, $\nu_{t+1} = \nu_tP_W$ and $\nu'_{t+1} = \nu'_tP_{W'}$, such that $\nu_t-\nu'_t = (\mu_{W'}-\mu_W)P^t_{W'}+\sum_{i=0}^{t-1}\nu_i(P_W-P_{W'})P^{t-1-i}_{W'}$, where every summand has zero total mass; hence, \eqref{eq:kernel-stationary-perturbation} and the contraction give $\normtv{\nu_t-\nu'_t}\le\epsilon_\pi[4\tau_{\mix}\tilde d_{\mix}(t)+\sum_{i=0}^{t-1}2d_{\mix}(i)\,\tilde d_{\mix}(t-1-i)]$, and \eqref{eq:mixing-distance-sum} induces $\sum_{t\ge0}\normtv{\nu_t-\nu'_t}\le32\tau_{\mix}^2\epsilon_\pi$.
By \eqref{eq:integral-perturbation} with $|\nu_t|({\cal S})\le2d_{\mix}(t)$, \eqref{eq:state-gradient-perturbation}, \eqref{eq:mixing-distance-sum}, and $\normnu{g_{W'}(s)}\le\nicefrac{R_rR_\pi}{(1-\gamma)}$,
\begin{equation*}
	\begin{aligned}
		&\normnu{D^{\rm tr}(W;s_0)-D^{\rm tr}(W';s_0)}\\
		&\quad\le \frac{4\tau_{\mix}R_A(R_\pi^2+L_\pi)\ell_2}{(1-\gamma)T}+\frac{32\tau_{\mix}^2R_rR_\pi^2\ell_2}{(1-\gamma)T}\\
		&\quad\le \frac{(36C_D^R+4C_D^S)\ell_2}{T}.
	\end{aligned}
\end{equation*}

\emph{Step 4: critic part.}~Let $\epsilon_W := \phi^\top v-V_W$ denote $\epsilon_v$ as a function of $W$ for the fixed $v$, such that $D^{\rm cr}(W;\omega) = \frac{1}{T}\sum_{t<T}\gamma^{T-t}\ee{\epsilon_W(s_T)\nabla_t}$ with $\nabla_t$ and the expectation taken under $\pi_W$, and fix $t<T$.
By the Markov property, $\ee{\epsilon_W(s_T)\nabla_t} = \int\varpi_W\,dP^t_W(s_0,\cdot)$ with $\varpi_W(s) := \mathbb{E}_{a\sim\pi_W(\cdot|s)}[\nabla\log\pi_W(a|s)F_W(s,a)]$ and $F_W(s,a) := \int\mathbb{P}(ds'|s,a)(P^{T-t-1}_W\epsilon_W)(s')$, where $|F_W|\le\bar\epsilon$ and $\normnu{\varpi_W}\le R_\pi\bar\epsilon$.
Since $\epsilon_W-\epsilon_{W'} = V_{W'}-V_W$, Step~2 and \eqref{eq:kernel-stationary-perturbation} give $|F_W-F_{W'}|\le(L_V+4\tau_{\mix}R_\pi\bar\epsilon)\ell_2$, and splitting $\varpi_W-\varpi_{W'}$ into the changes of the policy, of the score, and of $F$ bounds $\sup_s\normnu{\varpi_W(s)-\varpi_{W'}(s)}$ by $[\bar\epsilon(R_\pi^2+L_\pi)+R_\pi L_V+4\tau_{\mix}R_\pi^2\bar\epsilon]\,\ell_2$.
Hence, by \eqref{eq:integral-perturbation} and \eqref{eq:kernel-stationary-perturbation}, the change of $\ee{\epsilon_W(s_T)\nabla_t}$ is at most
\begin{equation*}
	\begin{split}
	\Big[\bar\epsilon\big((8\tau_{\mix}+1)R_\pi^2+L_\pi\big)+R_\pi L_V\Big]\ell_2
	&\le \Big[9\tau_{\mix}R_AR_\pi^2+\frac{R_\pi^2R_A}{1-\gamma}+R_AL_\pi\Big]\ell_2 \\
	&\le (1-\gamma)\big(10C_D^R+C_D^S\big)\ell_2.
	\end{split}
\end{equation*}

Moreover, the inequality $\sum_{t<T}\gamma^{T-t}\le\nicefrac{1}{(1-\gamma)}$ induces $\normnu{D^{\rm cr}(W;\omega)-D^{\rm cr}(W';\omega)}\le\nicefrac{(10C_D^R+C_D^S)\ell_2}{T}$.
Steps~3 and~4 give $\normnu{D(W;\omega)-D(W';\omega)}\le\nicefrac{(46C_D^R+5C_D^S)\ell_2}{T} = D_F\ell_2$.

\emph{Part (iii).}~We begin by bounding the mean-square error of the oracle estimator obtained by replacing the fitted critic with the exact value function $V_W$.

\begin{lemma}\label{lemmas:oracle-noise}
	Under Assumptions~\ref{assum:01} and~\ref{assum:is}, fix $W$ and a start state $s_0$, let $O$ be a $T$-step trajectory of the chain induced by $\pi_W$ from $s_0$, and let $\hat g^{\circ}(W;O)$ denote \eqref{eq:frame-policy-gradient} with $V_W$ in place of $\phi^\top v$ in \eqref{eq:bootstrapped-advantage}.
	Then $\ee{\normf{\hat g^{\circ}(W;O)-g(W)}^2}\le\nicefrac{(V^\circ)^2}{T}$.
\end{lemma}

\begin{proof}Abbreviate $Q := Q_W$, $V := V_W$, and $\nabla_t := \nabla\log\pi_W(a_t|s_t)$, and let $A_t := Q(s_t,a_t)-V(s_t)$ and $\xi_{t+1} := r(s_t,a_t)+\gamma V(s_{t+1})-Q(s_t,a_t) = \gamma[V(s_{t+1})-\int\mathbb{P}(ds'|s_t,a_t)V(s')]$, such that the TD error of $V$ is $\delta^\circ_t = A_t+\xi_{t+1}$.
The telescoped form \eqref{eq:bootstrapped-advantage} with $V$ in place of $\phi^\top v$ is $\hat\delta^{\circ}_t = \sum_{i=t}^{T-1}\gamma^{i-t}\delta^\circ_i$, and rearranging the double sum by the index of $A_i$ and $\xi_i$ gives
\begin{equation*}
	T\,\hat g^{\circ}(W;O)
	= \sum_{t=0}^{T-1}A_t\nabla_t+\underbrace{\sum_{i=1}^{T-1}A_iZ_i+\sum_{i=1}^{T}\xi_iZ'_i}_{=:\,S},
	\qquad
	Z_i := \sum_{t<i}\gamma^{i-t}\nabla_t,
	\enspace
	Z'_i := \sum_{t<i}\gamma^{i-1-t}\nabla_t.
\end{equation*}

\emph{Average part.}~The function $f(s,a) := (Q(s,a)-V(s))\nabla\log\pi_W(a|s)$ satisfies $\normf{f}\le\normnu{f}\le\nicefrac{2R_rR_\pi}{(1-\gamma)}$, since $|Q|,|V|\le\nicefrac{R_r}{(1-\gamma)}$, and has the mean $g(W)$ under $\mu_W\otimes\pi_W$ by \eqref{eq:actor-mean-field}; hence, Lemma~\ref{lemmas:markov-moment} gives $\ee{\normf{\frac{1}{T}\sum_{t<T}A_t\nabla_t-g(W)}^2}\le\nicefrac{16R_r^2R_\pi^2(1+4\tau_{\mix})}{[(1-\gamma)^2T]}$.

\emph{Martingale part.}~Let the $\sigma$-fields ${\cal G}_{2i} := \sigma(s_0,a_0,\ldots,a_{i-1},s_i)$ and ${\cal G}_{2i+1} := \sigma({\cal G}_{2i},a_i)$ reveal the states and the actions one at a time.
Then the matrix $\xi_iZ'_i$ is ${\cal G}_{2i}$-measurable with $\ee{\xi_iZ'_i\,|\,{\cal G}_{2i-1}} = Z'_i\,\ee{\xi_i\,|\,s_{i-1},a_{i-1}} = 0$ by the Markov property, since $Z'_i$ is ${\cal G}_{2i-1}$-measurable, and $A_iZ_i$ is ${\cal G}_{2i+1}$-measurable with $\ee{A_iZ_i\,|\,{\cal G}_{2i}} = Z_i\,\mathbb{E}_{a\sim\pi_W(\cdot|s_i)}[Q(s_i,a)-V(s_i)] = 0$, since $Z_i$ is ${\cal G}_{2i}$-measurable.
Hence, $S$ is a sum of martingale differences, which are orthogonal in the Frobenius inner product, and
\begin{equation*}
	\ee{\normf{S}^2} = \sum_{i=1}^{T-1}\ee{\normf{Z_i}^2\,\ee{A_i^2\,|\,{\cal G}_{2i}}}+\sum_{i=1}^{T}\ee{\normf{Z'_i}^2\,\ee{\xi_i^2\,|\,{\cal G}_{2i-1}}}
	\le \frac{2T\gamma^2R_r^2R_\pi^2}{(1-\gamma)^4}
\end{equation*}
since $\ee{A_i^2\,|\,{\cal G}_{2i}}$ is the variance of $Q(s_i,a)$ over $a\sim\pi_W(\cdot|s_i)$, which is at most $\nicefrac{R_r^2}{(1-\gamma)^2}$, $\ee{\xi_i^2\,|\,{\cal G}_{2i-1}}\le\nicefrac{\gamma^2R_r^2}{(1-\gamma)^2}$ likewise, $\normf{Z_i}\le\nicefrac{\gamma R_\pi}{(1-\gamma)}$, and $\normf{Z'_i}\le\nicefrac{R_\pi}{(1-\gamma)}$.
Minkowski's inequality combines the two parts into
\begin{equation*}
	\sqrt{\ee{\normf{\hat g^{\circ}(W;O)-g(W)}^2}}\le\frac{4R_rR_\pi\sqrt{1+4\tau_{\mix}}}{(1-\gamma)\sqrt{T}}+\frac{\sqrt2\,\gamma R_rR_\pi}{(1-\gamma)^2\sqrt{T}} = \frac{V^\circ}{\sqrt{T}}.
\end{equation*}

We complete the proof.
\end{proof}

\emph{Oracle noise.}~Conditionally on ${\cal F}_{k-1}$, $O_k$ is a $T$-step trajectory of the chain induced by $\pi_{W_k}$ from $s_{k,0}$, such that Lemma~\ref{lemmas:oracle-noise} gives $\ee{\normf{G_k^\circ-g(W_k)}^2\,|\,{\cal F}_{k-1}}\le\nicefrac{(V^\circ)^2}{T}$; since $g(W_k)$ is ${\cal F}_{k-1}$-measurable and the conditional mean minimizes the conditional mean-square distance, this proves the second claim of part (iii).

\emph{Critic noise.}~Let $\nabla_{k,t} := \nabla\log\pi_k(a_{k,t}|s_{k,t})$.
By the telescoped form \eqref{eq:bootstrapped-advantage} with $v = v_{k-1}$ and $W = W_k$, $G_k-G_k^\circ = \Delta^{\rm b}_k-\Delta^{\rm s}_k$ with
\begin{equation*}
	\Delta_k^{\rm b} := \frac{1}{T}\sum_{t=0}^{T-1}\gamma^{T-t}\epsilon_k(s_{k,T})\nabla_{k,t},
	\qquad
	\Delta_k^{\rm s} := \frac{1}{T}\sum_{t=0}^{T-1}\epsilon_k(s_{k,t})\nabla_{k,t}.
\end{equation*}
For the baseline part, let ${\cal F}_{k,t}$ extend ${\cal F}_{k-1}$ by $s_{k,0},a_{k,0},\ldots,a_{k,t-1},s_{k,t}$ and $\Xi_t := \epsilon_k(s_{k,t})\nabla_{k,t}$; since $\epsilon_k$ is ${\cal F}_{k-1}$-measurable and $a_{k,t}\sim\pi_k(\cdot|s_{k,t})$ given ${\cal F}_{k,t}$, $\ee{\Xi_t\,|\,{\cal F}_{k,t}} = 0$, and, for $t<t'$, $\Xi_t$ is ${\cal F}_{k,t'}$-measurable, such that $\ee{\inpf{\Xi_t,\Xi_{t'}}\,|\,{\cal F}_{k-1}} = 0$.
Hence, $\ee{\Delta^{\rm s}_k\,|\,{\cal F}_{k-1}} = 0$ and $\ee{\normf{\Delta_k^{\rm s}}^2\,|\,{\cal F}_{k-1}} = \frac{1}{T^2}\sum_{t<T}\ee{\normf{\Xi_t}^2\,|\,{\cal F}_{k-1}}\le\nicefrac{R_\pi^2\norm{\epsilon_k}_\infty^2}{T}$.
For the bootstrap part, \eqref{eq:score-bound} and $\sum_{t<T}\gamma^{T-t}\le\nicefrac{\gamma}{(1-\gamma)}$ give $\normf{\Delta_k^{\rm b}}\le\normnu{\Delta_k^{\rm b}}\le\nicefrac{\gamma R_\pi\norm{\epsilon_k}_\infty}{[(1-\gamma)T]}$ for every realization, and, since centering does not increase the conditional second moment, $\ee{\normf{\Delta^{\rm b}_k-\ee{\Delta^{\rm b}_k\,|\,{\cal F}_{k-1}}}^2\,|\,{\cal F}_{k-1}}\le\nicefrac{\gamma^2R_\pi^2\norm{\epsilon_k}_\infty^2}{[(1-\gamma)^2T^2]}$.

\emph{Conclusion of part (iii).}~The noise splits as $G_k-\ee{G_k\,|\,{\cal F}_{k-1}} = (G_k^\circ-\ee{G_k^\circ\,|\,{\cal F}_{k-1}})+(\Delta^{\rm b}_k-\ee{\Delta^{\rm b}_k\,|\,{\cal F}_{k-1}})-\Delta^{\rm s}_k$, such that Minkowski's inequality, the three bounds above, and $\sqrt{\ee{\norm{\epsilon_k}_\infty^2}}\le\epsilon_{\rm crit}$ yield
\begin{equation*}
	\sqrt{\ee{\normf{G_k-\ee{G_k\,|\,{\cal F}_{k-1}}}^2}}
	\le \frac{V^\circ+R_\pi\epsilon_{\rm crit}}{\sqrt{T}}+\frac{\gamma R_\pi\epsilon_{\rm crit}}{(1-\gamma)T}
	= \bar\sigma.
\end{equation*}

\emph{Part (iv).}~Let $v^*_{k-1}$ denote the population solution of Theorem~\ref{theorems:lstd-critic-error} for the policy under which $O_{k-1}$ is collected, i.e., $v^*_{k-1} = v^*(W_{k-1})$ for $k\ge2$ and $v^*_0 = v^*(W_1)$.
Splitting $\epsilon_k(s) = \phi^\top(s)(v_{k-1}-v^*_{k-1})+\phi^\top(s)(v^*_{k-1}-v^*(W_k))+(\phi^\top(s)v^*(W_k)-V_{W_k}(s))$ and using $\normsp{\phi(s)}\le1$, Lemma~\ref{lemmas:critic-target-lipschitz}, $\normsp{W_{k-1}-W_k}\le\bar\ell_2$ (the middle term vanishes for $k = 1$), and the definition of $\epsilon_{\rm app}$, we obtain $\norm{\epsilon_k}_\infty\le\normsp{v_{k-1}-v^*_{k-1}}+L_*\bar\ell_2+\epsilon_{\rm app}$, such that Minkowski's inequality gives $\sqrt{\ee{\norm{\epsilon_k}_\infty^2}}\le\sqrt{\ee{\normsp{v_{k-1}-v^*_{k-1}}^2}}+L_*\bar\ell_2+\epsilon_{\rm app}$.
Theorem~\ref{theorems:lstd-critic-error}, applied to frame $k-1$, which starts from the ${\cal F}_{k-2}$-measurable state $s_{k-1,0}$ (for $k = 1$, to the warm-up frame $O_0$ collected under $\pi_1$ from $s_{0,0}\sim\rho$), bounds the first term by $\epsilon_{\rm lstd}$ conditionally on ${\cal F}_{k-2}$ and hence, by the tower property, in $L^2$.

\section{Proof of Lemma \ref{lemmas:momentum-tracking} and Bounded Momenta}\label{apdx:lemma:is_track}
\begin{proof}[Proof of Lemma~\ref{lemmas:momentum-tracking}]
Let $e^M_k := M_k-\nabla J(W_k)$, $\xi_k := G_k-\nabla J(W_k)$, and, for $k\ge2$, $\varsigma_k := G_k-\widetilde G_k-[\nabla J(W_k)-\nabla J(W_{k-1})]$.
Since $W_k$ is ${\cal F}_{k-1}$-measurable, Lemmas~\ref{lemmas:importance-correction}(ii) and~\ref{lemmas:frame-bias-noise}(i),(iii) hold along the iterates, with $\epsilon_{\rm crit}$, $\epsilon_{\rm bias}$, and $\bar\sigma$ evaluated along them.
Subtracting $\nabla J(W_k) = (1-\eta)\nabla J(W_k)+\eta\nabla J(W_{k-1})+\eta[\nabla J(W_k)-\nabla J(W_{k-1})]$ from \eqref{eq:is-momentum} gives $e^M_1 = \xi_1$ and
\begin{equation}\label{eq:momentum-error-recursion}
	e^M_k = \eta\,e^M_{k-1}+(1-\eta)\,\xi_k+\eta\,\varsigma_k,\quad k\ge2.
\end{equation}
By \eqref{eq:actor-mean-field}, Lemma~\ref{lemmas:importance-correction}(ii), $\ee{G_j\,|\,{\cal F}_{j-1}} = h(W_j;\omega_j)$ (Section~\ref{sec:spg-analysis}), and the ${\cal F}_{j-1}$-measurability of $W_j$, $W_{j-1}$, and $\omega_j$, the conditional means are $B_j := \ee{\xi_j\,|\,{\cal F}_{j-1}} = B(W_j;\omega_j)$ for $j\ge1$ and $\bar\varsigma_j := \ee{\varsigma_j\,|\,{\cal F}_{j-1}} = B(W_j;\omega_j)-B(W_{j-1};\omega_j)$ for $j\ge2$.
With $u_1 := \xi_1-B_1$ and $u_j := (1-\eta)(\xi_j-B_j)+\eta(\varsigma_j-\bar\varsigma_j)$ for $j\ge2$, unrolling \eqref{eq:momentum-error-recursion} splits the tracking error as $e^M_k = e_k^{\rm n}+e_k^{\rm b}$ with
\begin{equation*}
	e_k^{\rm n} := \eta^{k-1}u_1+\sum_{j=2}^{k}\eta^{k-j}u_j,
	\qquad
	e_k^{\rm b} := \eta^{k-1}B_1+\sum_{j=2}^{k}\eta^{k-j}\big[(1-\eta)B_j+\eta \bar\varsigma_j\big].
\end{equation*}

\emph{Noise part.}~Each $u_j$ is ${\cal F}_j$-measurable with $\ee{u_j\,|\,{\cal F}_{j-1}} = 0$, such that $\ee{\inpf{u_i,u_j}} = 0$ for $i<j$ by the tower property.
Since $\xi_j-B_j = G_j-\ee{G_j\,|\,{\cal F}_{j-1}}$, Lemma~\ref{lemmas:frame-bias-noise}(iii) gives $\ee{\normf{\xi_j-B_j}^2}\le\bar\sigma^2$.
Since $\nabla J(W_j)-\nabla J(W_{j-1})$ is ${\cal F}_{j-1}$-measurable, $\varsigma_j-\bar\varsigma_j = (G_j-\widetilde G_j)-\ee{G_j-\widetilde G_j\,|\,{\cal F}_{j-1}}$, such that, for $\eta>0$, centering and the last claim of Lemma~\ref{lemmas:importance-correction} give $\ee{\normf{\varsigma_j-\bar\varsigma_j}^2}\le\ee{\normf{G_j-\widetilde G_j}^2}\le C_F^2\bar\ell_2^2$; for $\eta = 0$, this term does not enter $u_j$.
Together with $\normf{X+Y}^2\le2\normf{X}^2+2\normf{Y}^2$, this yields $\ee{\normf{u_1}^2}\le\bar\sigma^2$ and $\ee{\normf{u_j}^2}\le2(1-\eta)^2\bar\sigma^2+2\eta^2C_F^2\bar\ell_2^2$ for $j\ge2$, such that $\sum_{j=2}^{k}\eta^{2(k-j)}\le\nicefrac{1}{(1-\eta^2)}$ and $1-\eta^2\ge1-\eta$ give
\begin{equation}\label{eq:momentum-noise-bound}
	\ee{\normf{e_k^{\rm n}}^2}
	\le \eta^{2(k-1)}\bar\sigma^2+\frac{2(1-\eta)^2\bar\sigma^2+2\eta^2C_F^2\bar\ell_2^2}{1-\eta^2}
	\le \eta^{2(k-1)}\bar\sigma^2+2(1-\eta)\,\bar\sigma^2+\frac{2\eta^2C_F^2\bar\ell_2^2}{1-\eta}
\end{equation}
and Jensen's inequality and $\sqrt{x+y+z}\le\sqrt{x}+\sqrt{y}+\sqrt{z}$ yield $\ee{\normf{e_k^{\rm n}}}\le(\eta^{k-1}+\sqrt{2(1-\eta)})\,\bar\sigma+\nicefrac{\sqrt{2}\,\eta\,C_F\bar\ell_2}{\sqrt{1-\eta}}$, while $\normd{e_k^{\rm n}}\le\kappa_\dagger\normf{e_k^{\rm n}}$ by the definition of $\kappa_\dagger$.

\emph{Bias part.}~By the expressions of $B_j$ and $\bar\varsigma_j$, $(1-\eta)B_j+\eta \bar\varsigma_j = B(W_j;\omega_j)-\eta B(W_{j-1};\omega_j)$, and, with the split $B = \bar B+D$ of \eqref{eq:actor-mean-field}, the context-free part telescopes as
\begin{equation}\label{eq:context-free-bias-telescope}
	\begin{split}
	\eta^{k-1}\bar B(W_1)+\sum_{j=2}^{k}\eta^{k-j}\big[\bar B(W_j)-\eta\bar B(W_{j-1})\big]
&= \sum_{j=1}^{k}\eta^{k-j}\bar B(W_j)-\sum_{j=1}^{k-1}\eta^{k-j}\bar B(W_j) \\
&= \bar B(W_k)
	\end{split}
\end{equation}
such that
\begin{equation*}
	e_k^{\rm b}
	= \bar B(W_k)+\eta^{k-1}D(W_1;\omega_1)+\sum_{j=2}^{k}\eta^{k-j}\Big[(1-\eta)D(W_j;\omega_j)+\eta\big(D(W_j;\omega_j)-D(W_{j-1};\omega_j)\big)\Big]
\end{equation*}
where each difference $D(W_j;\omega_j)-D(W_{j-1};\omega_j)$ holds the context $\omega_j$ fixed, such that no bound on the change of the context between frames is needed.
First, $\normnu{\bar B(W_k)}\le\epsilon_{\rm dist}$.
Second, Lemma~\ref{lemmas:frame-bias-noise}(i) and Jensen's inequality with $\ee{\norm{\epsilon_j}_\infty}\le\epsilon_{\rm crit}$ give $\ee{\normnu{D(W_j;\omega_j)}}\le\epsilon_{\rm bias}-\epsilon_{\rm dist}$ for all $j$, and $\eta^{k-1}+(1-\eta)\sum_{j=2}^{k}\eta^{k-j} = 1$.
Third, Lemma~\ref{lemmas:frame-bias-noise}(ii) gives $\normnu{D(W_j;\omega_j)-D(W_{j-1};\omega_j)}\le D_F\bar\ell_2$ for $j\ge2$, and $\eta\sum_{j=2}^{k}\eta^{k-j}\le\nicefrac{\eta}{(1-\eta)}$.
Hence, $\ee{\normnu{e_k^{\rm b}}}\le\epsilon_{\rm bias}+\nicefrac{\eta D_F\bar\ell_2}{(1-\eta)}$, and, since $\normd{\cdot}\le\normnu{\cdot}$ for $\norm{\cdot}\ge\normsp{\cdot}$ (Appendix~\ref{apdx:notation}), $\ee{\normd{e^M_k}}\le\kappa_\dagger\ee{\normf{e_k^{\rm n}}}+\ee{\normnu{e_k^{\rm b}}}$ yields \eqref{eq:momentum-tracking-generic}; the bounds on $e_k^{\rm n}$ and $e_k^{\rm b}$ above are those stated below Lemma~\ref{lemmas:momentum-tracking}.

\emph{Second moment.}~Lemma~\ref{lemmas:frame-bias-noise}(i) and Minkowski's inequality with $\sqrt{\ee{\norm{\epsilon_j}_\infty^2}}\le\epsilon_{\rm crit}$ give $\sqrt{\ee{\normnu{D(W_j;\omega_j)}^2}}\le\epsilon_{\rm bias}-\epsilon_{\rm dist}$, such that the same three steps, with Minkowski's inequality in $L^2$, give $\sqrt{\ee{\normf{e_k^{\rm b}}^2}}\le\sqrt{\ee{\normnu{e_k^{\rm b}}^2}}\le\epsilon_{\rm bias}+\nicefrac{\eta D_F\bar\ell_2}{(1-\eta)}$; together with \eqref{eq:momentum-noise-bound} and $\normf{X+Y}^2\le2\normf{X}^2+2\normf{Y}^2$, this yields \eqref{eq:momentum-tracking-second-moment}.
For $\eta = 0$, the importance weights are not used, the condition on $\bar\ell_2$ is not needed, and $e^M_k = \xi_k$ gives directly, by Lemma~\ref{lemmas:frame-bias-noise}(i),(iii), $\ee{\normd{e^M_k}}\le\kappa_\dagger\bar\sigma+\epsilon_{\rm bias}$ and $\ee{\normf{e^M_k}^2}\le2\bar\sigma^2+2(\epsilon_{\rm bias})^2$, which are at most the corresponding bounds.
\end{proof}

\emph{Plain momentum (Remark~\ref{rem:is_plain}).}~For $M_k = \eta M_{k-1}+(1-\eta)G_k$ with $M_1 = G_1$, the same subtraction gives $e^M_1 = \xi_1$ and $e^M_k = \eta e^M_{k-1}+(1-\eta)\xi_k-\eta[\nabla J(W_k)-\nabla J(W_{k-1})]$ for $k\ge2$.
By the same orthogonality, the noise part $\eta^{k-1}(\xi_1-B_1)+(1-\eta)\sum_{j=2}^{k}\eta^{k-j}(\xi_j-B_j)$ has second moment at most $\eta^{2(k-1)}\bar\sigma^2+\nicefrac{(1-\eta)^2\bar\sigma^2}{(1-\eta^2)}\le\eta^{2(k-1)}\bar\sigma^2+(1-\eta)\bar\sigma^2$ in the Frobenius norm and hence expected dual norm at most $\kappa_\dagger(\eta^{k-1}+\sqrt{1-\eta})\,\bar\sigma$, and the bias part $\eta^{k-1}B_1+(1-\eta)\sum_{j=2}^{k}\eta^{k-j}B_j$ has expected nuclear norm at most $\epsilon_{\rm bias}$ by Lemma~\ref{lemmas:frame-bias-noise}(i), since $\eta^{k-1}+(1-\eta)\sum_{j=2}^{k}\eta^{k-j} = 1$; without the importance correction, the bias does not telescope, and its change between iterates does not enter.
Lemma~\ref{lemmas:return-smoothness} and $\normsp{W_j-W_{j-1}}\le\beta$, which hold for any normalized step with stepsize $\beta>0$, bound the drift $\sum_{j=2}^{k}\eta^{k-j+1}[\nabla J(W_j)-\nabla J(W_{j-1})]$ by $\nicefrac{\eta L\beta}{(1-\eta)}$ in the nuclear norm and hence in the dual norm, which gives the bound of Remark~\ref{rem:is_plain}.
The proof of Theorem~\ref{theorems:normalized-actor}, which uses Lemma~\ref{lemmas:momentum-tracking} only through \eqref{eq:momentum-tracking-generic} with $\bar\ell_2 = \beta$, applies with this bound, i.e., with $C_F = 0$, $\sqrt{1-\eta}$ in place of $\sqrt{2(1-\eta)}$, and $L$ in place of $D_F$, which gives the floor term $(1+\chi)cL$ of Remark~\ref{rem:is_plain}.

The following lemma bounds the momentum of unnormalized actor updates for every realization, which controls their importance weights.

\begin{lemma}\label{lemmas:baseline-momentum-tracking}
	Under Assumptions~\ref{assum:01} and~\ref{assum:is}, consider the ICM \eqref{eq:is-momentum} together with an actor update whose step $W_{k+1}-W_k$ is ${\cal F}_k$-measurable and satisfies $\normf{W_{k+1}-W_k}\le\bar\beta\normf{M_k}$ for a deterministic $\bar\beta>0$ with, if $\eta>0$, $2T R_\pi R_g\bar\beta\le1$ and $2\eta C_F\bar\beta\le1-\eta$.
	Then $\normf{M_k}\le2R_g$ for all $k\in[K]$, and Lemma~\ref{lemmas:momentum-tracking} applies with $\bar\ell_2 = 2\bar\beta R_g$.
\end{lemma}

\begin{proof}For $\eta = 0$, $M_k = G_k$, and Lemma~\ref{lemmas:critic-gradient-bounds}(iii) gives $\normf{M_k}\le R_g$.
Let $\eta>0$; then $\normf{M_1} = \normf{G_1}\le R_g$.
For $k\ge2$ with $\normf{M_{k-1}}\le2R_g$, the spectral norm of the step is at most its Frobenius norm, such that $\normsp{W_k-W_{k-1}}\le\bar\beta\normf{M_{k-1}}\le2\bar\beta R_g\le\nicefrac{1}{(T R_\pi)}$, such that Lemma~\ref{lemmas:importance-correction}(iii) with $W = W_k$, $W' = W_{k-1}$, $O = O_k$, and $v = v_{k-1}$ gives $\normf{G_k-\widetilde G_k}\le C_F\normsp{W_k-W_{k-1}}\le2C_F\bar\beta R_g$, and \eqref{eq:is-momentum}, Lemma~\ref{lemmas:critic-gradient-bounds}(iii), and $2\eta C_F\bar\beta\le1-\eta$ give
\begin{equation*}
	\begin{split}
	\normf{M_k}
	&\le (1-\eta)\normf{G_k}+\eta\normf{M_{k-1}}+\eta\normf{G_k-\widetilde G_k} \\
	&\le (1-\eta)R_g+2\eta R_g+2\eta C_F\bar\beta R_g
	\le 2R_g;
	\end{split}
\end{equation*}
the bound follows by induction.
In both cases, $\normsp{W_{k+1}-W_k}\le\bar\beta\normf{M_k}\le2\bar\beta R_g$ for all $k$, and $W_k$ is ${\cal F}_{k-1}$-measurable by induction, such that Lemma~\ref{lemmas:momentum-tracking} applies with $\bar\ell_2 = 2\bar\beta R_g$, where $T R_\pi\bar\ell_2\le1$ if $\eta>0$.
\end{proof}

\section{Proof of Theorem \ref{theorems:normalized-actor}}\label{apdx:thm:normalized}
\begin{proof}Let $e^M_k := M_k-\nabla J(W_k)$.
Since the step is ${\cal F}_k$-measurable, every $W_k$ is ${\cal F}_{k-1}$-measurable, and $\normsp{W_k-W_{k-1}} = \beta\normsp{N_{k-1}}\le\beta\norm{N_{k-1}}\le\beta$, such that Lemma~\ref{lemmas:momentum-tracking} applies with $\bar\ell_2 = \beta$, where $T R_\pi\beta\le1$ if $\eta>0$.
Substituting $W = W_k$ and $W' = W_{k+1} = W_k+\beta N_k$ into \eqref{eq:return-ascent-bound} and using $\normsp{N_k}\le\norm{N_k}\le1$ give $\beta\inpf{\nabla J(W_k),N_k}\le J(W_{k+1})-J(W_k)+\nicefrac{L\beta^2}{2}$.
By Definition~\ref{def:steps}(i) and the triangle inequality, $\inpf{M_k,N_k}\ge\chi^{-1}\normd{M_k}\ge\chi^{-1}(\normd{\nabla J(W_k)}-\normd{e^M_k})$, and the definition of the dual norm gives $\inpf{e^M_k,N_k}\le\normd{e^M_k}\norm{N_k}\le\normd{e^M_k}$, such that $\inpf{\nabla J(W_k),N_k} = \inpf{M_k,N_k}-\inpf{e^M_k,N_k}\ge\chi^{-1}\normd{\nabla J(W_k)}-(1+\chi^{-1})\normd{e^M_k}$; hence,
\begin{equation}\label{eq:normalized-one-step-ascent}
	\normd{\nabla J(W_k)}
	\le \chi\,\frac{J(W_{k+1})-J(W_k)}{\beta}+(1+\chi)\normd{e^M_k}+\frac{\chi L\beta}{2}.
\end{equation}
Summing \eqref{eq:normalized-one-step-ascent} over $k\in[K]$, using $J(W_{K+1})\le J^*$, taking expectations, dividing by $K$, and inserting \eqref{eq:momentum-tracking-generic} with $\bar\ell_2 = \beta$ and $\frac{1}{K}\sum_{k=1}^{K}\eta^{k-1}\le\frac{1}{K(1-\eta)}$ yield \eqref{eq:normalized-bound}.
With $1-\eta = K^{-2/3}$ and $\beta = cK^{-2/3}$, the condition $T R_\pi\beta\le1$ is equivalent to $K\ge(cT R_\pi)^{3/2}$, and substituting $\nicefrac{\Delta_1}{(\beta K)} = \nicefrac{\Delta_1}{c}\,K^{-1/3}$, $\nicefrac{L\beta}{2} = \nicefrac{cL}{2}\,K^{-2/3}$, $\nicefrac{\bar\sigma}{[K(1-\eta)]} = \bar\sigma K^{-1/3}$, $\sqrt{2(1-\eta)}\,\bar\sigma = \sqrt{2}\,\bar\sigma K^{-1/3}$, $\nicefrac{\sqrt{2}\eta C_F\beta}{\sqrt{1-\eta}}\le\sqrt{2}\,cC_F K^{-1/3}$, and $\nicefrac{\eta D_F\beta}{(1-\eta)}\le cD_F$ into \eqref{eq:normalized-bound} yields \eqref{eq:normalized-rate}.
\end{proof}

\section{Proof of Theorem \ref{theorems:unnormalized-actor}}\label{apdx:thm:unnormalized}
\begin{proof}
Let $u_k := \nicefrac{1}{\beta_k}$, $e^M_k := M_k-\nabla J(W_k)$, and $\Delta_k := J^*-J(W_k)\in[0,\Delta_{\max}]$, where $\Delta_{\max}\le2R_r$ because $|J(W)|\le R_r$ for every $W$.
The step $W_{k+1}-W_k = \beta_kM_k$ is ${\cal F}_k$-measurable with $\normf{\beta_kM_k}\le\bar\beta\normf{M_k}$ by Definition~\ref{def:steps}(ii), such that Lemma~\ref{lemmas:baseline-momentum-tracking} gives $\normf{M_k}\le2R_g$ and Lemma~\ref{lemmas:momentum-tracking} applies with $\bar\ell_2 = 2\bar\beta R_g$.
Substituting $W = W_k$ and $W' = W_k+\beta_kM_k$ into \eqref{eq:return-ascent-bound}, using $\normsp{M_k}\le\normf{M_k}$, and inserting $\inpf{\nabla J(W_k),M_k} = \frac{1}{2}(\normf{\nabla J(W_k)}^2+\normf{M_k}^2-\normf{e^M_k}^2)$ give
\begin{equation*}
	\frac{\beta_k}{2}\normf{\nabla J(W_k)}^2+\frac{\beta_k(1-L\beta_k)}{2}\normf{M_k}^2
	\le J(W_{k+1})-J(W_k)+\frac{\beta_k}{2}\normf{e^M_k}^2
\end{equation*}
where $L\beta_k\le L\bar\beta\le1$, such that $\normf{\nabla J(W_k)}^2\le2u_k(\Delta_k-\Delta_{k+1})+\normf{e^M_k}^2$.
Since $u_k$ need not be monotone, we sum by parts and use $u_1\le\nicefrac{\sqrt{\vartheta}}{\bar\beta}$, $\Delta_k\in[0,\Delta_{\max}]$, and the variation bound of Definition~\ref{def:steps}(ii):
\begin{equation*}
	\sum_{k=1}^{K}u_k(\Delta_k-\Delta_{k+1})
	= u_1\Delta_1+\sum_{k=2}^{K}(u_k-u_{k-1})\Delta_k-u_K\Delta_{K+1}
	\le \frac{\sqrt{\vartheta}\,\Delta_1+V_u\Delta_{\max}}{\bar\beta}
	= \frac{\Lambda}{2\bar\beta}.
\end{equation*}
Summing over $k\in[K]$, taking expectations, dividing by $K$, and inserting \eqref{eq:momentum-tracking-second-moment} with $\bar\ell_2 = 2\bar\beta R_g$ and $\frac{1}{K}\sum_{k=1}^{K}\eta^{2(k-1)}\le\frac{1}{K(1-\eta^2)}$ yield
\begin{equation*}
\begin{split}
	\frac{1}{K}\sum_{k=1}^{K}\ee{\normf{\nabla J(W_k)}^2}
	&\le \frac{\Lambda}{\bar\beta K}+\frac{2\bar\sigma^2}{K(1-\eta^2)}+4(1-\eta)\,\bar\sigma^2 \\
	&\quad+\frac{16\eta^2C_F^2\bar\beta^2R_g^2}{1-\eta}+2\Big(\epsilon_{\rm bias}+\frac{2\eta D_F\bar\beta R_g}{1-\eta}\Big)^2
\end{split}
\end{equation*}
and Jensen's inequality and $\sqrt{\sum_ix_i}\le\sum_i\sqrt{x_i}$ yield \eqref{eq:unnormalized-bound}.
With $1-\eta = K^{-1/2}$ and $\bar\beta = cK^{-1/2}$, the condition $L\bar\beta\le1$ follows from $K\ge(cL)^2$, the condition $2\eta C_F\bar\beta\le1-\eta$ from $c\le\nicefrac{1}{(2C_F)}$, and the condition $2T R_\pi R_g\bar\beta\le1$ from $c\le\nicefrac{1}{(2C_F)}$ and $\bar\beta\le c$, since $C_F\ge(\exp(1)-1)TR_AR_\pi^2$ and $R_g = R_AR_\pi$ give $2cTR_\pi R_g\le\nicefrac{1}{(\exp(1)-1)}<1$.
Substituting $\sqrt{\nicefrac{\Lambda}{(\bar\beta K)}} = \sqrt{\nicefrac{\Lambda}{c}}\,K^{-1/4}$, $\nicefrac{1}{\sqrt{K(1-\eta^2)}}\le\nicefrac{1}{\sqrt{K(1-\eta)}} = K^{-1/4}$, $\sqrt{1-\eta} = K^{-1/4}$, $\nicefrac{\eta\bar\beta}{\sqrt{1-\eta}}\le cK^{-1/4}$, and $\nicefrac{\eta\bar\beta}{(1-\eta)}\le c$ into \eqref{eq:unnormalized-bound} yields \eqref{eq:unnormalized-rate}.
\end{proof}

\section{Proofs for the Instances}\label{apdx:instances}

\subsection{Proof of Lemma~\ref{lemmas:nso-polar-convergence}}
\begin{proof}\emph{Part (i).}~If $M_k = 0$, the NSO iterates, $\Theta_k$, and $R_k$ vanish by convention.
Otherwise, let $M_k = U_k\Sigma_kV_k^\top$ be the compact singular value decomposition with $\Sigma_k\in\mathbb{R}^{\bar r_k\times\bar r_k}$, $\bar r_k := \rank{M_k}$, and $U_k^\top U_k = V_k^\top V_k = I_{\bar r_k}$, such that $\Theta_k = U_kV_k^\top$.
By induction over \eqref{eq:nso-step} from $N_{k,0} = \nicefrac{M_k}{\normf{M_k}}$, $N_{k,p} = U_k\Sigma_{k,p}V_k^\top$ with $\Sigma_{k,0} = \nicefrac{\Sigma_k}{\normf{M_k}}$: indeed, $I_n-N_{k,p}^\top N_{k,p} = (I_n-V_kV_k^\top)+V_k(I_{\bar r_k}-\Sigma_{k,p}^2)V_k^\top$ and $(I_n-V_kV_k^\top)V_k = 0$ give $N_{k,p}(I_n-N_{k,p}^\top N_{k,p})^i = U_k\Sigma_{k,p}(I_{\bar r_k}-\Sigma_{k,p}^2)^iV_k^\top$ for all $i\ge0$, such that \eqref{eq:nso-step} maps each singular value $\sigma_{k,p}$ of $N_{k,p}$ to
\begin{equation}\label{eq:nso-singular-value-update}
	\sigma_{k,p+1} = f(\sigma_{k,p}) = \sigma_{k,p}\sum_{i=0}^{\kappa_2-1}\frac{1}{4^i}\binom{2i}{i}(1-\sigma_{k,p}^2)^i.
\end{equation}

Matching the coefficients of $(1-x^2)^j$ gives $f'(x) = (2\kappa_2-1)\binom{2\kappa_2-2}{\kappa_2-1}4^{1-\kappa_2}(1-x^2)^{\kappa_2-1}\ge0$ on $[0,1]$, such that $f^p$ preserves the order of the singular values and the $i$th largest singular value of $N_{k,p}$ is $f^{p}(\nicefrac{\sigma_i(M_k)}{\normf{M_k}})$.

\emph{Monotone convergence and local order.}~Since $\normsp{M_k}\le\normf{M_k}$, every nonzero singular value satisfies $\sigma_{k,0}\in(0,1]$, and the Maclaurin series $(1-x)^{-1/2} = \sum_{i\ge0}4^{-i}\binom{2i}{i}x^i$, $x\in[0,1)$, with $x = 1-\sigma_{k,p}^2$ gives
\begin{equation}\label{eq:nso-binomial-identity}
	\sigma_{k,p}\sum_{i=0}^{\infty}\frac{1}{4^i}\binom{2i}{i}(1-\sigma_{k,p}^2)^i = 1.
\end{equation}
Since its summands are nonnegative, \eqref{eq:nso-singular-value-update} gives $\sigma_{k,p+1}\le1$; since the $i = 0$ term of the sum in \eqref{eq:nso-singular-value-update} equals one and the others are positive for $\sigma_{k,p}\in(0,1)$, $\sigma_{k,p+1}>\sigma_{k,p}$ there, and $\sigma = 1$ is the unique fixed point in $(0,1]$.
Hence, $\normsp{N_{k,p}}\le1$, each $\sigma_{k,p}$ increases monotonically to one, and $N_{k,p}\to\Theta_k$.
With $e_{k,p} := 1-\sigma_{k,p}$, subtracting \eqref{eq:nso-singular-value-update} from \eqref{eq:nso-binomial-identity} gives, as $e_{k,p}\to0$,
\begin{equation*}
	e_{k,p+1}
	= (1-e_{k,p})\sum_{i=\kappa_2}^{\infty}\frac{1}{4^i}\binom{2i}{i}\big[e_{k,p}(2-e_{k,p})\big]^i
	= \frac{1}{2^{\kappa_2}}\binom{2\kappa_2}{\kappa_2}e_{k,p}^{\kappa_2}+\oo{e_{k,p}^{\kappa_2+1}}
\end{equation*}
such that, by \citet[Definition~4.1.1]{Kelley1995}, each singular value converges to one locally with order $\kappa_2$, and so does $N_{k,p}$ to $\Theta_k$, since $\normsp{N_{k,p}-\Theta_k} = \normsp{\Sigma_{k,p}-I_{\bar r_k}} = \max_ie_{k,p,i}$ with the error $e_{k,p,i}$ of the $i$th singular value.

\emph{Global contraction.}~Let $\delta_{k,p} := 1-\sigma_{k,p}^2$ and $\bar\psi(\delta) := \sum_{i=0}^{\kappa_2-1}4^{-i}\binom{2i}{i}\delta^i$.
Since $\bar\psi$ truncates the series of $(1-\delta)^{-1/2}$, whose square is $\sum_{j\ge0}\delta^j$, the coefficients of $\bar\psi^2$ equal one up to degree $\kappa_2-1$ and are nonnegative beyond; hence, $\bar\psi(\delta)^2\ge\sum_{j=0}^{\kappa_2-1}\delta^j$ on $[0,1]$, and \eqref{eq:nso-singular-value-update} gives $0\le\delta_{k,p+1} = 1-(1-\delta_{k,p})\bar\psi(\delta_{k,p})^2\le\delta_{k,p}^{\kappa_2}$.
Therefore, $1-\sigma_{k,P_2}\le1-(1-(1-\sigma_{k,0}^2)^{\kappa_2^{P_2}})^{1/2}\le(1-\sigma_{k,0}^2)^{\kappa_2^{P_2}}$ for every nonzero singular value, and $\sigma_{k,0}\ge a_k>0$ gives $\normsp{R_k} = \max_ie_{k,P_2,i} = 1-\sigma_{\min}^+(N_{k,P_2})\le(1-a_k^2)^{\kappa_2^{P_2}}<1$.

\emph{Part (ii).}~If $M_k = 0$, both sides of \eqref{eq:nso-alignment} vanish.
Otherwise, with the nonzero singular values $\sigma_1,\ldots,\sigma_{\bar r_k}$ of $M_k$ and $x_i := \nicefrac{\sigma_i}{\normf{M_k}}\in(0,1]$, part (i) gives $\normnu{M_k}-\inpf{M_k,N_{k,P_2}} = \sum_{i=1}^{\bar r_k}\sigma_i(1-f^{P_2}(x_i)) = \normf{M_k}\sum_{i=1}^{\bar r_k}x_i(1-f^{P_2}(x_i))\le\bar r_ks_{P_2}\normf{M_k}\le r\,s_{P_2}\normnu{M_k}$, since $\bar r_k\le r$ and $\normf{M_k}\le\normnu{M_k}$.
Finally, the global contraction gives $x(1-f^{P}(x))\le x(1-x^2)^{\kappa_2^{P}}\le x\exp(-\kappa_2^{P}x^2)\le(2e\kappa_2^{P})^{-1/2}$ for $x\in[0,1]$, where the last inequality maximizes over $x$; hence, $s_P\le(2e\kappa_2^{P})^{-1/2}$, and $\kappa_2^{P_2}\ge4r^2$ gives $r\,s_{P_2}\le(8e)^{-1/2}<0.22$.
\end{proof}

\subsection{Proofs of Corollaries~\ref{cor:nsac}--\ref{cor:sgdm-adanorm}}

Hereinafter, we provide the proofs of Corollaries~\ref{cor:nsac}--\ref{cor:sgdm-adanorm}.

\emph{\bfseries Muon.}~The NSO output $N_{k,P_2}$ is a function of $M_k$ and hence ${\cal F}_k$-measurable, and Lemma~\ref{lemmas:nso-polar-convergence} gives $\normsp{N_{k,P_2}}\le1$ and $\inpf{M_k,N_{k,P_2}}\ge(1-r\,s_{P_2})\normnu{M_k} = \chi^{-1}\normnu{M_k}$ with $\chi\ge1$ if $r\,s_{P_2}<1$.
Hence, the Muon step is normalized in the spectral norm, whose dual norm is the nuclear norm with $\kappa_\dagger = r^{1/2}$, and Theorem~\ref{theorems:normalized-actor} gives \eqref{eq:normalized-bound} and \eqref{eq:normalized-rate} in the nuclear norm, i.e., \eqref{eq:nsac-stationarity-rate}; Corollary~\ref{cor:eps-guarantee}(i) applies verbatim.
If $\kappa_2^{P_2}\ge4r^2$, then $r\,s_{P_2}\le(8e)^{-1/2}<0.215$, such that $\chi<\nicefrac{1}{0.785}<1.28$.

\emph{\bfseries NSGDM.}~The direction $N_k := \nicefrac{M_k}{\normf{M_k}}$ if $M_k\neq0$ and $N_k := 0$ otherwise is ${\cal F}_k$-measurable with $\normf{N_k}\le1$ and $\inpf{M_k,N_k} = \normf{M_k}$.
Hence, the NSGDM step is normalized with $\chi = 1$ in the Frobenius norm, which satisfies $\normf{\cdot}\ge\normsp{\cdot}$ and is self-dual with $\kappa_\dagger = 1$, and Theorem~\ref{theorems:normalized-actor} bounds $\frac{1}{K}\sum_{k=1}^{K}\ee{\normf{\nabla J(W_k)}}$ by \eqref{eq:normalized-bound} and, with its schedule, by the bracket of \eqref{eq:nsgdm-stationarity-rate}, such that $\normnu{\nabla J(W_k)}\le r^{1/2}\normf{\nabla J(W_k)}$, which holds since $\nabla J(W_k)$ has rank at most $r$, yields \eqref{eq:nsgdm-stationarity-rate}.
At the same $c$, \eqref{eq:nsgdm-stationarity-rate} is \eqref{eq:nsac-stationarity-rate} with $r^{1/2}$ in place of $\chi$ on the optimization and second-order terms, $2$ in place of $1+\chi$ on the noise and correction terms, and $2r^{1/2}(\epsilon_{\rm bias}+cD_F)$ in place of the floor $(1+\chi)(\epsilon_{\rm bias}+cD_F)$; together with the monotonicity of each term of \eqref{eq:nsac-stationarity-rate} in $c$ and $\nicefrac{(1+\chi)}{2}\le\chi$, this proves the comparison in Section~\ref{sec:comparison}.

\emph{\bfseries SGDM.}~The step $\beta_kM_k$ with the deterministic $\beta_k = \beta$ is unnormalized with $\bar\beta = \beta$, $\vartheta = 1$, and $V_u = 0$, such that $\Lambda = 2\Delta_1$, and Theorem~\ref{theorems:unnormalized-actor} and $\normnu{\nabla J(W_k)}\le r^{1/2}\normf{\nabla J(W_k)}$ yield the nuclear form of \eqref{eq:unnormalized-bound} and, with $\beta = cK^{-1/2}$, \eqref{eq:sgdm-stationarity-rate}.

\emph{\bfseries AdaNorm.}~Let $\beta_k := \nicefrac{\beta}{\sqrt{n_k}}$ and $u_k := \nicefrac{1}{\beta_k}$, and recall $\bar\beta = \beta\sqrt{\nicefrac{2}{n_0}}$ and $\vartheta = 2+\nicefrac{R_g^2}{n_0}$.
Unrolling \eqref{eq:madagrad-update} gives $n_k = (1-\alpha)^kn_0+\alpha\sum_{j=1}^{k}(1-\alpha)^{k-j}\normf{G_j}^2$, a convex combination of $n_0$ and $\normf{G_1}^2,\ldots,\normf{G_k}^2$ in which the squared gradient norms have the total weight $\alpha_k := 1-(1-\alpha)^k\le k\alpha\le\nicefrac{1}{2}$ for $k\in[K]$ by Bernoulli's inequality and $\alpha K\le\nicefrac{1}{2}$; hence, $\normf{G_j}\le R_g$ (Lemma~\ref{lemmas:critic-gradient-bounds}(iii)) gives, for $k\in[K]$,
\begin{equation}\label{eq:madagrad-accumulator-bounds}
	\frac{n_0}{2}\le(1-\alpha_k)\,n_0\le n_k\le(1-\alpha_k)\,n_0+\alpha_kR_g^2\le n_0+\frac{R_g^2}{2},
	\quad\mbox{i.e.,}\quad
	\frac{\bar\beta}{\sqrt{\vartheta}}\le\beta_k\le\bar\beta.
\end{equation}

Moreover, $|n_k-n_{k-1}| = \alpha|\normf{G_k}^2-n_{k-1}|\le\alpha(n_0+R_g^2)$, since both $\normf{G_k}^2$ and $n_{k-1}$ lie in $[0,n_0+R_g^2]$, and $\sqrt{n_k}+\sqrt{n_{k-1}}\ge\sqrt{2n_0}$, such that $\alpha K\le\nicefrac{1}{2}$ gives
\begin{equation*}
	\sum_{k=2}^{K}|u_k-u_{k-1}|
	= \frac{1}{\beta}\sum_{k=2}^{K}\frac{|n_k-n_{k-1}|}{\sqrt{n_k}+\sqrt{n_{k-1}}}
	\le \frac{K\alpha\,(n_0+R_g^2)}{\beta\sqrt{2n_0}}
	\le \frac{n_0+R_g^2}{2\beta\sqrt{2n_0}}
	= \frac{\vartheta-1}{2\bar\beta}.
\end{equation*}

Since $n_k$, and hence $\beta_k$, is ${\cal F}_k$-measurable, \eqref{eq:madagrad-accumulator-bounds} and this variation bound show that the AdaNorm step is unnormalized with $\bar\beta$, $\vartheta$, and $V_u = \nicefrac{(\vartheta-1)}{2}$, such that $\Lambda = 2\sqrt{\vartheta}\,\Delta_1+(\vartheta-1)\Delta_{\max}$, and Theorem~\ref{theorems:unnormalized-actor} and $\normnu{\nabla J(W_k)}\le r^{1/2}\normf{\nabla J(W_k)}$ yield the nuclear form of \eqref{eq:unnormalized-bound}.
With $\beta = c\sqrt{\nicefrac{n_0}{2}}\,K^{-1/2}$, we have $\bar\beta = cK^{-1/2}$, and, for a $K$-independent $n_0$, $\vartheta$ and $\Lambda$ do not depend on $K$, such that the schedule of Theorem~\ref{theorems:unnormalized-actor} yields \eqref{eq:sgdm-stationarity-rate}.

\section{Consequences of the Actor Bounds}\label{apdx:actor-implications}
\subsection{Residual Error}\label{apdx:actor-floor}
By Lemma~\ref{lemmas:frame-bias-noise}(iv) and the schedule of Theorem~\ref{theorems:normalized-actor}, the floor term of \eqref{eq:normalized-rate} is at most
\begin{equation}\label{eq:nsac-error-floor}
	\begin{split}
		(1+\chi)\,\Big[&\epsilon_{\rm dist}+\frac{4R_rR_\pi\tau_{\mix}}{(1-\gamma)T}+cD_F \\
		&+\frac{\gamma R_\pi}{(1-\gamma)T}\Big(R_v\exp(-C_E\kappa_1^{P_1})
		+ 2C_v\sqrt{\frac{1+4\tau_{\mix}}{T}}
		+ \frac{cL_*}{K^{2/3}} + \epsilon_{\rm app}\Big)\Big].
	\end{split}
\end{equation}

Note that \eqref{eq:nsac-error-floor} separates the stationary-to-discounted mismatch, the start-state transient, the drift of the frame bias, and the critic bootstrap.
Within the bootstrap term, the NSI, sampling, and approximation errors are attenuated by $\nicefrac{\gamma R_\pi}{[(1-\gamma)T]}$, whereas the policy drift of the lagged critic belongs to the transient.
In addition to the $\nicefrac{1}{T}$ decay in Remark~\ref{rem:scope}, the sampling part of the critic term decays as $T^{-3/2}$ for fixed $\lambda>0$ and $c$.

\emph{Mismatch despite one-step mixing.}~Let ${\cal S}={\cal A}=\{0,1\}$, $s_{t+1}=a_t$, $\rho=\delta_0$, $r(s,a)=(s-\bar c)a$, and $\pi_w(1|s)=p:=(1+e^{-w})^{-1}$.
Although the chain mixes in one step, we have
\[
J(w)=\gamma p^2-\bar c p,\qquad
J'(w)=p(1-p)(2\gamma p-\bar c),\qquad
g(w)=p(1-p)((1+\gamma)p-\bar c).
\]

At $\gamma=1/2$, $\bar c=3/4$, and $p=1/2$, the stationary mean satisfies $g(w)=0$, whereas $J'(w)=-1/16$.
Therefore, fast mixing alone does not remove the stationary-to-discounted mismatch.

\subsection{Frame Length and Transition Complexity}\label{apdx:frame-complexity}
Reducing the $\oo{\nicefrac{1}{T}}$ part of the floor requires a larger $T$, which also enlarges the correction constant $C_F$.
For every $c>0$, the inequality of arithmetic and geometric means gives $\Gamma\ge2\big[\sqrt{2}\chi(1+\chi)\kappa_\dagger\Delta_1C_F\big]^{1/2}$. 
Since $C_F\ge(\exp(1)-1)R_AR_\pi^2T$, the sufficient horizon $(\nicefrac{\Gamma}{\epsilon})^3$ of Corollary~\ref{cor:eps-guarantee}(i) grows at least as $T^{3/2}$ when $\Delta_1>0$.
Note that such growth is a property of the sufficient bound rather than a lower bound on the number of updates required by Algorithm~\ref{alg:01}.

For fixed $\lambda$ and problem constants, we consider a target $\epsilon\in(0,1]$ for the part of \eqref{eq:normalized-bound} other than $(1+\chi)\epsilon_{\rm dist}$.
The $\oo{\nicefrac{1}{T}}$ part of the floor is of order $\epsilon$ for $T$ of order $\nicefrac{1}{\epsilon}$.
Moreover, in \eqref{eq:normalized-bound}, the choices $1-\eta$ of order $\nicefrac{\epsilon^2}{\bar\sigma^2}$, $\beta$ of order $\nicefrac{\epsilon(1-\eta)^{1/2}}{C_F}$, and $K$ of order $\nicefrac{1}{(\epsilon\beta)}$, i.e., of order $\nicefrac{\bar\sigma C_F}{\epsilon^3}$, make every other term of order $\epsilon$, including the drift $\nicefrac{\eta D_F\beta}{(1-\eta)}$ that is of order $\nicefrac{D_F\bar\sigma}{C_F}$, and satisfy $TR_\pi\beta\le1$ for small $\epsilon$.
Note that these choices are the schedule $1-\eta\propto K^{-2/3}$ and $\beta\propto K^{-2/3}$ with constants that depend on $\bar\sigma$ and $C_F$.
With $\bar\sigma = \oo{T^{-1/2}}$ and $C_F = \oo{T}$, these choices give $\oo{T^{1/2}\epsilon^{-3}} = \oo{\epsilon^{-7/2}}$ updates and, at $T$ transitions per update, $\oo{\epsilon^{-9/2}}$ transitions.
With the fixed proportion $1-\eta = K^{-2/3}$ of Theorem~\ref{theorems:normalized-actor} and the stepsize constant $c$ of order $T^{-1/2}$ that minimizes $\Gamma$ as shown above, the horizon is of order $T^{3/2}\epsilon^{-3}$, i.e., $\oo{\epsilon^{-9/2}}$ updates and $\oo{\epsilon^{-11/2}}$ transitions.
Without momentum, i.e., $\eta = 0$, \eqref{eq:normalized-bound} needs neither the importance correction nor the step condition, and contains the noise term $\sqrt{2}(1+\chi)\kappa_\dagger\bar\sigma = \oo{T^{-1/2}}$ instead of the correction noise and the drift; hence, $T$ of order $\epsilon^{-2}$, $\beta$ of order $\epsilon$, and $K$ of order $\epsilon^{-2}$ give $\oo{\epsilon^{-2}}$ updates and $\oo{\epsilon^{-4}}$ transitions.
Therefore, the ICM pays for the growth of $C_F$ with $T$ in transition complexity, whereas the comparison at a fixed frame length concerns the reducible part of the floor, which decreases from $\oo{T^{-1/2}}$ for $\eta = 0$ to $\oo{\nicefrac{1}{T}}$.

\end{appendices}

\bibliography{ml-muon-a2c}

\end{document}